\documentclass[10pt]{article} 
\usepackage[preprint]{tmlr}

\usepackage{amsmath,amsfonts,bm}

\def\eqref#1{equation~\ref{#1}}

\def\1{\bm{1}}

\DeclareMathAlphabet{\mathsfit}{\encodingdefault}{\sfdefault}{m}{sl}
\SetMathAlphabet{\mathsfit}{bold}{\encodingdefault}{\sfdefault}{bx}{n}

\usepackage{hyperref}
\usepackage{url}

\usepackage{url}            
\usepackage{booktabs}       
\usepackage{amsfonts}       
\usepackage{nicefrac}       
\usepackage{microtype}      
\usepackage{xcolor}         
\usepackage{natbib}
\usepackage{graphicx}
\usepackage{hyperref}       
\usepackage{wrapfig}
\usepackage{amsmath}
\usepackage{tabularx}
\usepackage{multirow}
\usepackage{caption}
\usepackage{afterpage}
\usepackage{subfig}
\usepackage{array}
\usepackage{wrapfig}
\usepackage{enumitem}
\usepackage{listings}
\usepackage{pythonhighlight}

\usepackage{xspace}
\def\modelname{GIT\xspace}
\def\modelnamewospace{GIT}

\title{
	GIT: A Generative Image-to-text Transformer \\for Vision and Language
}

\author{\name Jianfeng Wang \email jianfw@microsoft.com \\
      \name Zhengyuan Yang \email zhengyang@microsoft.com \\
      \name Xiaowei Hu \email xiaowei.hu@microsoft.com \\
      \name Linjie Li \email lindsey.li@microsoft.com \\
      \name Kevin Lin \email keli@microsoft.com \\
      \name Zhe Gan \email zhe.gan@microsoft.com \\
      \name Zicheng Liu \email zliu@microsoft.com \\
      \name Ce Liu \email ce.liu@microsoft.com \\
      \name  Lijuan Wang \email lijuanw@microsoft.com \\
      \addr Microsoft Cloud and AI
      }

\def\month{11}  
\def\year{2022} 
\def\openreview{\url{https://openreview.net/forum?id=b4tMhpN0JC}} 

\begin{document}

\maketitle

\vspace{-0.5cm}
\begin{abstract}
In this paper, we design and train 
a \textbf{G}enerative \textbf{I}mage-to-text \textbf{T}ransformer, \modelname, to unify vision-language tasks such as image/video captioning and question answering. 
While 
generative models provide a consistent network architecture between pre-training and fine-tuning, existing work typically contains complex structures (uni/multi-modal encoder/decoder) and depends on external modules such as object detectors/taggers and optical character recognition (OCR). 
In \modelname, we simplify the architecture as  one image encoder and one
text decoder under a single language modeling task.
We also scale up the 
pre-training data and the model size
to boost the model performance.
Without bells and whistles, our \modelname 
establishes new  state of the arts on numerous challenging benchmarks
with a large margin.
For instance, 
our model
surpasses the human performance for the first time on TextCaps (138.2 vs. 125.5 in CIDEr).
Furthermore, we present a new scheme of generation-based image classification and scene text recognition, achieving decent performance on standard benchmarks.

\end{abstract}

\vspace{-0.5cm}
\section{Introduction}


\begin{table}[h]
    \centering
    \small
    \caption{
    Comparison with prior SOTA on image/video captioning and question answering (QA) tasks.
    *: evaluated on the public server. CIDEr scores are reported for Captioning tasks.
    Prior SOTA:
	COCO\citep{abs-2101-00529}, 
    nocaps \citep{yu2022coca}, 
    VizWiz-Caption \citep{GongZWCZSL21},   
    TextCaps \citep{abs-2012-04638},
    ST-VQA \citep{biten2022latr},
    VizWiz-VQA \citep{alayrac2022flamingo},
    OCR-VQA \citep{biten2022latr},
    MSVD \citep{abs-2111-13196},
    MSRVTT \citep{abs-2201-08264},
    VATEX \citep{abs-2110-05204},
    TVC \citep{abs-2110-05204},
    MSVD-QA \citep{wang2022all},
    TGIF-Frame \citep{ZellersLHYPCFC21},
    Text Recog. \citep{lyu2022maskocr}.
    ~Details of \modelnamewospace2 are presented in supplementary materials.
    }
    \label{tab:sota}
    \begin{tabular}{@{}c@{~}c@{~~}c@{~~}c@{~~}c@{~~}c@{~~}c@{~~}c@{~~}c@{~~}c@{~~}c@{~~}c@{~~}c@{~}c@{}c@{}} \toprule
                                & \multicolumn{4}{c}{Image captioning}     & \multicolumn{3}{c}{Image QA} & \multicolumn{4}{c}{Video captioning} & \multicolumn{2}{c}{Video QA} & Text Rec. \\
                                \cmidrule(lr){2-5} \cmidrule(lr){6-8} \cmidrule(lr){9-12} \cmidrule(lr){13-14}
                                \cmidrule(lr){15-15}
                                & \rotatebox[origin=c]{90}{COCO$^*$}   
                                & \rotatebox[origin=c]{90}{nocaps$^*$} 
                                & \rotatebox[origin=c]{90}{VizWiz$^*$}
                                & \rotatebox[origin=c]{90}{TextCaps$^*$}
                                & \rotatebox[origin=c]{90}{ST-VQA$^*$}
                                & \rotatebox[origin=c]{90}{VizWiz$^*$}
                                & \rotatebox[origin=c]{90}{OCR-VQA}
                                & \rotatebox[origin=c]{90}{MSVD}
                                & \rotatebox[origin=c]{90}{MSRVTT}
                                & \rotatebox[origin=c]{90}{VATEX$^*$}
                                & \rotatebox[origin=c]{90}{TVC$^*$}
                                & \rotatebox[origin=c]{90}{MSVD-QA}
                                & \rotatebox[origin=c]{90}{TGIF-Frame}
                                & \rotatebox[origin=c]{90}{Avg on 6} \\
                   \midrule
    Prior SOTA\protect\footnotemark &  138.7  & 120.6   & 94.1   &  109.7 & 69.6  & 65.4  & 67.9    & 120.6 & 60      & 86.5 & 64.5  &  48.3        & 69.5 & 93.8 \\
                                \midrule
    \modelname (ours)           & 148.8   & 123.4 & 114.4   & 138.2 & 69.6   & 67.5   & 68.1  &    180.2   &  73.9   & 93.8  & 61.2   & 56.8   &       72.8 & 92.9 \\
    $\Delta$                    & +10.1    & +2.8  & +20.3   & +28.5 & +0.0   & +2.1   & +0.2  & +59.6      & +13.9    & +7.3    & -3.3   & +8.5  &      +3.3   & -0.9 \\
    \midrule
    \modelnamewospace2 (ours)   & 149.8   & 124.8 & 120.8   & 145.0 & 75.8    & 70.1   & 70.3  &    185.4   &  75.9   & 96.6 & 65.0    & 58.2   &       74.9 & 94.5 \\
    $\Delta$                    & +11.1   & + 4.2 & +26.7   & +35.3 & +6.2 & +4.7 & +2.4 & +64.8 & +15.9 & +10.1 & +0.5 & +9.9 & +5.4 & +0.7 \\
    \bottomrule
    \end{tabular}
\end{table}

\footnotetext{Prior SOTA: among all the numbers reported in publications 
before 8/2022, as far as we know.}

Tremendous advances have been made in recent years 
on vision-language (VL) pre-training, especially based on 
the large-scale data of image-text pairs, \emph{e.g.}, CLIP~\citep{clip}, 
Florence~\citep{abs-2111-11432}, and SimVLM~\citep{wang2021simvlm}.
The learned representation greatly boosts the performance on various downstream tasks, 
such as image captioning~\citep{LinMBHPRDZ14}, visual question answering (VQA)~\citep{GoyalKSBP16}, and image-text retrieval.


During pre-training, Masked Language Modeling (MLM)  and 
Image-Text Matching (ITM) tasks
have been widely used~\citep{abs-2012-06946,FangW0WY021,Li0LZHZWH0WCG20,abs-2101-00529,ChenLYK0G0020,abs-2111-02387,ufo,KimSK21}. 
However, these losses are different from the
downstream tasks, and task-specific adaptation has to be made. 
For example, ITM is removed for image captioning~\citep{ufo,Li0LZHZWH0WCG20}, and 
an extra randomly initialized multi-layer perceptron is added for VQA~\citep{wang2021simvlm,Li0LZHZWH0WCG20}.
To reduce this discrepancy, recent approaches~\citep{cho2021unifying,wang2021simvlm,abs-2111-12085,abs-2202-03052} have attempted to design unified generative models for pre-training, as most VL tasks can be cast as 
generation problems.
These approaches typically 
leverage a multi-modal encoder and a text 
decoder with careful design on the text input and the text target.
To further push the frontier of this 
direction, we present a simple Generative Image-to-text Transformer,
named \modelname,
which consists only of one image encoder and one text decoder.
The pre-training task is just to map the input image to the entire associated text description
with the language modeling objective.
Despite its simplicity, \modelname~achieves new  state of the arts 
across numerous challenging benchmarks
with a large margin, as summarized in Table~\ref{tab:sota}.



\begin{figure}[t!]
    \centering
    \includegraphics[width=\textwidth]{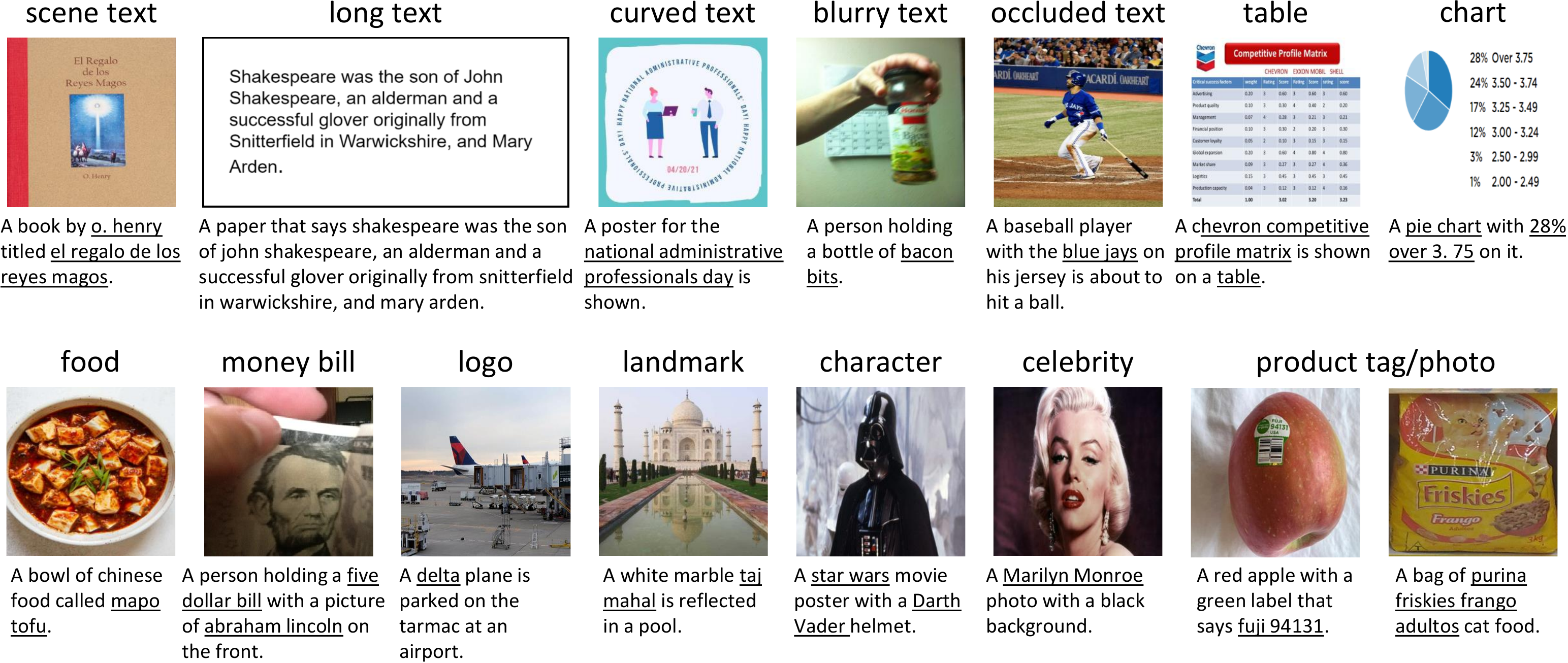}
    \caption{Example captions generated by \modelname. The model demonstrates strong capability of recognizing  scene text, tables/charts, food, banknote, logos, landmarks, characters, products, \emph{etc.}} 
    \label{fig:teaser}
\end{figure}


The image encoder is a Swin-like vision transformer~\citep{DosovitskiyB0WZ21,abs-2111-11432} 
pre-trained on massive image-text pairs based on the contrastive 
task~\citep{JiaYXCPPLSLD21,clip,abs-2111-11432}.
This eliminates the dependency on the object detector, which 
is used in many existing approaches~\citep{00010BT0GZ18,Li0LZHZWH0WCG20,abs-2012-06946,abs-2101-00529,ChenLYK0G0020,FangW0WY021}.
To extend it to the video domain, we simply extract 
the features of multiple sampled frames and concatenate them as the video representation.
The text decoder is a transformer network 
to predict the associated text.
The entire network is trained with the language modeling task.
For VQA,  the input question is treated as a text prefix, and the answer is generated in an auto-regressive way. 
Furthermore, we present a new generation-based scheme for ImageNet classification, 
where the predicted labels come directly from our generative model without pre-defining the vocabulary.

The approach is simple, but the performance is surprisingly impressive after we scale up the pre-training data and the model size.
Fig.~\ref{fig:teaser} shows captions generated by the \modelname fine-tuned with TextCaps.
The samples demonstrate the model's strong capability of 
recognizing and describing scene text, tables, charts, food, banknote, logos, landmarks, characters, celebrities, products, \emph{etc.}, indicating that our \modelname model has encoded rich multi-modal knowledge about the visual world.

Our main contributions are as follows.
\begin{itemize}[leftmargin=*]
\setlength\itemsep{0.1mm}
    \item We present \modelname, which consists of only one image encoder and one text decoder, pre-trained on 0.8 billion image-text pairs with the language modeling task.
    \item We demonstrate new  state-of-the-art performance over numerous tasks on image/video captioning and QA (Table~\ref{tab:sota}), without the dependency on object detectors, object tags, and OCR.
    On TextCaps, we surpass the human performance for the first time. 
    This implies that a simple network architecture can also achieve strong performance with scaling.
    \item We demonstrate that \modelname~pre-trained on the image-text pairs 
    is capable of achieving new state-of-the-art performance even on video tasks without video-dedicated encoders. 
    \item We present a new scheme of generation-based image classification. On ImageNet-1K, we show a decent performance (88.79\% top-1 accuracy) with our \modelname.
\end{itemize}
\section{Related Work}
In VL pre-training, multi-task pre-training has been widely used to 
empower the network with multiple or enhanced capabilities.
For example, MLM and ITM are widely adopted pre-training tasks~\citep{Li0LZHZWH0WCG20,KimSK21,abs-2101-00529,abs-2012-06946,xue2021probing,LuBPL19,TanB19}.
Recently, the image-text contrastive loss has also been added
in~\cite{yu2022coca,li2021align,ufo}.
Since most VL tasks can be formulated as the text generation task~\citep{cho2021unifying},
a single generation model can be pre-trained to support various downstream tasks.
The input and output texts are usually carefully designed to pre-train such a generation model. 
For example in~\cite{cho2021unifying}, the text is properly
masked as the network input and the goal is to recover the masked 
text span. 
SimVLM~\citep{wang2021simvlm} randomly splits a text sentence into the input and the target output.
In these methods, a multi-modal transformer encoder is utilized 
to incorporate the text inputs before decoding the output.

For image representation, Faster RCNN has been used 
in most existing approaches~\citep{00010BT0GZ18,Li0LZHZWH0WCG20,abs-2012-06946,abs-2101-00529,ChenLYK0G0020,FangW0WY021}
to extract the region features.
Recently, a growing interest is in dense representation~\citep{abs-2004-00849,wang2021simvlm,ufo,KimSK21,abs-2112-05230,abs-2111-02387,li2021align} from the feature map,
which requires no bounding box annotations. 
Meanwhile, it is easy to train the entire network in an end-to-end way. 
In addition to the representation from the feature map, object tags~\citep{Li0LZHZWH0WCG20,abs-2012-06946,abs-2101-00529,abs-2111-12727,abs-2112-05230} are leveraged to
facilitate the transformer to understand the context, especially the novel objects.
For scene-text-related tasks, OCR is invoked to generate 
the scene text as additional network input, \emph{e.g.}, 
in~\cite{HuSDR20,abs-2012-04638}.
For the text prediction, 
A transformer network is typically used,
which can incorporate the cross-attention module to fuse the image tokens, \emph{e.g.},~\cite{cho2021unifying,alayrac2022flamingo,abs-2111-12085,yu2022coca},
or 
only the self-attention modules where the image tokens are 
concatenated with the text tokens, 
\emph{e.g.},~\cite{Li0LZHZWH0WCG20,ChenLYK0G0020,abs-2101-00529,abs-2012-06946,abs-2112-05230}.

Along the direction of scaling on VL tasks, LEMON~\citep{abs-2111-12233}
studies the behavior of the detector-based captioning model with MLM.
CoCa~\citep{yu2022coca} studies different model sizes, but on the same pre-training data.
In this paper, we present
a comprehensive study on 9 various benchmarks (3 in main paper and 6 in supplementary materials, image/video captioning \& QA tasks) with 3 different model sizes and 3 different pre-training data scales (9 data points for each benchmark). 

\section{Generative Image-to-text Transformer} \label{sec:method}

\begin{figure}[t]
    \centering
    \small
    \includegraphics[width=0.99\linewidth]{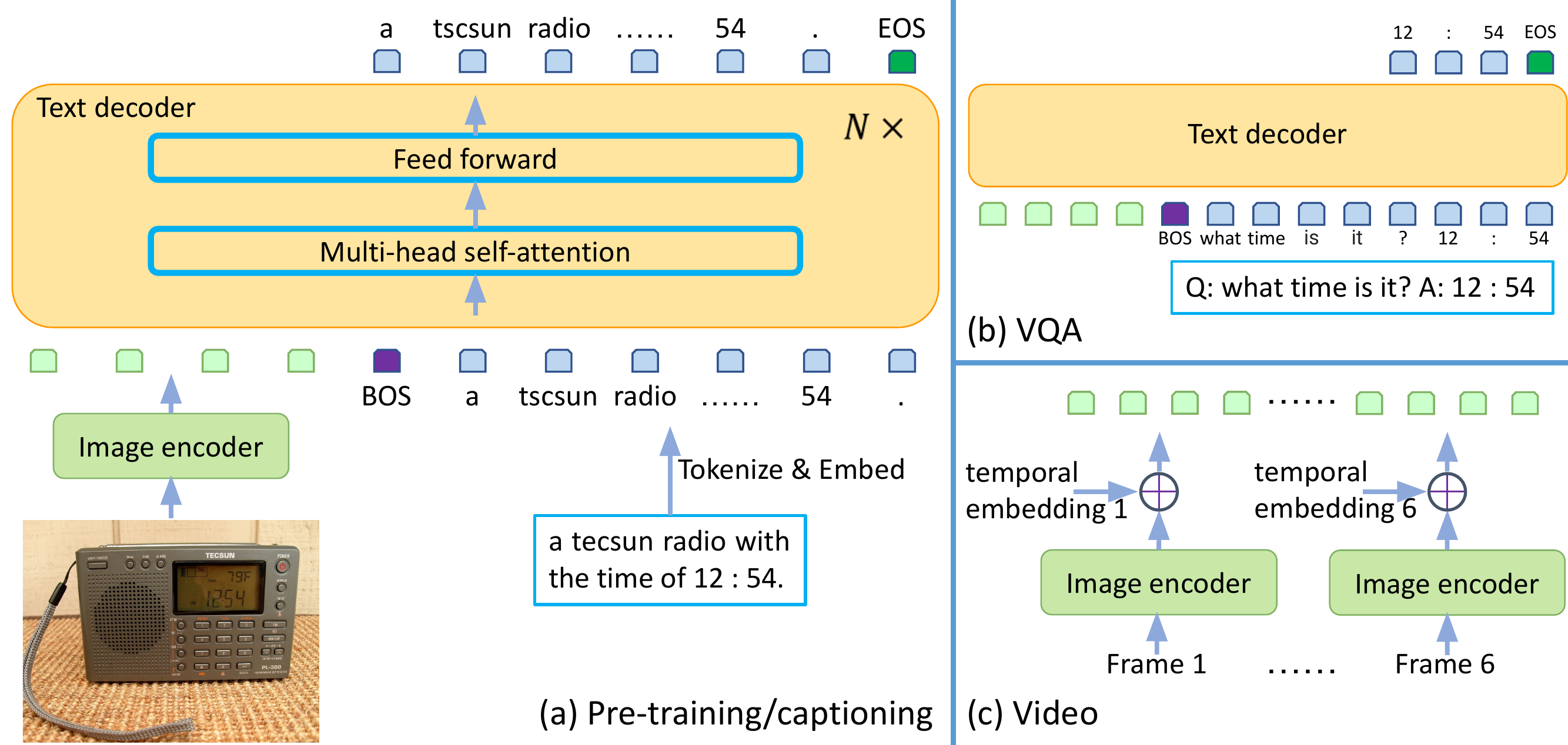}
    \caption{Network architecture of our \modelname, composed of 
    one image encoder
    and one text decoder.  
    (a): The training task in both pre-training and captioning is the language modeling task to predict the associated description.
    (b): In VQA, the question is placed as the text prefix.
    (c): For video, multiple frames are sampled and encoded independently. The features are added with an extra learnable temporal embedding (initialized as 0) 
    before concatenation.}
    \label{fig:framework}
\end{figure}

With large-scale image-text pairs, our goal is
to pre-train a VL model which is simple yet
effective to benefit image/video captioning and QA tasks.
As the input is the image and the output is the text, 
the minimal set of components could be one image encoder 
and one text decoder, which are the only components
of our \modelname as illustrated
in Fig.~\ref{fig:framework}.

\subsection{Network Architecture}
The image encoder is based on the contrastive pre-trained 
model~\citep{abs-2111-11432}.
The input is the raw image and the output is a compact 2D 
feature map, which is flattened into a list of features.
With an extra linear layer and a layernorm layer, the image features
are projected into $D$ dimensions,
which are the input to the text decoder.  
We use the image encoder pre-trained with contrastive tasks because
recent studies show superior performance with such image encoder, 
e.g.~\cite{abs-2111-11432,abs-2111-02387,alayrac2022flamingo}.
In Sec~\ref{sec:analysis} and supplementary materials, 
we also observe the VL performance boosts significantly with a stronger 
image encoder.
This is consistent with the observation in object detection-based approaches, e.g. in~\cite{abs-2012-06946,abs-2101-00529}.
The concurrent work of CoCa~\citep{yu2022coca}
unifies the contrastive task and the generation task.
as one pre-training phase. 
Our approach is equivalent to separating the two tasks sequentially: 
($i$) using the contrastive task to pre-train the image encoder followed by ($ii$) using the generation task
to pre-train both the image encoder and text decoder.

\begin{wrapfigure}{R}{0.3\textwidth}
 \vspace{-2mm}
  \includegraphics[width=0.3\textwidth]{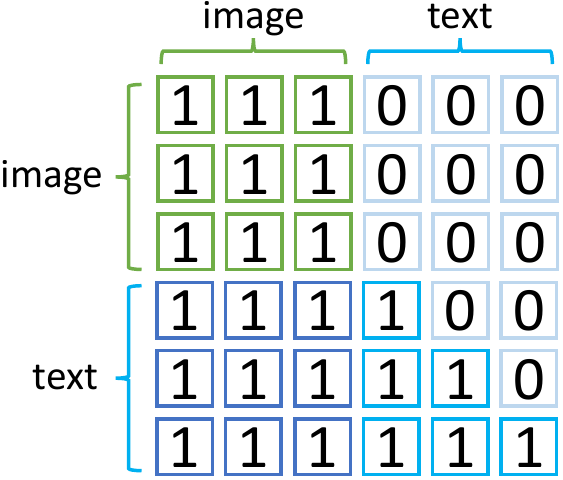}
  \caption{\small
    \texttt{seq2seq} attention mask is applied to the transformer.
    If ($i$, $j$) is 1, the $i$-th output can depend on the $j$-th input; 
    otherwise, not.
  }
  \label{fig:seq}
\end{wrapfigure}

The text decoder is a transformer module to predict the text description. 
The transformer module consists of multiple transformer blocks, 
each of which is composed of one self-attention layer and one feed-forward layer.
The text is tokenized and embedded into $D$ dimensions, followed by an addition
of the positional encoding and a layernorm layer.
The image features are concatenated with the text embeddings as the input
to the transformer module.
The text begins with the \texttt{[BOS]} token, and is decoded in an auto-regressive 
way until the \texttt{[EOS]} token or reaching
the maximum steps.
The \texttt{seq2seq} attention mask as in Fig.~\ref{fig:seq} is applied such that
the text token only depends on the preceding tokens and all image tokens, and 
image tokens can attend to each other. 
This is different from a unidirectional attention mask, where not every image token can rely on all other image tokens.

Instead of well initializing the image encoder, we randomly initialize the 
text decoder. This design choice is highly motivated from the experiment studies of~\cite{abs-2012-06946}, in which the random initialization shows similar 
performance, compared with the BERT initialization.
This could be because the BERT initialization cannot understand the image signal, 
which is critical for VL tasks.
Without dependency of the initialization, we can easily explore different design choices.
The concurrent work of Flamingo~\citep{alayrac2022flamingo}
employs a similar architecture of image encoder + text decoder, but their decoder is pre-trained and frozen
to preserve the generalization capability of the large language model. 
In our \modelname, all parameters are updated to better fit the VL tasks. 

An alternative architecture is the cross-attention-based decoder
to incorporate the image signals instead of concatenation with self-attention.
Empirically as shown in supplementary material (Appendix G.2), with large-scale pre-training, we find the self-attention-based decoder achieves better
performance overall, while in small-scale setting, the cross-attention-based approach wins. 
A plausible explanation is that with sufficient training, the decoder parameters can well process both the image and the text, and the image tokens can be better updated with the self-attention for text generation. With cross-attention, the image tokens cannot attend to each other. 


\subsection{Pre-training}
For each image-text pair,
let $I$ be the image,
$y_i, i \in \{1,\cdots, N\}$ be the text tokens,
$y_0$ be the \texttt{[BOS]} token and $y_{N + 1}$ be the \texttt{[EOS]} token.
We apply the language modeling (LM) loss to train the model.
That is, \begin{align}
    l = \frac{1}{N + 1}\sum_{i = 1}^{N + 1} \text{CE}(y_i, p(y_i | I, \{y_{j}, j=0, \cdots, i - 1)\}),
    \label{eqn:lm_loss}
\end{align}
where CE is the cross-entropy loss with label smoothing of 0.1.

An alternative choice is 
MLM, which predicts typically 15\% of input tokens in each iteration. 
To predict all tokens, we have to run at least $1/0.15=6.7$ epochs.
For LM, each iteration can predict all tokens, which is more efficient for large-scale pre-training data.
In~\cite{abs-2111-12233}, the ablation studies also show that LM can achieve better performance with limited epochs.
In our large-scale training, the number of epoch is only 2 due to computational resource limitation, and thus we choose LM.
Meanwhile, most of the recent large-scale language models are also based on LM, e.g.~\cite{abs-2005-14165,chowdhery2022palm}.

Without the image input, the model is reduced to a 
decoder-only language model, 
similar to GPT3~\citep{abs-2005-14165} in the architecture wise.
Thus, this design also enables the 
possibility to leverage the text-only data to enrich the
decoding capability with a scaled-up decoder. 
We leave this as future work.

\subsection{Fine-tuning}
For the image captioning task, as the training data format 
is the same as that in pre-training, 
we
apply
the same LM task to fine-tune our \modelname.

For visual question answering, 
the question and the ground-truth answer 
are concatenated
as a new special caption during the fine-tuning,
but the LM loss is only applied on the answer and 
the \texttt{[EOS]} tokens.
During inference, the question is interpreted as
the caption prefix and the completed part
is the prediction. 
Compared with the existing approaches~\citep{ufo,wang2021simvlm,abs-2101-00529,abs-2201-12086} for VQAv2~\citep{GoyalKSBP16},
our model is generative without pre-defining the candidate answers, even in inference. 
This imposes
more challenges as the model has to predict at least two correct tokens: one for the answer and another for \texttt{[EOS]}. In contrast, the existing work pre-collects
the answer candidate, recasts the problem as a classification problem, and only needs to predict once.
However, considering the benefit of the free-form answer, we choose the generative approach. 
Due to difficulty of the generative model, we observe slightly worse performance on
VQAv2 than the discriminative existing work.
For the scene-text related VQA tasks, existing approaches~\citep{abs-2012-04638,HuSDR20} typically leverages the OCR
engine to generate the scene text and use dynamic pointer network to decide 
the current output token should be OCR or the general text. 
Here, our approach depends on no OCR engine, and thus no dynamic pointer network.
Empirically, we find the model gradually learns how to read the scene text
with large-scale pre-training, and our model achieves new SoTA performance on these tasks.

Our model is not specifically designed for the video domain, but we find 
our model can also achieve competitive or even new SOTA performance with a simple architecture change. That is, 
we sample multiple frames
from each video clip, and encode each
frame via the image encoder independently.
Afterwards, we add 
a learnable temporal embedding (initialized as zeros),
and concatenate the features from sampled frames.
The final representation is used in a similar way as the image representation for 
captioning and question answering.

We also apply our generation model to the image classification task,
where the class names are interpreted as image captions, and 
our \modelname is fine-tuned to predict the result in an auto-regressive way. 
This 
is different from existing work
which normally
pre-defines the 
vocabulary and uses a linear layer (with softmax)
to predict the likelihood of each category.
This new generation-based scheme is beneficial when new data and new categories are added
to the existing dataset. In this case, the network can continuously 
train on the new data without introducing new parameters.

\section{Experiments}
\subsection{Setting}
We collect 0.8B image-text pairs for pre-training, which 
include
COCO~\citep{LinMBHPRDZ14},
Conceptual Captions (CC3M)~\citep{SoricutDSG18}, 
SBU~\citep{OrdonezKB11},
Visual Genome (VG)~\citep{KrishnaZGJHKCKL16},
Conceptual Captions (CC12M)~\citep{changpinyo2021cc12m},
ALT200M~\citep{abs-2111-12233},
and an extra 
0.6B 
data following a similar collection procedure in~\cite{abs-2111-12233}.
The image encoder is initialized from the pre-trained contrastive model~\citep{abs-2111-11432}.
The hidden dimension ($D$) is 768. 
The text decoder consists of 6 randomly-initialized 
transformer blocks.
The total number of model parameters is 0.7 billion.
The learning rates of the image encoder
and the decoder
are $1e^{-5}$ and $5e^{-5}$, respectively,
and follow the cosine decay to 0.
The total number of epochs is $2$. 
During inference, the beam size is $4$ and the 
length penalty~\citep{WuSCLNMKCGMKSJL16} is 0.6 by default.

Supplementary materials show results on two smaller model variants (\modelnamewospace$_B$ and \modelnamewospace$_L$) 
and one even larger model (\modelnamewospace2) with full details.
When comparing with existing approaches, the reference numbers are the 
best one reported in the corresponding paper unless explicitly specified.

\begin{table}[t]
\small
\centering
\caption{Results on image captioning. 
*: the nubmers are from~\cite{abs-2003-12462}; 
CE: cross-entropy optimization. 
All numbers are CIDEr scores, and other metrics are shown in supplementary materials. 
$^\#$: winner entry of the CVPR 2021 workshop challenge
Anc.-Cap.: ~\cite{abs-2105-03236}  
AoANet: ~\cite{huang2019attention}
BUTD: ~\cite{00010BT0GZ18},   
CoCa: ~\cite{yu2022coca},                 
DistillVLM: ~\cite{FangW0WY021},
Flamingo: ~\cite{alayrac2022flamingo},
Human: ~\cite{abs-1812-08658},
LEMON: ~\cite{abs-2111-12233},     
M4C-Cap.: ~\cite{HuSDR20} 
MiniVLM: \cite{abs-2012-06946},
MTMA: ~\cite{GongZWCZSL21},
OFA: ~\cite{abs-2202-03052},              
OSCAR: ~\cite{Li0LZHZWH0WCG20},
UFO: ~\cite{ufo},           
UniversalCap: ~\citep{abs-2111-12727} 
ViTCap: ~\cite{abs-2112-05230},
VinVL: ~\cite{abs-2101-00529},          
VIVO: ~\cite{abs-2009-13682}         
SimVLM: ~\cite{wang2021simvlm},           
TAP: ~\cite{abs-2012-04638}.
}
\label{tab:caption}
\begin{tabular}{@{}p{4.0cm}@{}p{3.5cm}@{}p{3.5cm}@{}p{3.2cm}@{}}  
    \vtop{\vskip 0pt \vskip -\ht\strutbox 
                    \begin{tabular}{c}
                	\begin{tabular}{@{}lc@{}}
                		\toprule
                                         Method                     & CE       \\ 
                		                                            \midrule            
                		MiniVLM               & 119.8            \\
                 		DistillVLM               & 120.8            \\
                		ViTCap                & 125.2            \\
                		OSCAR                & 127.8            \\
                		VinVL                 & 130.8            \\
                		UFO                              & 131.2            \\ 
                		Flamingo         & 138.1            \\
                LEMON                         & 139.1            \\
                		SimVLM                & 143.3            \\
                		CoCa                      & 143.6            \\
                		OFA                   & \bf145.3 \\
                		\midrule  
                		\modelname                                  & 144.8  \\ 
                        \bottomrule
                	\end{tabular}
                	\\
                	(a) COCO Karp.
                	 \end{tabular}
    \vskip -\dp\strutbox }%
                    & 
    \vtop{\vskip 0pt \vskip -\ht\strutbox 
                	\begin{tabular}{c}
                	    \begin{tabular}{l@{~~~~~~}c}
                        \toprule
                        Method                          & C               \\
                                  \midrule
                        BUTD        & 120.5 \\
                        VinVL       & 138.7 \\
                        \midrule
                    \modelname                          & \textbf{148.8} \\ 
                        \bottomrule
                        \end{tabular}     
                        \\
                        (b) COCO test (c40) 
                        \\
                        \\
                        \begin{tabular}{l@{~~~~}c}
                        \toprule
                         Method                  & test-std \\
                        \midrule    
                        MTMA & 94.1 \\ \midrule
                        \modelname                   & \bf114.4 \\ 
                        \bottomrule
                        \end{tabular}
                        \\
                        (c) VizWiz-Captions
                	\end{tabular}
        \vskip -\dp\strutbox }%
                    &
    \vtop{\vskip 0pt \vskip -\ht\strutbox 
                    \begin{tabular}{c}
                    \begin{tabular}{@{~}l@{~}c}
                         \toprule
                		Method                              & Test \\ 
                		\midrule                            
                		OSCAR        & 80.9      \\
                		Human         & 85.3      \\
                		VIVO          & 86.6      \\
                		VinVL         & 92.5      \\
                		UFO                      & 92.3      \\ 
                		SimVLM        & 115.2     \\
                		LEMON         & 114.3      \\
                		UniversalCap  & 119.3      \\
                		CoCa              & 120.6      \\
                		\midrule
                	    \modelname                          & \bf123.4  \\ 
                		\bottomrule
                	\end{tabular}
                	\\
                	(d) nocaps
                   \end{tabular}
        \vskip -\dp\strutbox }%
                	&
    \vtop{\vskip 0pt \vskip -\ht\strutbox 
                	\begin{tabular}{c}
                    \begin{tabular}{l@{}c}
                    \toprule
                     Method                         &  Test \\
                     \midrule
                    BUTD*       & 33.8   \\
                  AoANet* & 34.6   \\
                  M4C-Cap.*          & 81.0   \\
                  Anc.-Cap.   & 87.4 \\
                  TAP         & 103.2  \\
                  TAP$^\#$    & 109.7  \\
                  Human       & 125.5  \\   \midrule
                \modelname                          & \bf138.2 \\ 
                    \bottomrule
                    \end{tabular}
                    \\
                    (e) TextCaps
                \end{tabular} 
        \vskip -\dp\strutbox }%
    \\  
\end{tabular} 
\end{table}

\begin{table}[h]
    \centering
    \caption{
    Zero/Few/Full-shot evaluation on Flickr30K with Karpathy split.
    }
    \label{tab:flickr_few}
    \begin{tabular}{cccccc}
    \toprule
    Shot       &  0   & 16   & 32   & 290 (1\%) & full \\
    \midrule
\cite{ZhouPZHCG20} & - & -   & -    & -         & 68.5  \\
    Flamingo   & 67.2 & 78.9 & 75.4 &  -        &    -  \\
\modelname     & 49.6 & 78.0 & 80.5 &  86.6     & 98.5 \\
\bottomrule
    \end{tabular}
\end{table}

\subsection{Results on Image Captioning and Question Answering}
We comprehensively evaluate the captioning performance on 
the widely-used 
Karpathy split~\citep{KarpathyL15} of COCO~\citep{LinMBHPRDZ14} and Flickr30K~\citep{young2014image},
the COCO test set,
nocaps~\citep{abs-1812-08658}\footnote{
We compare all approaches including using external image-text datasets.}
which focuses on novel objects,
TextCaps~\citep{abs-2003-12462} which focuses on scene-text understanding,
and VizWiz-Captions~\citep{abs-2002-08565} which focuses on the real use case by the vision-impaired people.
The results in CIDEr~\citep{VedantamZP15} are shown in Table~\ref{tab:caption} and~\ref{tab:flickr_few}.
From the results, we can see our model achieves the new SOTA performance on all 
these metrics except on COCO Karpathy test. 
On nocaps, compared with CoCa~\citep{yu2022coca}, 
our model is much smaller in the model size (0.7B vs 2.1B), but achieves higher performance (123.0 vs 120.6 in CIDEr).
On Textcaps, our solution outperforms the previous SOTA (TAP~\cite{abs-2012-04638}) 
by a breakthrough margin (28.5 points in CIDEr), and also surpasses the human performance for the first time.
{
For zero/few-shot evaluation as shown in Table~\ref{tab:flickr_few}, our model can significantly benefit from more shots.
With 32-shots, our approach is also better than Flamingo.
}

On VQA, the evaluation benchmarks include VQAv2~\citep{GoyalKSBP16}, TextVQA~\citep{singh2019towards}, 
VizWiz-VQA~\citep{Gurari0SGLGLB18}.
ST-VQA~\citep{biten2019scene}, and OCR-VQA~\citep{mishra2019ocr}.
Before fine-tuning the model, we run an intermediate fine-tuning on the combination of the training data of
VQAv2, TextVQA, ST-VQA, OCR-VQA, VizWiz-VQA, 
Visual Genome QA~\citep{KrishnaZGJHKCKL16}, 
GQA~\citep{HudsonM19},
and 
OK-VQA~\citep{marino2019ok}.
To avoid data contamination, 
we remove the duplicate images of the test and validation set of the target benchmarks.
As illustrated in Table~\ref{tab:vqa},
we achieve new SOTA on VizWiz-VQA and OCR-VQA, and same performance with prior SOTA of LaTr~\citep{biten2022latr}
on ST-VQA. 
Compared with the concurrent work of Flamingo~\citep{alayrac2022flamingo},
we achieve higher accuracy (+5.4) on TextVQA and lower (-3.29) on VQAv2. 
Note that Flamingo's model size is 80B, which is 114 times of ours (0.7B).
On VQAv2, we observe that our model performs worse in 1.5 points than 
the discriminative model of Florence~\citep{abs-2111-11432}, which shares the same image encoder. 
The reason might be the 
increased difficulty
of the generative model. That is, each correct answer requires at least two correct predictions (answer and \texttt{[EOS]}; 2.2 on average), 
while the discriminative model requires only one correct prediction. 
In~\citep{wang2021simvlm},
the ablation study also shows the better performance by around 1 point than the discriminative counterpart.
Another reason could be that the model of Florence for VQA leverages RoBerta~\citep{liu2019roberta} as the text encoder, which implicitly uses the text-only data to improve the performance.

\begin{table}[t]
    \small
    \caption{Results on visual question answering. 
    (a): for VQAv2, approaches are divided according to whether the answer vocabulary is pre-defined (Closed) or not (Open) during inference. The model with closed vocabulary can be 
        a classification model or generation model with constrained outputs, \emph{e.g.},~\cite{abs-2202-03052,abs-2201-12086}.
        The two numbers in parenthesis are the number of parameters and the number of images (the images for pre-trained modules are not counted) in VL pretraining.
    (b): for TextVQA, Mia~\citep{abs-2106-15332}$^\#$ is the winner entry of TextVQA Challenge 2021 with a fine-tuned  T5-3B~\citep{RaffelSRLNMZLL20} model.
    (c): $^{\#\#}$: winner entry of 2021 VizWiz Grand Challenge Workshop.
ALBEF: ~\cite{li2021align},
BLIP: ~\cite{abs-2201-12086},
BLOCK+CNN+W2V: ~\cite{mishra2019ocr},
CLIP-ViL: ~\cite{abs-2107-06383},
CoCa: ~\cite{yu2022coca},
CRN: ~\cite{LiuXWDJT20},
Flamingo: ~\cite{alayrac2022flamingo},
Florence: ~\cite{abs-2111-11432},
LaAP-Net: ~\cite{HanHH20},
LaTr: ~\cite{biten2022latr},
M4C: ~\cite{HuSDR20},
M4C: ~\cite{HuSDR20},
METER: ~\cite{abs-2111-02387},
Mia: ~\cite{abs-2106-15332},
mPlug: ~\cite{li2022mplug},
OSCAR: ~\citep{Li0LZHZWH0WCG20},
OFA: ~\cite{abs-2202-03052},
UFO: ~\cite{ufo},
UNITER: ~\citep{ChenLYK0G0020},
UNIMO: ~\cite{li2020unimo},
SA-M4C: ~\cite{KantBASPLA20},
SimVLM: ~\cite{wang2021simvlm},
SMA~\cite{abs-2006-00753},
SMA: ~\cite{abs-2006-00753},
TAP: ~\cite{abs-2012-04638},
VinVL: ~\cite{abs-2101-00529},
VILLA: ~\cite{Gan0LZ0020}.
}
    \label{tab:vqa}
    \centering
    \begin{tabular}{@{}p{6cm}p{4cm}p{4.5cm}@{}}
    \vtop{\vskip 0pt \vskip -\ht\strutbox 
           \begin{tabular}{c}
                    \begin{tabular}{@{}l@{}l@{}c@{}} \toprule
                        Vocabulary                             &  Method                                  & test-std \\ \midrule
                         \multirow{15}{*}{Closed}              & OSCAR            & 73.82 \\
                                                               & UNITER             & 74.02 \\
                                                               & VILLA                 & 74.87 \\
                                                               & UNIMO                & 75.27 \\
                                                               & ALBEF                & 76.04 \\
                                                               & VinVL             & 76.60 \\
                                                               & UFO                          & 76.76 \\
                                                               & CLIP-ViL          & 76.70 \\
                                                               & METER             & 77.64 \\
                                                               & BLIP              & 78.32 \\
                                                               & SimVLM (-, 1.8B)  & 80.34 \\
                                                               & Florence (0.9B, 14M)    & 80.36 \\ 
                                                               & mPlug (0.6B, 14M)    & 81.26 \\
                                                               & OFA (0.9B, 54M)   & 82.0  \\ 
                                                               & CoCa (2.1B, 4.8B)     & \bf82.3  \\
                                                               \midrule
                         \multirow{2}{*}{Open}                 & Flamingo (80B, 2.3B)     & \bf82.1  \\
                                                               & \modelname (0.7B, 0.8B)                             &   78.81    \\ 
                        \bottomrule
                    \end{tabular}
              \\
              (a) VQAv2
        \end{tabular}
    \vskip -\dp\strutbox }%
    &
    \vtop{\vskip 0pt \vskip -\ht\strutbox 
        \begin{tabular}{@{}c@{}}
                    \begin{tabular}{@{}l@{~~}c@{}}
                    \toprule
                    Method                          & test       \\
                    \midrule
                    M4C             & 40.46      \\
                    LaAP-Net        & 41.41      \\
                    SA-M4C     & 44.6         \\
                    SMA      & 45.51    \\
                    TAP      & 53.97     \\
                Flamingo & 54.1      \\
                    Mia & \bf73.67       \\
                    \midrule
                    \modelname                     & 59.75     \\ 
                    \bottomrule
                    \end{tabular}
            \\
            (b) TextVQA
             \\  
             \\
                        \begin{tabular}{@{}l@{~~~~}c@{}} \toprule
                            Method                             & test \\ \midrule
                            \citep{vizwiz2021winner}$^{\#\#}$  & 60.6 \\
                        Flamingo   &  65.4 \\
                        \midrule
                            \modelname                        & \bf67.5 \\ 
                            \bottomrule
                        \end{tabular}
                        \\
                        (c) VizWiz-QA
        \end{tabular}
    \vskip -\dp\strutbox }%
     &
    \vtop{\vskip 0pt \vskip -\ht\strutbox 
        \begin{tabular}{cc}
            \begin{tabular}{@{}l@{~~~~~~~~~~~~}c@{}} \toprule
                Method                            & Test ANLS\\
                \hline
                M4C               & 46.2 \\
                SMA        & 46.6 \\
                CRN            & 48.3 \\
                LaAP-Net          & 48.5 \\
                SA-M4C       & 50.4 \\
                TAP        & 59.7 \\     
                LaTr        & \bf69.6 \\
                \midrule
                \modelname                       & \bf69.6     \\  
                \bottomrule
            \end{tabular} \\
            (d) ST-VQA
            \\
            \begin{tabular}{@{}l@{~~}c@{~~}c@{}} \toprule
                Method                            & test \\ \midrule
             BLOCK+CNN+W2V  & 48.3 \\
                M4C               & 63.9 \\
                LaAP-Net          & 64.1 \\
                LaTr        & 67.9 \\
                \midrule
                \modelname                       & \bf68.1 \\  
                \bottomrule 
            \end{tabular}
        \\
        (e) OCR-VQA
        \end{tabular}
    \vskip -\dp\strutbox }%
    \end{tabular}
\end{table}

\subsection{Results on Video Captioning and Question Answering}

On the video captioning task, the performance is evaluated on 
MSVD~\citep{chen-dolan-2011-collecting} with the widely-used splits from~\cite{VenugopalanXDRMS14},
MSRVTT~\citep{xu2016msr},
YouCook2~\citep{zhou2018towards} (results in supplementary materials.)
VATEX~\citep{wang2019vatex},
and
TVC~\citep{lei2020tvr} (results in supplementary materials.).
On VATEX, the performance is evaluated on both the public test and private test (evaluated on the server).
Video QA is evaluated on 
MSVD-QA~\citep{xu2017video,chen-dolan-2011-collecting}, 
MSRVTT-QA~\citep{xu2017video,xu2016msr},
and TGIF-Frame~\citep{JangSYKK17},
which are all open-ended tasks.
The results are shown in Table~\ref{tab:video_caption} and Table~\ref{tab:video_qa}
for captioning and QA, respectively.
Although our model is not dedicated for video tasks, our model achieve new SOTA 
on MSRVD, MSRVTT, and VATEX for captioning and on MSVD-QA and TGIF-Frame for QA. 
For example on VATEX private test, our results are even better (93.8 vs 86.5) than 
CLIP4Caption++~\citep{abs-2110-05204}, which relies on model ensemble and additional subtitle input.
This is also better than Flamingo~\citep{alayrac2022flamingo} (84.2) with 80B parameters.

\begin{table}[t!]
\centering
\small
\caption{Results on video captioning. 
$^E$: model ensemble; 
$^T$: with the subtitle as additional input.
C.4Cap.: ~\cite{abs-2110-05204}
GRU-EVE: ~\cite{aafaq2019spatio}
MGSA: ~\cite{chen2019motion}
MGSA: ~\cite{chen2019motion}
MV-GPT: ~\cite{abs-2201-08264}
PickNet: ~\cite{chen2018less}
PMI-CAP: ~\cite{chen2020learning}
SibNet: ~\cite{liu2020sibnet}
OA-BTG: ~\cite{zhang2019object}
ORG-TRL: ~\cite{Zhang_2020_CVPR}
OpenBook: ~\cite{zhang2021open}
POS+VCT: ~\cite{hou2019joint}
POS+CG: ~\cite{wang2019controllable}
SAAT: ~\cite{zheng2020syntax},
STG-KD: ~\cite{pan2020spatio}
SwinBERT: ~\cite{abs-2111-13196}
Support-set: ~\cite{patrick2020support}
VaTeX: ~\cite{wang2019vatex}
VALUE: ~\cite{VALUE}
}
\label{tab:video_caption}
\begin{tabular}{@{}p{5cm}p{5cm}p{4cm}@{}}
    \vtop{\vskip 0pt \vskip -\ht\strutbox 
    \begin{tabular}{c}
        \begin{tabular}{lcc} \toprule
            Method                              &B@4      & C     \\ \midrule
            PickNet         & 52.3    & 76.5  \\
            GRU-EVE      & 47.9    & 78.1  \\
            SAAT         & 46.5    & 81.0  \\
            MGSA          & 53.4    & 86.7  \\
            POS+VCT         & 52.8    & 87.8  \\
            SibNet         & 54.2    & 88.2  \\
            POS+CG  & 52.5    & 88.7  \\
            OA-BTG       & 56.9    & 90.6  \\
            STG-KD         & 52.2    & 93.0  \\
            PMI-CAP     & 54.6    & 95.1  \\
            ORG-TRL      & 54.3    & 95.2  \\
            SwinBERT      & 58.2    & 120.6 \\ \midrule
            \modelname                          & \bf79.5 & \bf 180.2 \\  
            \bottomrule
        \end{tabular}
        \\
        (a) MSVD
    \end{tabular}
    \vskip -\dp\strutbox }%
    &
    \vtop{\vskip 0pt \vskip -\ht\strutbox
    \begin{tabular}{c}
        \begin{tabular}{lcc} \toprule
             Method                              &B@4      & C     \\ \midrule
            SAAT          & 39.9    & 51.0 \\
            MGSA           & 42.4    & 47.5 \\
            POS+VCT          & 42.3    & 49.1 \\
            SibNet          & 40.9    & 47.5 \\
            POS+CG   & 42.0    & 48.7 \\
            OA-BTG        & 41.4    & 46.9 \\
            STG-KD          & 40.5    & 47.1 \\
            Support-set   & 38.9    & 48.6 \\
            PMI-CAP      & 42.1    & 49.4 \\
            ORG-TRL       & 43.6    & 50.9 \\
            OpenBook        & 33.9    & 52.9 \\
            SwinBERT       & 41.9    & 53.8 \\ 
            MV-GPT$^T$     & 48.9    & 60   \\
            \midrule
            \modelname                           & \bf53.8 & \bf73.9 \\  
            \bottomrule
        \end{tabular} \\
        (b) MSRVTT
    \end{tabular}
    \vskip -\dp\strutbox }%
    &
        \vtop{\vskip 0pt \vskip -\ht\strutbox
    \begin{tabular}{c}
        \begin{tabular}{@{}lc@{}} 
            \toprule
                     Method                              & C \\ \midrule
                    VaTeX          & 45.1 \\
                 OpenBook          & 57.5 \\
                VALUE$^{T}$                & 58.1 \\
                SwinBERT          & 73.0 \\ 
                C.4Cap.$^{ET}$    & 85.7  \\
                \midrule
                \modelname                              & \bf91.5  \\ 
                \bottomrule
        \end{tabular} 
        \\
        (a) VATEX public test
        \\
        \\
        \begin{tabular}{@{}lc@{}} \toprule
                     Method                              & C \\ \midrule
  X-L.+T.$^E$                        & 81.4 \\
                     Flamingo & 84.2 \\
            C.4Cap.$^{ET}$            &  86.5     \\
                \midrule
                \modelname                              & \bf93.8  \\ 
                \bottomrule
        \end{tabular} \\
        (e) VATEX private test 
    \end{tabular}
    \vskip -\dp\strutbox }
\end{tabular}
\end{table}

\begin{table}[]
    \small
    \centering
    \caption{Results on video question answering. All are open-ended question answering tasks. 
    All-in-one: ~\cite{wang2022all},
    ClipBERT: \cite{LeiLZGBB021},
    CoMVT: ~\cite{SeoNS21},
    Flamingo: ~\cite{alayrac2022flamingo},
    JustAsk: ~\cite{YangMSLS21},
    MERLOT: \cite{ZellersLHYPCFC21},
    MV-GPT: ~\cite{abs-2201-08264},
    QueST: ~\cite{JiangCLZG20},
    HCRN: ~\cite{LeLVT21},
    VIOLET: ~\cite{abs-2111-12681}.
    }
    \label{tab:video_qa}
    \begin{tabular}{ccc}
                \begin{tabular}{lc}
                \toprule
                Method          & Accuracy \\
                            \midrule
            QueST        & 34.6 \\
            HCRN             & 36.1 \\
            CoMVT            & 42.6  \\
            JustAsk       & 46.3  \\
            VIOLET    &    47.9 \\
            All-in-one   & 48.3 \\
                \midrule
                \modelname                  & \bf56.8 \\ 
                \bottomrule
                \end{tabular}
         & 
                \begin{tabular}{lc}
                \toprule
                Method          & Accuracy \\
                            \midrule
            JustAsk       & 41.5  \\
            MV-GPT    & 41.7 \\
            MERLOT   & 43.1 \\
            VIOLET    & 43.9 \\
         All-in-one      & 46.8 \\
        Flamingo & \bf47.4 \\
                \midrule
                \modelname                  & 43.2 \\ 
                \bottomrule
                \end{tabular}
         &
            \begin{tabular}{lc}
            \toprule
            Method          &  Accuracy \\
            \midrule
        HCRN             & 55.9 \\
        QueST        & 59.7 \\
        ClipBERT      &   60.3 \\
           All-in-one     & 66.3 \\
        VIOLET    &    68.9 \\
        MERLOT   & 69.5 \\
            \midrule
            \modelname      &  \bf72.8 \\ 
            \bottomrule
            \end{tabular}
            \\
            (a) MSVD-QA & (b) MSRVTT-QA & (c) TGIF-Frame 
    \end{tabular}
\end{table}

\subsection{Results on Image Classification}
We fine-tune \modelname on ImageNet-1k. 
Each category is mapped to a unique 
class name, and the prediction is correct only if it is exactly matched 
with the ground-truth label subject to more or fewer whitespaces\footnote{pred.replace(` ', `') == gt.replace(` ', `')}. 
As shown in Table~\ref{tab:main_imagenet}, our approach can achieve descent
accuracy without pre-defining the vocabulary.
Compared with Florence~\citep{abs-2111-11432} (same image encoder), our approach is worse in about 1.2 points.
The reason might be similar to the case on VQAv2. That is, the generative approach needs to 
predict more tokens correctly to make one correct prediction,
which increases the difficulty. 

\begin{table}[t]
    \centering
    \begin{tabular}{cc}
        \begin{minipage}[t]{0.45\linewidth}
        \centering
        \small
         \caption{Results on ImageNet-1k classification task.
         Our approach takes the class name as the caption and predict 
         the label in an auto-regressive way without pre-defining the vocabulary.}
         \label{tab:main_imagenet}
        \begin{tabular}{@{}l@{}l@{}c@{}}
        \toprule
        Vocabulary          &  Method                        & Top-1 \\
        \midrule
    \multirow{2}{*}{Closed} & ALIGN~\citep{JiaYXCPPLSLD21}    & 88.64 \\
                            & Florence~\citep{abs-2111-11432} & 90.05 \\
                            & CoCa~\citep{yu2022coca}         & \bf91.0  \\
        \midrule
        Open                & \modelname                     & 88.79 \\ 
        \bottomrule
        \end{tabular}      
        \end{minipage}
         &
    \begin{minipage}[t]{0.45\linewidth}
         \centering
         \small
         \caption{Results on scene text recognition. MJ and ST indicate the MJSynth (MJ)~\citep{jaderberg2014synthetic,jaderberg2016reading} and SynthText (ST)~\citep{gupta2016synthetic} datasets used for training scene text recognition models.}
         \label{tab:scene_text_rec_avg}
         \begin{tabular}{l @{~~} c@{~~}  c@{~~}c@{~~}c@{~~}c@{~~}c@{~~}c@{~~}  c}
         \toprule
        Method & FT data &  Average \\
        \midrule
         SAM~\citep{liao2019mask}                     & MJ+ST     & 87.8 \\
         Ro.Scanner~\citep{yue2020robustscanner}      & MJ+ST     & 87.5 \\
         SRN~\citep{yu2020towards}                    & MJ+ST     & 89.6 \\
         ABINet~\citep{fang2021read}                  & MJ+ST     & 91.9 \\
         S-GTR~\citep{he2021visual}                   & MJ+ST     & 91.9 \\
         MaskOCR~\citep{lyu2022maskocr}               & MJ+ST     & \bf93.8 \\
         \midrule
         \multirow{2}{*}{\modelname}                 & TextCaps  & 89.9 \\
                                                     & MJ+ST     & 92.9 \\
         \bottomrule
         \end{tabular}
        \end{minipage}
    \end{tabular}
\end{table}

{
\textbf{Zero-shot/Few-shot.}
The result is shown in Table~\ref{tab:imagenet_few_shot}.
With no knowledge of the vocabulary, the pretrained \modelname cannot infer the expected vocabulary,
and thus the exactly-match accuracy is only 1.93\% (in the column of \textit{equal}).
However, if we relax the requirement and take it correct if the prediction contains the ground-truth,
the accuracy is 40.88\% (in the column of \textit{in}), which shows the predicted caption can well identify the image content.
If we have the vocabulary as a prior and limit the output tokens to be within the vocabulary, 
the accuracy
drops to 33.48\% (in the column of \textit{voc-prior}). 
This may suggest the network is less natural to directly predict the category name.
By fine-tuning the model with only 1 shot or 5 shots per category, we observe that 
the accuracy is significantly 
improved. 
This demonstrates our model can be easily adapted to 
downstream tasks even with a few training samples.
With the shot increased from 1 to 5, the gap between \textit{voc-prior} and the other two columns (\textit{equal} and \textit{in}) becomes smaller. This is expected as more shots can be better to
guide the network to predict in-vocabulary output.

Compared with Flamingo, our \modelname achieves higher accuracy.
Flamingo conducts the few-shot learning without parameter update, but each test image is combined 
with the support training examples as extra network inputs.
Meanwhile, different test image requires different support shots 
based on~\cite{yang2022empirical}.
These may increase the inference cost.
In contrast, our model updates the parameters by a lightweight fine-tuning once, and then all these training shots are not required during inference.

\begin{table}[]
    \centering
    \caption{
    Zero/Few-shot evaluation on ImageNet with 3 metrics. 
    \textit{equal}: the unrestricted prediction should be exactly matched to the ground-truth.
    \textit{in}: the unrestricted prediction should contain the ground-truth label name.
    \textit{voc-prior}: the vocabulary is pre-defined as a prior. For our \modelname, a trie structure is constructed motivated from \cite{abs-2202-03052} 
    to limit the candidate tokens during each token prediction, such that the predicted result is guaranteed to be within the vocabulary. 
    }
    \label{tab:imagenet_few_shot}
    \begin{tabular}{cccccccccc}
    \toprule
                 & \multicolumn{3}{c}{Zero-shot}    & \multicolumn{3}{c}{1-shot per class}  & \multicolumn{3}{c}{5-shot per class} \\
                 \cmidrule(lr){2-4}   \cmidrule(lr){5-7} \cmidrule(lr){8-10}
Accuracy type    & equal & in     & voc-prior       & equal & in    & voc-prior   & equal & in    & voc-prior   \\
\midrule          
Flamingo         &   -   & -      &  -              & -     & -     & 71.7        & -     & -     & 77.3        \\
\midrule
\modelname       & 1.93  & 40.88  & 33.48           & 64.54 & 66.76 & 72.45       & 79.79 & 80.15 & 80.95  \\
\bottomrule
    \end{tabular}
\end{table}

}

\subsection{Results on Scene Text Recognition}

The task~\citep{graves2006connectionist} aims to read scene text directly from the image.
We evaluate our model in two settings. One is the \modelname fine-tuned on TextCaps.
The prediction is considered correct if the caption contains the ground-truth scene text word. The other is to fine-tune the model on two large scene text datasets: MJSynth (MJ)~\citep{jaderberg2014synthetic,jaderberg2016reading} and SynthText (ST)~\citep{gupta2016synthetic}, where the ground-truth scene text is used as the \textit{caption}.
The prediction is correct if the output is the exact match to the ground-truth.
Following the established setup, we evaluate on six standard benchmarks,
including ICDAR 2013 (IC13)~\citep{karatzas2013icdar}, ICDAR 2015 (IC15)~\citep{karatzas2015icdar}, IIIT 5K-Words (IIIT)~\citep{mishra2012scene}, Street View Text (SVT)~\citep{wang2011end}, Street View Text-Perspective (SVTP)~\citep{phan2013recognizing}, and CUTE80 (CUTE)~\citep{risnumawan2014robust}.
The average accuracy is reported in Table~\ref{tab:scene_text_rec_avg}. The accuracy on individual test sets is in supplementary materials.
Our TextCaps-fine-tuned captioning model achieves an 89.9 accuracy,
which demonstrates the strong scene text comprehension capability of our captioning model. 
After fine-tuning the model on the standard MJ+ST datasets, \modelname achieves 92.9 that surpasses the prior arts~\citep{fang2021read,he2021visual} of 91.9. 
\subsection{Analysis}\label{sec:analysis}
\begin{figure}
    \centering
    \small
    \begin{tabular}{@{}c@{}c@{}c@{}}
    \includegraphics[width=0.33\linewidth]{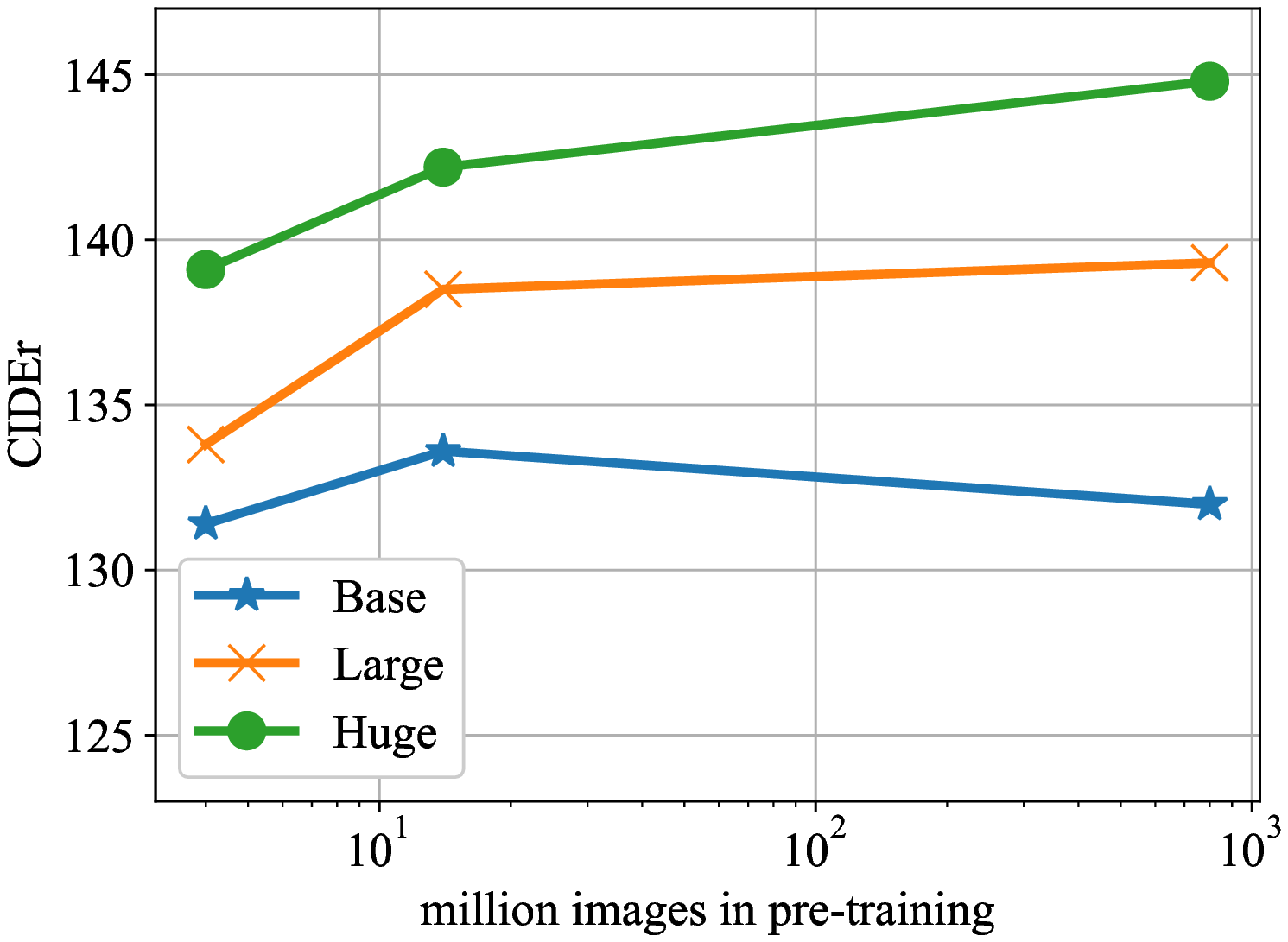}         &      
    \includegraphics[width=0.33\linewidth]{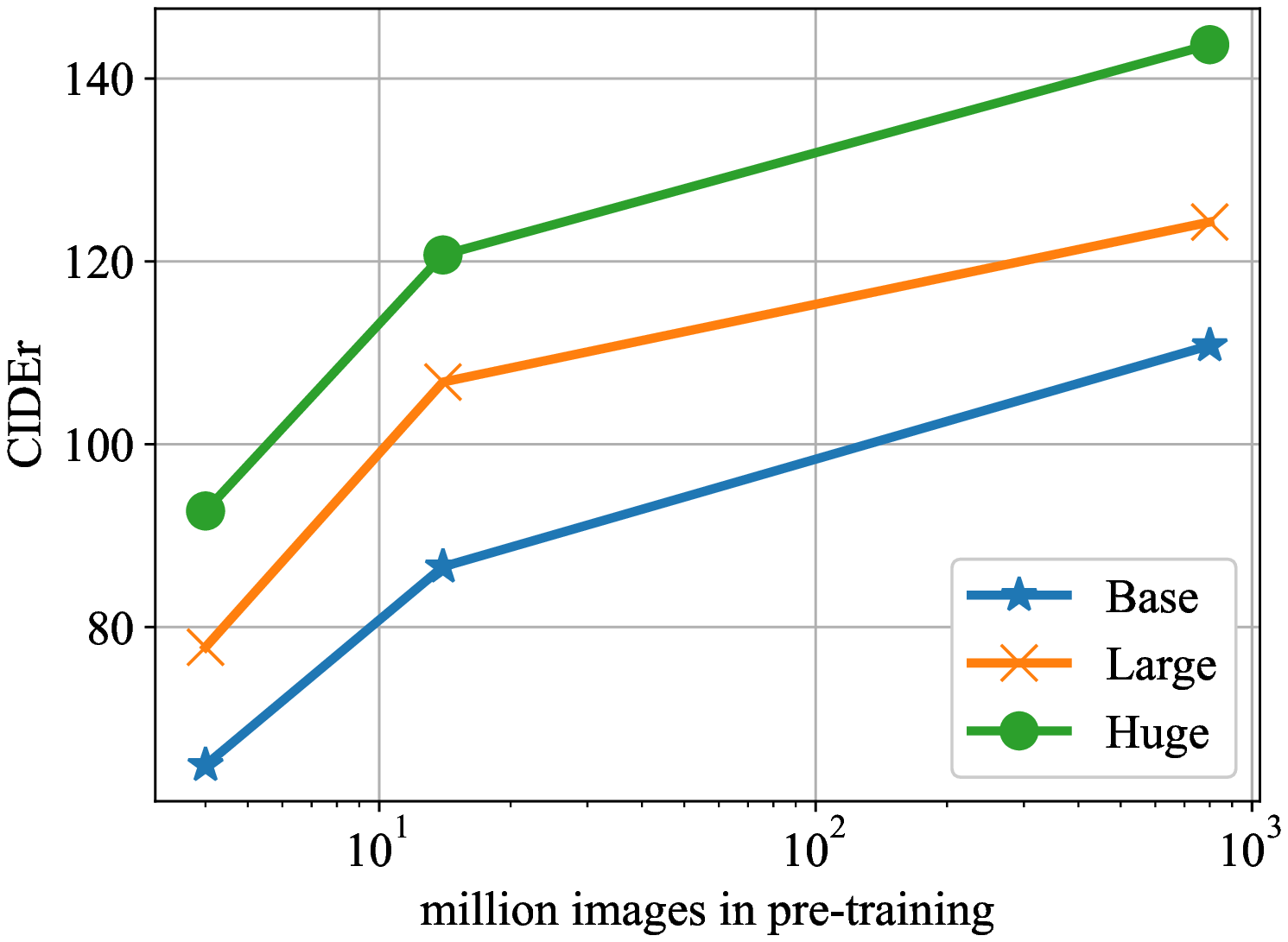}     & 
    \includegraphics[width=0.33\linewidth]{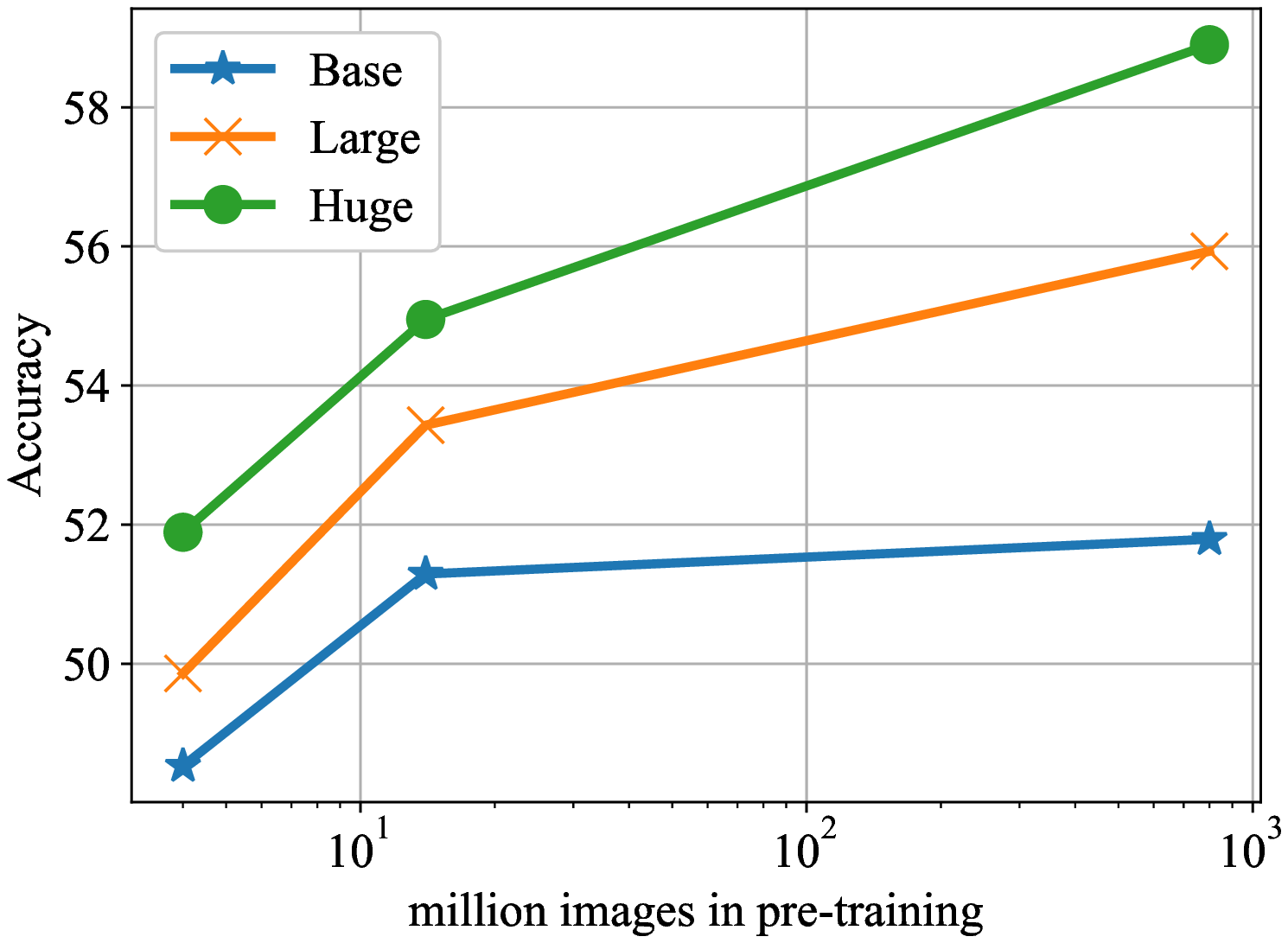}     \\
            (a) COCO    &   (b) TextCaps  & (c) VizWiz-QA
    \end{tabular}
    \caption{
    Performance with different pre-training data scales 
    and different model sizes. 
    }
    \label{fig:scale}
\end{figure}

\noindent\textbf{Model and data scaling.}
To study the trending with data scales, 
we construct two smaller pre-training datasets: one is the combination of COCO, SBU, CC3M and VG, leading to 4M images or 10M image-text pairs;
the other is to further combine CC12M, leading to about 14M images or 20M image-text pairs.
When pre-training on small-scale datasets, we use 30 epochs rather than 2 epochs as on the 0.8B data.
For the network structure, we name our model as \textit{Huge} and
replace the image encoder
with ViT-B/16 and ViT-L/14 from CLIP~\cite{clip} as \textit{Base} and \textit{Large}, respectively. 
Fig.~\ref{fig:scale} shows the results on COCO, TextCaps, and VizWiz-QA.
On COCO, the base model benefits from 4M to 14M, but the performance drops with 0.8B data.
The 14M data are more similar to COCO than the majority of the noisy 0.8B data.
Meanwhile, the Base model with limited capacity
may not be able to benefit effectively from large-scale data. 
Similar observations are also reported in~\cite{kolesnikov2020big} for ImageNet-1k classification. 
On TextCaps and VizWiz-QA, all model variants benefit significantly from more pre-training data.
Also, a larger backbone improves more especially with 0.8B data. 

Here, we scale the image encoder.
Empirically, we find it is difficult to effectively scale up the text decoder. Preliminary results are shown 
in Table~\ref{tab:varying_decoder_layers}, which shows a larger decoder shows no improvement.
The reason might be that it is difficult to effectively train with limited amount of text by LM. 
Another plausible reason is that the image encoder is responsible for object recognition,
and the decoder is responsible for 
organizing the object terms in a natural language way. The latter task might be easy since most of the descriptions follow similar patterns, e.g. object + verb + subject, and thus a small decoder is enough during end-to-end training. Larger decoders increase the learning difficulty, which might degrade the performance. 

Flamingo~\citep{alayrac2022flamingo} shows a larger decoder improves the
performance. However, their decoder is pre-trained and frozen during 
the VL pre-training, which avoids the problem of how to effectively train the decoder.
In LEMON~\citep{abs-2111-12233}, the transformer can be scaled up to 32 layers. The reason could be that LEMON uses MLM, instead of LM, which might be more difficult to train.  

\begin{table}[t]
    \centering
    \small
    \caption{Ablation study of larger text decoders. The models are pre-trained on a subset of 0.4B image-text pairs. No beam search and no SCST are performed.}
    \begin{tabular}{ccccccc}
    \toprule
    \multirow{2}{*}{Layers}     &    \multicolumn{4}{c}{COCO}        & \multicolumn{2}{c}{nocaps}    \\
            \cmidrule(lr){2-5} \cmidrule(lr){6-7}
               & B@4     &  M    & C         & S                     & C     & S             \\ 
               \midrule
    6          & 38.9    & 30.7  & 136.4     & 24.6                  & 119.3 & 15.9      \\ 
    12         & 38.9    & 30.6  & 136.0     & 24.2                  & 118.1 & 15.5 \\     
    24         & 39.1    & 30.2  & 134.6     & 23.8                  & 115.4 & 15.1 \\ 
    \bottomrule
    \end{tabular}
    \label{tab:varying_decoder_layers}
\end{table}

\noindent\textbf{Scene text in pre-training data.}
To understand the capability of scene text comprehension, we examine the pre-training dataset and study how many image-text pairs contain the scene text. We first run the Microsoft Azure OCR API\footnote{
{https://docs.microsoft.com/en-us/azure/cognitive-services/computer-vision/concept-recognizing-text}} against all images in CC12M and 
500K images in the web crawled images. 
The OCR result is compared with the
associated text.
It is considered \textit{matched} only if the text contains 
an OCR result that is longer than 5 characters.
It is estimated that 15\% of CC12M and 31\% of the downloaded images contain scene text descriptions.
As the training task is to predict the texts, the network gradually learns to read the scene text. 



\section{Conclusion}
In the paper, we design and train a simple generative model, named \modelname, 
to map the input image
to the associated text description on large-scale image-text pairs.
On image/video captioning and question answering tasks,
our model achieves new  state-of-the-art performance across numerous benchmarks and surpasses the human performance on TextCaps for the first time.
For the image classification, we apply the generation task
to predict the label name directly. 
The strategy is
different from the existing work with a pre-defined and fixed vocabulary,
and is beneficial especially when  new category data are added. 

\noindent\textbf{Limitations.}
We focus on the pretraining-and-finetuning strategy to 
improve the absolute performance. 
Empirically, we find it is unclear on how to control the generated
caption and how to perform in-context learning without parameter update,
which we leave as future work. 

\noindent\textbf{Societal impact.}
Compared with the existing work, our model clearly improves the performance 
and be more appropriate to help visually-impaired people.
The model is pre-trained on large-scale data, and the data
are not guaranteed to contain no toxic language, which may poison
the output. Although we observe few such instances 
qualitatively, special care should be taken to deploy the model
in practice and more research exploration is required to control
the output.

%






\appendix
\newcommand{\fivewidth}{0.18}
\appendix
\section*{Appendix}
The supplementary materials provide more details on the experiments,
including results with different model variants,
more visualizations,
ablation analysis on decoder architectures,
more results on data and model scaling, \emph{etc}.



\section{Setting}
\subsection{Data Preprocessing}
We follow~\cite{ufo} to preprocess the pre-training data. That is, make sure the shorter length of the image no larger than 384 and the longer side no larger than 640 while maintaining the aspect ratio. 
Meanwhile, all images are re-saved with quality being 90 in the JPEG format.
This results in 39 terabytes.
No such preprocessing is applied on the fine-tuning dataset.

\subsection{Platform}
The data are stored in Azure Blob Storage\footnote{\url{https://azure.microsoft.com/en-us/services/storage/blobs}},
and the training is conducted on A100 
provisioned by 
Azure Machine Learning\footnote{\url{https://docs.microsoft.com/en-us/azure/machine-learning/}}.
The code is in python with packages
including 
Pytorch\footnote{\url{https://pytorch.org/}, license: \url{https://github.com/pytorch/pytorch/blob/master/LICENSE}}
DeepSpeed\footnote{\url{https://github.com/microsoft/DeepSpeed}, MIT license)},
Transformers\footnote{\url{https://github.com/huggingface/transformers}, Apache License 2.0},
maskrcnn-benchmark\footnote{\url{https://github.com/facebookresearch/maskrcnn-benchmark}, MIT license},
CLIP\footnote{\url{https://github.com/openai/CLIP}, MIT license},
OSCAR\footnote{\url{https://github.com/microsoft/Oscar}, MIT license},
and VirTex~\citep{virtex}\footnote{\url{https://github.com/kdexd/virtex}, 
MIT license}.

\subsection{Network}
In the main paper, we present the results of our \modelname. 
Here, we construct two smaller model variants, named 
\modelnamewospace$_\text{B}$ and \modelnamewospace$_\text{L}$ on smaller pre-training dataset.
As shown in Table~\ref{tab:networks_config}, 
\modelnamewospace$_\text{B}$ uses CLIP/ViT-B/16~\citep{clip}
as the image encoder and is pre-trained on 10M image-text pairs or 4M images, which 
is a combination of COCO, SBU, CC3M and VG. 
\modelnamewospace$_\text{L}$ uses CLIP/ViT-L/14~\citep{clip}
as the image encoder and is pre-trained on 20M image-text pairs or 14M images, which is a
combination of the 10M image-text pairs with CC12M.

The three model variants share the same pre-training hyperparameters. 
The learning rate is warmed up in the first 500 iterations, and then follows cosine decay to 0. 
The learning rate is $1e^{-5}$ for the image encoder and is multiplied by 5 
for the randomly initialized text decoder. 
The batch size is 4096. Parameters are updated
by AdamW~\citep{adamw} with $\beta_1 = 0.9$
and $\beta_2 = 0.999$.
The number of epochs is 2.


\begin{table}[]
    \centering
    \caption{Model configurations in pre-training. 
    The decoder is a 6-layer transformer network. 
    The hidden size is 768 with 12 attention heads except \modelnamewospace2. 
    Parameters of text token embeddings and the last projection weight before the softmax layer are shared and not counted 
    in the model size. 
    }
    \label{tab:networks_config}
    \begin{tabular}{l@{~~}c@{~~}c@{~~}c@{~~}c@{~~}c@{~~}c}
    \toprule
    Name              &  images         & image-text pairs & image encoder                 & epochs & model size  & image size\\
    \midrule
    \modelnamewospace$_\text{B}$ &  4M          & 10M              & CLIP/ViT-B/16~\citep{clip}     & 30     & 129M & 224\\
    \modelnamewospace$_\text{L}$ &  14M         & 20M              & CLIP/ViT-L/14~\citep{clip}     & 30     & 347M & 224\\
    \modelname            &  0.8B        & 0.8B             & Florence/CoSwin~\citep{abs-2111-11432}   & 2               & 681M & 384 \\
    \midrule
    \modelnamewospace2    & 10.5B        & 12.9B            & DaViT~\citep{ding2022davit} (4.8B)  & 2 & 5.1B & 384 \\
    \bottomrule
    \end{tabular}
\end{table}

As the performance exhibits no signs of
plateau, we further scale up the model size to 5.1B and the
number of pretraining images to 10.5B (12.9B image-text pairs).
The image encoder is scaled to 4.8B based on DaViT~\citep{ding2022davit}
and is pre-trained with the UniCL~\citep{Yang_2022_CVPR,abs-2111-11432} task.
The text decoder is enlarged to 0.3B, the hyperparameters (number of transformer layers, hidden dimension, etc) of which follow BERT-Large~\citep{devlin2018bert}.
The model is named as \modelnamewospace2.

\subsection{Implementation of the Data Loader}
A challenging problem is to implement the data loader efficiently as the total data size (39TB for the 0.8B images) is much larger than the local disk size 
(around 7TB). 
As the data are stored in Azure Storage, we download the data to the local disk before reading it rather than 
directly from the cloud.
Considering the data scale may increase even larger in the future, 
we should make sure each operation is independent to the dataset size.
In the meanwhile, the data downloading should be overlapped with the GPU computing, 
such that
the data are always locally available when needed.
The solution is outlined as follows.
\begin{enumerate}
    \item The image-text pairs are evenly split among $C$ compute nodes. Each node only accesses the corresponding part.
    \item Each node consumes the data trunk by trunk. 
    Each trunk is $2^{20}$ image-text pairs except the last which may have fewer than $2^{20}$ data.
    \item The data in each trunk is randomly shuffled. 
    We shuffle the data in the trunk level such that the cost is not related with the dataset size, and hence it can be applied to even larger dataset.
    \item The shuffled trunk data are split evenly among the GPUs within the node. 
    \item One extra process on each node (launched by local rank = 0) is created to pre-fetch at most 7 future trunks. 
    As each trunk is designed for all ranks in one node, it is not required for other ranks to launch the pre-fetching process, 
    which avoids the race condition. 
    \item Local storage contains at most 12 trunk data, and the oldest will be removed.
\end{enumerate}

Empirically, we observe almost no\footnote{That is, the data preprocessing is faster than the training and is overlapped with the GPU training.} time cost on the data loading 
during model training and the speed is 
also stable.

\section{Results on Image Captioning}
On each task, the model is fine-tuned with 10 epochs. The batch size is 512 and the learning rate is $2.5e^{-6}$. 
SCST~\citep{scst} follows the same hyperparameters if performed.

\begin{table}[]
	\centering
	\small
	\caption{
	Results on COCO captioning with 
	Karpathy~\citep{KarpathyL15} split.
	SimVLM: C4 (800GB) dataset are used and not included in the table;
    Flamingo: 27M video-text pairs are not counted in the table.
    UniversalCaptioner: the extra 0.3B in parameters is CLIP/ViT-L, which is used as feature and keyword extractor. 
    the data for pre-training CLIP/ViT-L are not counted .
    VinVL/LEMON/OSCAR/MiniVLM/DistillVLM: the extra parameters are for object detector; data for the object detectors are not counted. 
    CoCa: ~\cite{yu2022coca},
    BLIP: ~\cite{abs-2201-12086},
    mPLUG: ~\cite{li2022mplug},
    MiniVLM: ~\cite{abs-2012-06946},
    DistillVLM: ~\cite{FangW0WY021},
    Flamingo: ~\cite{alayrac2022flamingo},
    LEMON: ~\cite{abs-2111-12233},
    OSCAR: ~\cite{Li0LZHZWH0WCG20},
    OFA: ~\cite{abs-2202-03052},
    UFO: ~\cite{ufo},
    UniversalCap: ~\cite{abs-2111-12727},
    VinVL: ~\cite{abs-2101-00529},
    ViTCap: ~\cite{abs-2112-05230},
    SimVLM: ~\cite{wang2021simvlm}.
    }
	\label{tab:full_coco_caption}
	\begin{tabular}{@{}lc@{~~}c@{~~}cccccccc@{}}
		\toprule
        \multirow{2}{*}{Method}               & \multirow{2}{*}{\#Param.} & \multirow{2}{*}{\#Images}  & \multicolumn{4}{c}{Cross-Entropy}                     & \multicolumn{4}{c}{SCST}                       \\ \cmidrule(lr){4-7} \cmidrule(lr){8-11}
		                                      &      &          & B@4               &  M               & C              & S             & B                 &  M               & C              & S      \\\midrule            
        \multicolumn{11}{c}{Tiny-sized models} \\ \midrule
		MiniVLM         & 46M+8M  & 11M      & 35.6              & 28.6             & 119.8          & 21.6          &  39.2             & 29.7             & 131.7          & 23.5   \\
 		DistillVLM         & 46M+8M  & 4M       & 35.6              & 28.7             & 120.8          & 22.1          &  -                &  -               & -              & -      \\ \midrule
 		
 		\multicolumn{11}{c}{Base-sized models} \\ \midrule
		ViTCap          & 0.2B  & 4M      & 36.3              & 29.3             & 125.2          & 22.6          & 41.2              & 30.1             & 138.1          & 24.1    \\ 
OSCAR$_\text{B}$       & 0.1B+64M  & 4M      & 36.5              &  30.3            & 123.7          & 23.1          &  40.5             & 29.7             & 137.6          & 22.8 \\
VinVL$_\text{B}$        & 0.1B+0.2B  & 6M      & 38.2              & 30.3             & 129.3          & 23.6          &  40.9             & 30.9             & 140.4          & 25.1 \\
		UFO$_\text{B}$             & 0.1B  & 4M      & 36.0              & 28.9             & 122.8          & 22.2          & -                 & -                &  -             &  - \\ 
UniversalCap$_\text{B}$  & 0.2B+0.3B & 36M      & -                 & -                 & -              & -          & 42.9               & 31.4                &  149.7             &  25.0 \\ 
		\midrule
 		\modelnamewospace$_\text{B}$          & 0.1B  & 4M      & 40.4              & 30.0             & 131.4          & 23.0          & 41.3              & 30.4             & 139.1          & 24.3 \\ \midrule 
		
		\multicolumn{11}{c}{Large-sized models} \\ \midrule
OSCAR$_\text{L}$       & 0.3B+64M  & 4M      & 37.4              & 30.7             & 127.8          & 23.5          &  41.7             & 30.6             & 140.0          & 24.5   \\
VinVL$_\text{L}$        & 0.3B+0.2B  & 6M      & 38.5              & 30.4             & 130.8          & 23.4          &  41.0             & 31.1             & 140.9          & 25.2    \\
		UFO$_\text{L}$             & 0.3B  & 4M      & 38.7              & 30.0             & 131.2          & 23.3          &   -               & -                & -              & -  \\ 
BLIP$_\text{ViT-L}$     & -     & 129M    & 40.4              & -                & 136.7          &  -            &   -               & -                & -              & -  \\
UniversalCap$_\text{L}$ & 0.5B+0.3B  & 36M     & -                 & -                & -              & -             & 42.9              & 31.5             & 150.2          & 25.2 
\\
 mPLUG                    &   0.6B     & 14M   & 43.1 & 31.4 & 141.0 & 24.2 & \bf46.5 &  32.0 &  \bf155.1 & 26.0 \\
\midrule  
 		\modelnamewospace$_\text{L}$          & 0.3B  & 14M     & 42.0              & 30.8             & 138.5          & 23.8          & 42.3              & 31.2             & 144.6          & 25.4 \\ \midrule  
        
        \multicolumn{11}{c}{Huge/Giant-sized models} \\ \midrule
		Flamingo   & 80B   & 2.3B   &              -    & -                & 138.1          & -             & -                 & -                & -              & -  \\
LEMON$_\text{huge}$     & 0.7B+0.2B  & 0.2B   & 41.5              & 30.8             & 139.1          & 24.1          & 42.6              & 31.4             & 145.5          & 25.5 \\
SimVLM$_\text{Huge}$    & -     & 1.8B   & 40.6              & 33.7    & 143.3          & \textbf{25.4} &  -                &  -               & -              & -  \\
OFA    & 0.9B  &  54M      & 43.9 & 31.8 & \bf145.3 & 24.8 & \bf44.9 & \bf32.5 & 154.9 & \bf26.6 \\
		CoCa                & 2.1B  & 4.8B   & 40.9              & \bf33.9             & 143.6          & 24.7          & -                 &  -               & -              & -  \\
		\midrule
		\modelname                            & 0.7B  & 0.8B   & \textbf{44.1}     & 31.5             & 144.8 & 24.7          & 44.1     & 32.2    & 151.1 & 26.3  \\ 
		\midrule
		\modelnamewospace2   & 5.1B  & 10.5B       & \bf44.1     & 31.4 &  145.0 & 24.8           & 44.0     & 32.2 &  152.7 & 26.4 \\ 
        \bottomrule
	\end{tabular}
\end{table}

\begin{table}[]
    \centering
    \caption{Results on  COCO test set evaluated on the public server. c5/c40: Each image is paired with $5$ or $40$ reference ground-truth captions. B: BLEU~\citep{PapineniRWZ02}; M: METEOR~\citep{DenkowskiL14}; R: ROUGE-L~\citep{LinO04}; C: CIDEr-D~\citep{VedantamZP15}.
    BUTD: ~\cite{00010BT0GZ18},
    VinVL: ~\cite{abs-2101-00529}.
    }
    \label{tab:coco_full_test}
    \begin{tabular}{@{}l@{~~}c@{~~}c@{~~~}c@{~~}c@{~~~}c@{~~}c@{~~~}c@{~~}c@{~~~}c@{~~}c@{~~~}c@{~~}c@{~~~}c@{~~}c@{}}
    \toprule
    \multirow{2}{*}{Method}  & \multicolumn{2}{c}{B@1} & \multicolumn{2}{c}{B@2} & \multicolumn{2}{c}{B@3} & \multicolumn{2}{c}{B@4} & \multicolumn{2}{c}{M} & \multicolumn{2}{c}{R} & \multicolumn{2}{c}{C} \\
            \cmidrule(lr){2-3} \cmidrule(lr){4-5} \cmidrule(lr){6-7} \cmidrule(lr){8-9} \cmidrule(lr){10-11} \cmidrule(lr){12-13} \cmidrule(lr){14-15}
              & c5    & c40  & c5   & c40  & c5   & c40  & c5   & c40  & c5   & c40  & c5   & c40  & c5    & c40 \\
              \midrule
    BUTD      &  80.2 & 95.2 & 64.1 & 88.8 & 49.1 & 79.4 & 36.9 & 68.5 & 27.6 & 36.7 & 57.1 & 72.4 & 117.9 & 120.5 \\
    VinVL     & 81.9  & 96.9 & 66.9 & 92.4 & 52.6 & 84.7 & 40.4 & 74.9 & 30.6 & 40.8 & 60.4 & 76.8 & 134.7 & 138.7 \\
    \midrule
\modelnamewospace & 84.0	 & 97.9     & 69.8    & 94.4 & 55.6    & 87.6 & 43.2 & 78.3 & 31.9 & 42.0 & 62.0 & 78.4 & 145.5 & 148.8 \\ 
\midrule
\modelnamewospace2           & 84.5 & 98.1  & 70.0 & 94.4 & 55.7 & 87.6 & 43.2 & 78.3   & 31.9 & 42.1 & 62.0 & 78.4 & 146.4 & \bf149.8 \\
    \bottomrule
    \end{tabular}
\end{table}

\noindent\textbf{COCO}
Fig.~\ref{tab:full_coco_caption} shows the complete results including \modelnamewospace$_\text{B}$
and \modelnamewospace$_\text{L}$ on COCO Karpathy split~\citep{KarpathyL15}.
For the base-sized and large-sized models, our model achieves competitive performance with existing approaches but with a simplified architecture.
We observe that UniversalCaptioner~\citep{abs-2111-12727} achieves much better performance.
As a strong image encoder of CLIP/ViT-L with 0.3B parameters
is used in UniversalCaptioner for both the base and large model, effectively, the model size is much larger than those in respective categories.
In the meanwhile, both UniversalCaptioner~\citep{abs-2111-12727} and 
OFA~\citep{abs-2202-03052} use more data than our approach within base/large-sized model sizes.
Fig.~\ref{tab:coco_full_test} shows the full results on the COCO test set.

\noindent\textbf{nocaps.}
The main paper presents the overall performance on nocaps. Table~\ref{tab:full_nocaps} contains 
the complete results for each sub domain and other model variants.
\begin{table*}[t]
	\centering
	\small
	\caption{Results on nocaps.
	in.: in-domain; near.: near domain; out.: out-of-domain; C: CIDEr. S: SPICE.
	OSCAR: ~\cite{Li0LZHZWH0WCG20},
	Human: ~\cite{abs-1812-08658},
	VIVO: ~\cite{abs-2009-13682},
	VinVL: ~\cite{abs-2101-00529},
	UFO: ~\cite{ufo},
	mPLUG: ~\cite{li2022mplug},
	SimVLM: ~\cite{wang2021simvlm},
	LEMON: ~\cite{abs-2111-12233},
	UniversalCap: ~\cite{abs-2111-12727},
	CoCa: ~\cite{yu2022coca}.
	}
	\label{tab:full_nocaps}
    \begin{tabular}{@{~}l@{~}c@{~}c@{~}c@{~}c@{~}c@{~}c@{~}c@{~}c@{~~~}c@{~}c@{~}c@{~}c@{~}c@{~}c@{~}c@{~}c@{~}c@{~}c@{~}c@{~}c@{~}c@{~}c@{~}c@{~}c@{~}}
         \toprule
		\multirow{3}{*}{Method}      & \multicolumn{8}{c}{Validataion set}  & \multicolumn{8}{c}{Test set} \\ \cmidrule(lr){2-9} \cmidrule(l){10-17}
		& \multicolumn{2}{c}{{in.}}  & \multicolumn{2}{c}{{near.}}  & \multicolumn{2}{c}{{out.}} & \multicolumn{2}{c}{{overall}} & \multicolumn{2}{c}{{in.}}  & \multicolumn{2}{c}{{near.}}  & \multicolumn{2}{c}{{out.}} & \multicolumn{2}{c}{{overall}} \\
		\cmidrule(lr){2-3}               \cmidrule(lr){4-5}                \cmidrule(lr){6-7}                  \cmidrule(lr){8-9} \cmidrule(lr){10-11} \cmidrule(lr){12-13} \cmidrule(lr){14-15} \cmidrule(lr){16-17}
		                                    & C              & S       & C          & S           & C          & S         & C          & S   & C              & S       & C          & S           & C          & S         & C          & S \\
		\midrule                         
		OSCAR        & 85.4           & 11.9    & 84.0       & 11.7        & 80.3       & 10.0      & 83.4       & 11.4  & 84.8       & 12.1     & 82.1     & 11.5      & 73.8      & 9.7       & 80.9      & 11.3    \\
		Human         & 84.4           & 14.3    & 85.0       & 14.3        & 95.7       & 14.0      & 87.1       & 14.2    & 80.6       & 15.0     & 84.6     & 14.7      & 91.6      & 14.2      & 85.3      & 14.6     \\
		VIVO          & 92.2           & 12.9    & 87.8       & 12.6        & 87.5       & 11.5      & 88.3       & 12.4    & 89.0       & 12.9     & 87.8     & 12.6      & 80.1      & 11.1      & 86.6      & 12.4         \\
		VinVL         & 103.7          & 13.7    & 95.6       & 13.4        & 83.8       & 11.9      & 94.3       & 13.1    & 98.0       & 13.6     & 95.2     & 13.4      & 78.0      & 11.5      & 92.5      & 13.1        \\
		UFO           & 103.9          & 14.5    & 95.5       & 13.8        & 83.5       & 12.3      & 94.3       & 13.6   & 98.9       & 14.3     & 94.7     & 13.9      & 77.9      & 12.1      & 92.3      & 13.6   \\ 
		mPLUG           & - & - & - & - & - & - & 114.8 & 14.8 & - & - & - & - & - & - & - & - \\
		SimVLM        & 113.7          & -       & 110.9      & -           & 115.2      & -         & 115.2      & -            & 113.7      & -        & 110.9    & -         & 115.2     & -         & 115.2     & -              \\
		LEMON         & 118.0          & 15.4    & 116.3      & 15.1        & 120.2      & 14.5      & 117.3      & 15.0  & 112.8       & 15.2      & 115.5    & 15.1 & 110.1  & 13.7      & 114.3     & 14.9  \\

		UniversalCap  & 123.2          & 15.0    & 121.5      & 15.3        & 123.4      & 14.4      & 122.1      & 15.0  & 118.9      & 15.4     & 120.6    & 15.3      & 114.3     & 14.1      & 119.3     & 15.1 \\
		CoCa              & -              & -       & -          & -           & -          & -         & 122.4      & 15.5  & -          & -        & -        & -         & -         & -          & 120.6    & 15.5 \\
		\midrule
		
	    \modelnamewospace$_\text{B}$   & 100.7  & 13.8   & 97.7   & 13.5  & 89.6  & 12.5  & 96.6    & 13.4  &  -        &   -     &   -     & -        &  -       &      -    &   -      &   -       \\ 
	    \modelnamewospace$_\text{L}$   & 107.7  & 14.9   & 107.8  & 14.5  & 102.5 & 13.7  & 106.9   & 14.4  &  -        &   -     &   -     & -        &  -       &      -    &   -      &   -       \\ 
	    \modelname                   & \bf129.8 & \bf16.3& 124.1     & 16.0    & 127.1  & 15.7  & 125.5  & 16.0     & 122.4  & 16.2 & 123.9& 16.0  & 122.0 & \bf15.7   & 123.4 & 15.9   \\
	    \midrule
	    \modelnamewospace2      & 126.9 & 16.1& \bf125.8     & \bf16.2    & \bf130.6  & \bf15.8  & \bf126.9  & \bf16.1     & \bf124.2  & \bf16.4 & \bf125.5& \bf16.1  & \bf122.3 & 15.6   & \bf124.8 & \bf16.1   \\ 
		\bottomrule
	\end{tabular}
\end{table*}

\begin{figure}
\centering
\small
\renewcommand{\arraystretch}{0.8}
\begin{tabular}{p{\fivewidth\linewidth}@{~~}p{\fivewidth\linewidth}@{~~}p{\fivewidth\linewidth}@{~~}p{\fivewidth\linewidth}@{~~}p{\fivewidth\linewidth}}
\includegraphics[width=1\linewidth,height=1\linewidth]{fig/image/nocaps/589eb1664bed4c06.jpg} & \includegraphics[width=1\linewidth,height=1\linewidth]{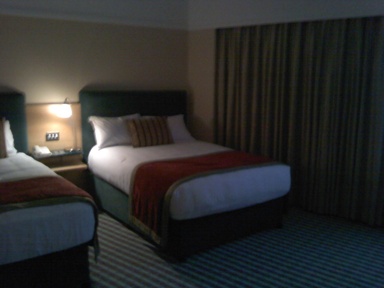} & \includegraphics[width=1\linewidth,height=1\linewidth]{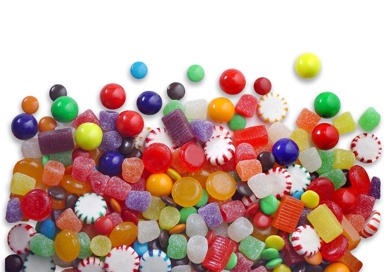} & \includegraphics[width=1\linewidth,height=1\linewidth]{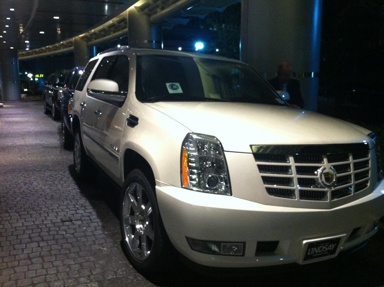} & \includegraphics[width=1\linewidth,height=1\linewidth]{fig/image/nocaps/2a76b21412394e0b.jpg}\\
\textbf{Pred}: a bunch of green grapes hanging from a vine. & \textbf{Pred}: a hotel room with two beds in a room. & \textbf{Pred}: a pile of candy on a white background. & \textbf{Pred}: a group of cars parked on the side of a building. & \textbf{Pred}: a blue and white spotted \underline{stingray} laying on the sand.\\
\includegraphics[width=1\linewidth,height=1\linewidth]{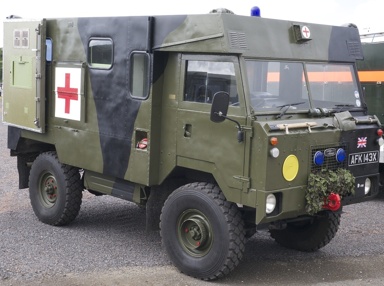} & \includegraphics[width=1\linewidth,height=1\linewidth]{fig/image/nocaps/7a87a07e3e0a1637.jpg} & \includegraphics[width=1\linewidth,height=1\linewidth]{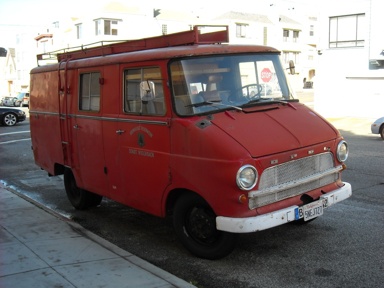} & \includegraphics[width=1\linewidth,height=1\linewidth]{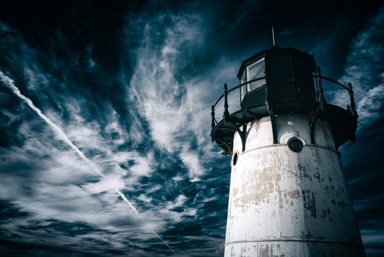} & \includegraphics[width=1\linewidth,height=1\linewidth]{fig/image/nocaps/50ff8448a2872f3d.jpg}\\
\textbf{Pred}: a green medical truck parked in a parking lot. & \textbf{Pred}: a red drum set and two guitars in a room. & \textbf{Pred}: a red truck parked on the side of a street. & \textbf{Pred}: a white lighthouse with a cloudy sky. & \textbf{Pred}: a lion laying in the grass next to a log.\\
\includegraphics[width=1\linewidth,height=1\linewidth]{fig/image/nocaps/9fbcea57c04cad15.jpg} & \includegraphics[width=1\linewidth,height=1\linewidth]{fig/image/nocaps/9c4d6c06fe89e558.jpg} & \includegraphics[width=1\linewidth,height=1\linewidth]{fig/image/nocaps/d6388aa972872b97.jpg} & \includegraphics[width=1\linewidth,height=1\linewidth]{fig/image/nocaps/438118dbd16f942f.jpg} & \includegraphics[width=1\linewidth,height=1\linewidth]{fig/image/nocaps/50f231b09e848f15.jpg}\\
\textbf{Pred}: a glass of coffee and a glass of milk. & \textbf{Pred}: a group of bread buns sitting on a cooling rack. & \textbf{Pred}: a bunch of oysters cooking on a grill. & \textbf{Pred}: a close up of a cat sitting in the grass. & \textbf{Pred}: a spider on its web with a spider in it.\\
\includegraphics[width=1\linewidth,height=1\linewidth]{fig/image/nocaps/3a9bc59b36b64e73.jpg} & \includegraphics[width=1\linewidth,height=1\linewidth]{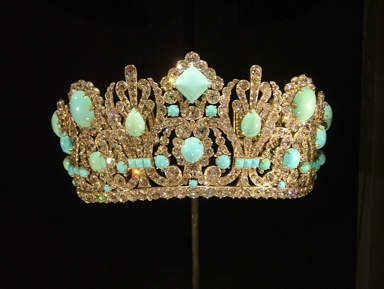} & \includegraphics[width=1\linewidth,height=1\linewidth]{fig/image/nocaps/627f9e6d648f2768.jpg} & \includegraphics[width=1\linewidth,height=1\linewidth]{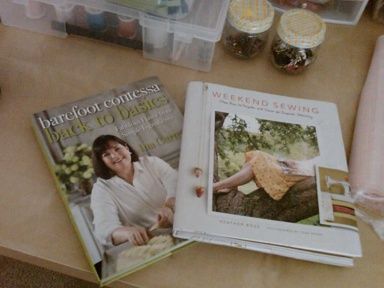} & \includegraphics[width=1\linewidth,height=1\linewidth]{fig/image/nocaps/ed897e3c646c618c.jpg}\\
\textbf{Pred}: a stack of pancakes sitting on a plate with syrup. & \textbf{Pred}: a blue crown on a stand on a black background. & \textbf{Pred}: a group of blue flowers in a field of tall grass. & \textbf{Pred}: two books sitting on top of a table. & \textbf{Pred}: a wooden cabinet sitting on top of a wooden floor.\\
\includegraphics[width=1\linewidth,height=1\linewidth]{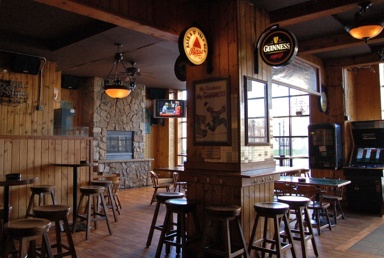} & \includegraphics[width=1\linewidth,height=1\linewidth]{fig/image/nocaps/51968e261767baae.jpg} & \includegraphics[width=1\linewidth,height=1\linewidth]{fig/image/nocaps/0210f9c01e31f873.jpg} & \includegraphics[width=1\linewidth,height=1\linewidth]{fig/image/nocaps/1390f4bb25697890.jpg} & \includegraphics[width=1\linewidth,height=1\linewidth]{fig/image/nocaps/7f13b5581a303c26.jpg}\\
\textbf{Pred}: a bar or pub with stools and a table. & \textbf{Pred}: a pool with a bed next to it in a yard. & \textbf{Pred}: a couple of cars parked in a showroom. & \textbf{Pred}: a close up of a white ceiling fan. & \textbf{Pred}: a group of dogs in the snow.\\
\end{tabular}
\caption{Captioning results of our COCO-fine-tuned GIT on random samples from the nocaps validation set. Words not in COCO training captions are underlined.}
\label{fig:nocaps_random}
\end{figure}

\begin{figure}
\centering
\small
\renewcommand{\arraystretch}{0.8}
\begin{tabular}{p{\fivewidth\linewidth}@{~~}p{\fivewidth\linewidth}@{~~}p{\fivewidth\linewidth}@{~~}p{\fivewidth\linewidth}@{~~}p{\fivewidth\linewidth}}
\includegraphics[width=1\linewidth,height=1\linewidth]{fig/image/nocaps/1b7b604aa0d293cd.jpg} & \includegraphics[width=1\linewidth,height=1\linewidth]{fig/image/nocaps/993839d04015d05b.jpg} & \includegraphics[width=1\linewidth,height=1\linewidth]{fig/image/nocaps/14a7fbad2d02e768.jpg} & \includegraphics[width=1\linewidth,height=1\linewidth]{fig/image/nocaps/34b8358262d2a434.jpg} & \includegraphics[width=1\linewidth,height=1\linewidth]{fig/image/nocaps/69f87f239d88063f.jpg}\\
\textbf{Pred}: a white \underline{stingray} swimming in an aquarium. & \textbf{Pred}: a \underline{bumble} bee sitting on a white flower. & \textbf{Pred}: a large \underline{starfish} and fish swimming in the water. & \textbf{Pred}: a blue \underline{starfish} sitting on top of a coral. & \textbf{Pred}: a small \underline{dragonfly} sitting on top of a blade of grass.\\
\includegraphics[width=1\linewidth,height=1\linewidth]{fig/image/nocaps/b160fdbcb56bbced.jpg} & \includegraphics[width=1\linewidth,height=1\linewidth]{fig/image/nocaps/720bb6cc50390263.jpg} & \includegraphics[width=1\linewidth,height=1\linewidth]{fig/image/nocaps/f7bb0283e5f50e4c.jpg} & \includegraphics[width=1\linewidth,height=1\linewidth]{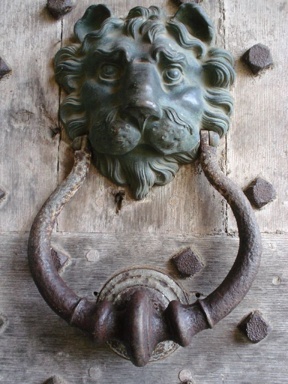} & \includegraphics[width=1\linewidth,height=1\linewidth]{fig/image/nocaps/bddfd477fa68ee18.jpg}\\
\textbf{Pred}: a black and white \underline{lemur} sitting in a tree. & \textbf{Pred}: a green \underline{roulette} table in a room with chairs. & \textbf{Pred}: a couple of \underline{barbies} on a cake. & \textbf{Pred}: a lion head door \underline{knocker} on a wooden door. & \textbf{Pred}: a group of \underline{lipsticks} sitting next to each other.\\
\includegraphics[width=1\linewidth,height=1\linewidth]{fig/image/nocaps/b72faf4782ff46a0.jpg} & \includegraphics[width=1\linewidth,height=1\linewidth]{fig/image/nocaps/d133c517c7947101.jpg} & \includegraphics[width=1\linewidth,height=1\linewidth]{fig/image/nocaps/0049a724f5dc20e4.jpg} & \includegraphics[width=1\linewidth,height=1\linewidth]{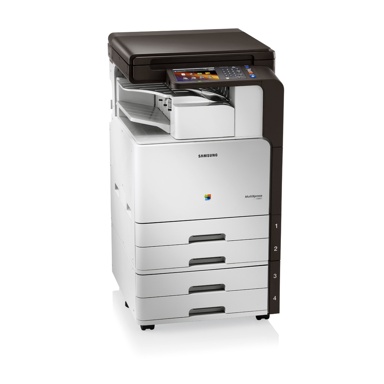} & \includegraphics[width=1\linewidth,height=1\linewidth]{fig/image/nocaps/0794df9274debd2f.jpg}\\
\textbf{Pred}: a bug sitting on top of a yellow \underline{dandelion.} & \textbf{Pred}: a \underline{banjo} sitting on top of a table with a \underline{giveaway} sign. & \textbf{Pred}: a group of \underline{violins} hanging on a wall. & \textbf{Pred}: a \underline{photocopier} machine sitting on top of a white background. & \textbf{Pred}: a closet with a white \underline{kallax} shelves and clothes.\\
\includegraphics[width=1\linewidth,height=1\linewidth]{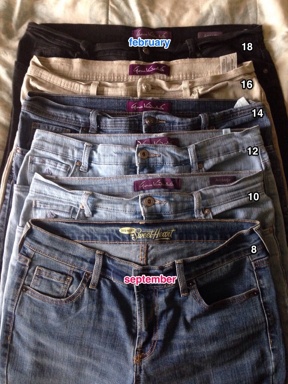} & \includegraphics[width=1\linewidth,height=1\linewidth]{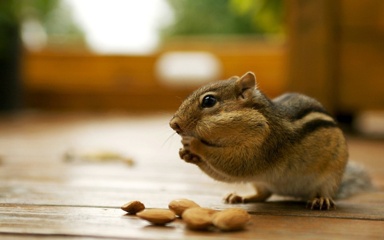} & \includegraphics[width=1\linewidth,height=1\linewidth]{fig/image/nocaps/8af320676429623d.jpg} & \includegraphics[width=1\linewidth,height=1\linewidth]{fig/image/nocaps/24aa01ab4f072bd4.jpg} & \includegraphics[width=1\linewidth,height=1\linewidth]{fig/image/nocaps/2a76b21412394e0b.jpg}\\
\textbf{Pred}: a row of jeans stacked up with the date of \underline{september.} & \textbf{Pred}: a small \underline{chipmunk} eating nuts on the floor. & \textbf{Pred}: a close up of a \underline{wasp} nest on a green leaf. & \textbf{Pred}: a \underline{sonicare} electric toothbrush in a package. & \textbf{Pred}: a blue and white spotted \underline{stingray} laying on the sand.\\
\end{tabular}
\caption{Captioning results of our COCO-fine-tuned GIT on random samples whose prediction contains novel terms from the nocaps validation set. Novel terms, which are not in COCO training captions, are underlined.}
\label{fig:nocaps_novel}
\end{figure}

Fig.~\ref{fig:nocaps_random} shows random\footnote{Disgusting images and images containing clear people identification information are excluded.} prediction examples
on the nocaps validation set.
To visualize the novel concept recognition capability, we also collect
sample images whose prediction contains at least one word not in the COCO training set,
as illustrated in Fig.~\ref{fig:nocaps_novel}. 
As we can see, the model can well identify the novel object without the object tags as the network input.


\begin{table}[]
    \centering
    \caption{Results on TextCaps~\citep{abs-2003-12462}.
    Test set is evaluated by the server. 
    *: the numbers are from~\cite{abs-2003-12462}. 
    B: BLEU@4; M: METEOR; R: ROUGE-L; S: SPICE; C: CIDEr. 
    $^\#$: winner entry of the CVPR 2021 workshop challenge.
    BUTD: ~\cite{00010BT0GZ18},
  AoANet: ~\cite{huang2019attention},
  M4C-Cap.: ~\cite{HuSDR20},
  Anc.-Cap.: ~\cite{abs-2105-03236},
  TAP: ~\cite{abs-2012-04638},
  Human: ~\cite{abs-2003-12462}.
    }
    \label{tab:full_textcaps}
    \begin{tabular}{l@{~~~~~}c@{~~~~~}c@{~~~~~}c@{~~~~~}c@{~~~~~}c@{~~~~~}c@{~~~~~}c@{~~~~~}c@{~~~~~}c@{~~~~~}c}
    \toprule
     Method             & \multicolumn{5}{c}{Validation set} & \multicolumn{5}{c}{Test set} \\
                  \cmidrule(lr){2-6} \cmidrule(l){7-11}
                                    &  B       & M        & R        & S        & C        & B       & M       & R       & S       & C      \\\midrule       
    BUTD*       &  20.1    & 17.8     & 42.9     & 11.7     & 41.9     & 14.9    & 15.2    & 39.9    & 8.8     & 33.8   \\
  AoANet* & 20.4     & 18.9     & 42.9     & 13.2     & 42.7     & 15.9    & 16.6    & 40.4    & 10.5    & 34.6   \\
  M4C-Cap.*          & 23.3     & 22.0     & 46.2     & 15.6     & 89.6     & 18.9    & 19.8    & 43.2    & 12.8    & 81.0   \\
  Anc.-Cap.   & 24.7     & 22.5     & 47.1     & 15.9     & 95.5     & 20.7    & 20.7    & 44.6    & 13.4    & 87.4 \\
  TAP         & 25.8     & 23.8     & 47.9     & 17.1     & 109.2    & 21.9    & 21.8    & 45.6    & 14.6    & 103.2  \\
  TAP$^\#$    & 28.1     & 24.4     & 49.3     & 17.7     & 119.0    & 22.9    & 22.0    & 46.5    & 14.6    & 109.7  \\
  Human       & -        &  -       &  -       &  -       & -        & 24.4    & 26.1    & 47.0    & 18.8    & 125.5  \\   \midrule
\modelnamewospace$_\text{B}$              & 24.1     & 21.1     & 45.2     & 15.7     & 64.9     & -       & -       & -       &  -      & -       \\ 
\modelnamewospace$_\text{L}$              & 30.6     & 24.6     & 50.3     & 18.6     & 106.3    & -       & -       & -       & -       & -       \\ 
\modelname                   & 37.0  & 27.6  & 54.1  & 21.1  & 143.7 & 33.1 & 26.2 & 52.2 & 19.6 & 138.2 \\ 
\midrule
\modelnamewospace2 & \bf38.4  & \bf28.3  & \bf54.6  & \bf21.9  & \bf148.6 & \bf33.8 & \bf27.0 & \bf53.0 & \bf20.2 & \bf145.0 \\ 
    \bottomrule
    \end{tabular}
\end{table}

\begin{figure}
\centering
\small
\renewcommand{\arraystretch}{0.8}
\begin{tabular}{p{\fivewidth\linewidth}@{~~}p{\fivewidth\linewidth}@{~~}p{\fivewidth\linewidth}@{~~}p{\fivewidth\linewidth}@{~~}p{\fivewidth\linewidth}}
\includegraphics[width=1\linewidth,height=1\linewidth]{fig/image/TextCaption/random/c6d14c2ec786cce3.jpg} & \includegraphics[width=1\linewidth,height=1\linewidth]{fig/image/TextCaption/random/03486477f41d883f.jpg} & \includegraphics[width=1\linewidth,height=1\linewidth]{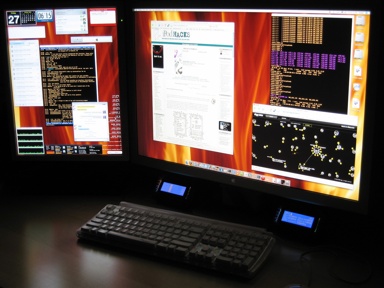} & \includegraphics[width=1\linewidth,height=1\linewidth]{fig/image/TextCaption/random/683faf24809dbcc4.jpg} & \includegraphics[width=1\linewidth,height=1\linewidth]{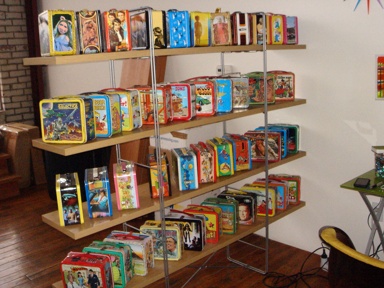}\\
\textbf{Pred}: a bottle of dog house wine sits on a table. & \textbf{Pred}: a green bay packers football field with the goal posts. & \textbf{Pred}: two computer monitors are on a desk, one of which is called ipad hacks. & \textbf{Pred}: a black and red poster that says end police impunity. & \textbf{Pred}: a collection of tin boxes on a shelf with one that says star trek.\\
\includegraphics[width=1\linewidth,height=1\linewidth]{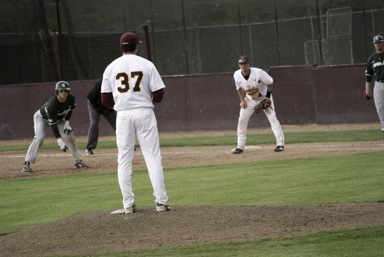} & \includegraphics[width=1\linewidth,height=1\linewidth]{fig/image/TextCaption/random/994a1dba94f4d46a.jpg} & \includegraphics[width=1\linewidth,height=1\linewidth]{fig/image/TextCaption/random/de37d4819233d6dc.jpg} & \includegraphics[width=1\linewidth,height=1\linewidth]{fig/image/TextCaption/random/15f40b53304867ca.jpg} & \includegraphics[width=1\linewidth,height=1\linewidth]{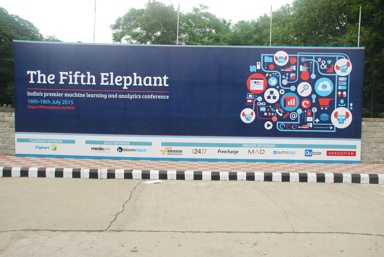}\\
\textbf{Pred}: a baseball player with the number 37 stands on the mound. & \textbf{Pred}: a baseball player with the number 35 on his jersey & \textbf{Pred}: a black and white photo of a floppy disk that says you're an asshole. & \textbf{Pred}: an orange train with the number 16433r on the side. & \textbf{Pred}: a billboard for the fifth elephant machine learning and analytics conference.\\
\includegraphics[width=1\linewidth,height=1\linewidth]{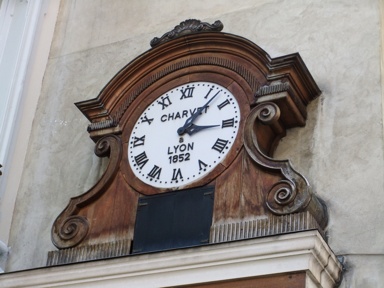} & \includegraphics[width=1\linewidth,height=1\linewidth]{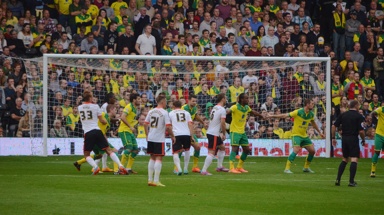} & \includegraphics[width=1\linewidth,height=1\linewidth]{fig/image/TextCaption/random/88f4ce44737175a4.jpg} & \includegraphics[width=1\linewidth,height=1\linewidth]{fig/image/TextCaption/random/1f73da0dee5318fe.jpg} & \includegraphics[width=1\linewidth,height=1\linewidth]{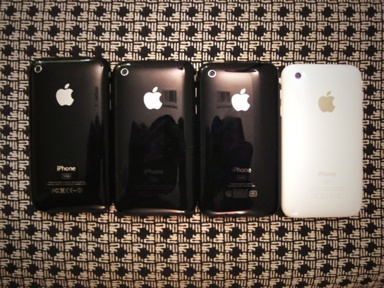}\\
\textbf{Pred}: a clock with the name charvet on it & \textbf{Pred}: soccer players with the number 13 on their jersey & \textbf{Pred}: an airasia zest plane is parked on the tarmac. & \textbf{Pred}: a can of sanpellegrino chino sits on a table. & \textbf{Pred}: four iphones are lined up on a carpet.\\
\includegraphics[width=1\linewidth,height=1\linewidth]{fig/image/TextCaption/random/4c79aab76aba13b6.jpg} & \includegraphics[width=1\linewidth,height=1\linewidth]{fig/image/TextCaption/random/36ab0ae0998bfd7b.jpg} & \includegraphics[width=1\linewidth,height=1\linewidth]{fig/image/TextCaption/random/dc2ea493b38c8751.jpg} & \includegraphics[width=1\linewidth,height=1\linewidth]{fig/image/TextCaption/random/bd63f6f2a002ee10.jpg} & \includegraphics[width=1\linewidth,height=1\linewidth]{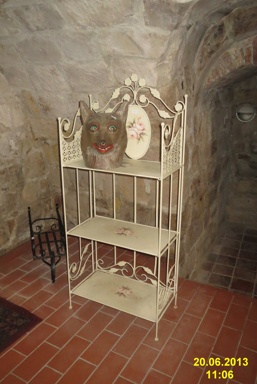}\\
\textbf{Pred}: a grocery store aisle full of canned peas. & \textbf{Pred}: a white lenovo laptop with a black screen on it. & \textbf{Pred}: a green street sign for troublesome valley road. & \textbf{Pred}: a cd cover that says " rarities of piano music at schloss vor kusum ". & \textbf{Pred}: a white shelf with a picture of a dog on it.\\
\includegraphics[width=1\linewidth,height=1\linewidth]{fig/image/TextCaption/random/a6ba8a7b7f10f515.jpg} & \includegraphics[width=1\linewidth,height=1\linewidth]{fig/image/TextCaption/random/df1c84eb74c79d38.jpg} & \includegraphics[width=1\linewidth,height=1\linewidth]{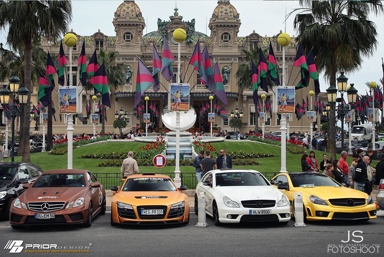} & \includegraphics[width=1\linewidth,height=1\linewidth]{fig/image/TextCaption/random/493f5e5e22c7b221.jpg} & \includegraphics[width=1\linewidth,height=1\linewidth]{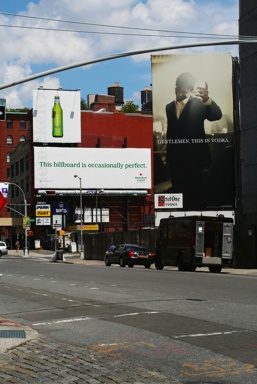}\\
\textbf{Pred}: a poster for the game dynasty warriors gundam reborn. & \textbf{Pred}: the cover of the book the energy glut by ian roberts and phil edwards. & \textbf{Pred}: several cars are lined up in front of a casino. & \textbf{Pred}: a bottle of il bruciato wine from 2009. & \textbf{Pred}: a billboard that says this billboard is occasionally perfect.\\
\end{tabular}
\caption{Visualization of our model on random validation images of TextCaps.}
\label{fig:TextCaption_random}
\end{figure}

\begin{figure}[]
\centering
\includegraphics[width=0.87\textwidth]{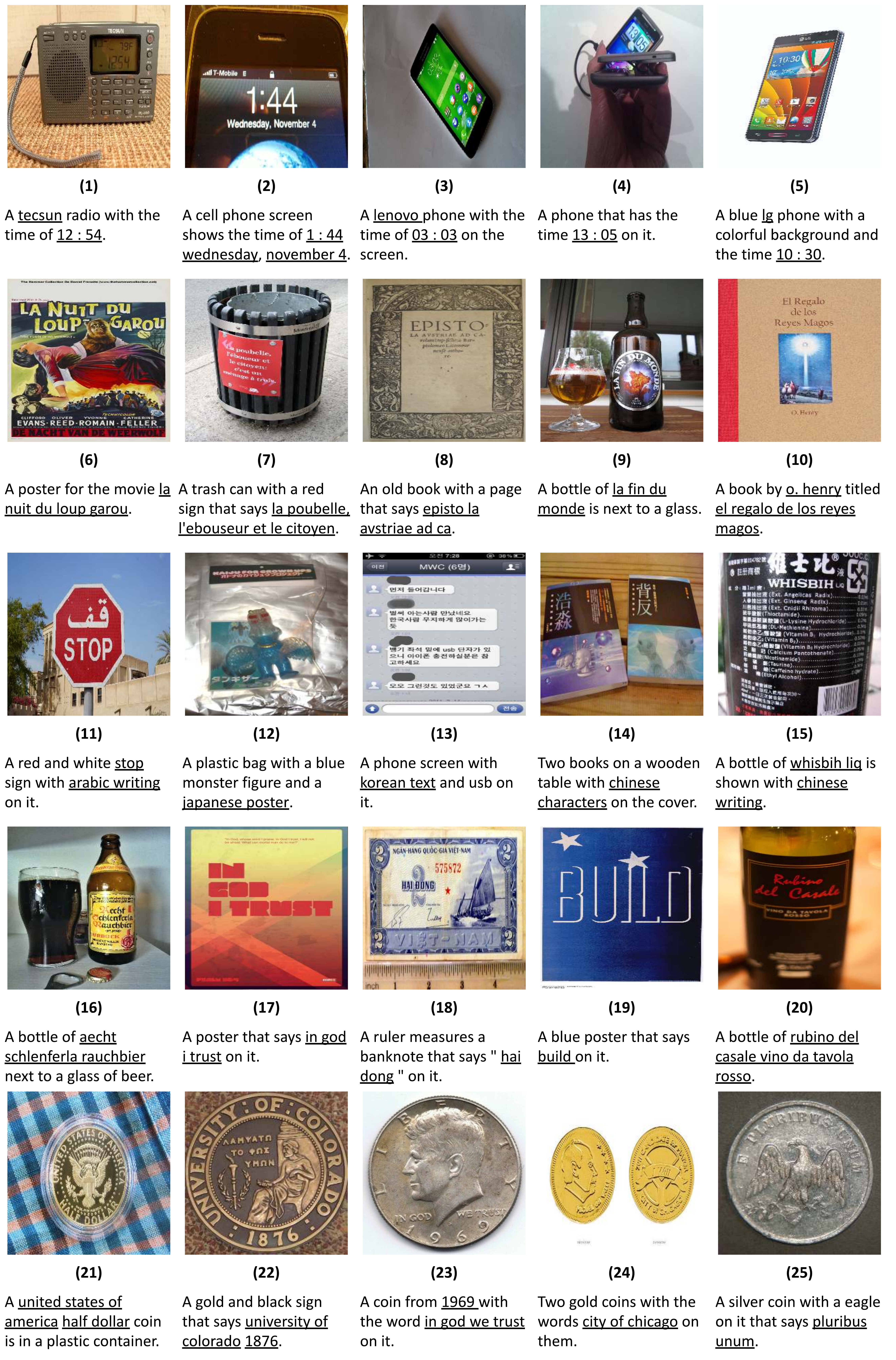}
\caption{
Grouped caption predictions from TextCaps.
The scene text is underlined in descriptions.
(1-5) Screen time.  (6-10) Language-French/Spanish.  (11-15) Language-Arabic/Japanese/Korean/Chinese.  (16-20) Scene text in stylized fonts. (21-25) Coin/Curved text.
}
\label{fig:visuA}
\end{figure}

\begin{figure}[]
\centering
\includegraphics[width=0.85\textwidth]{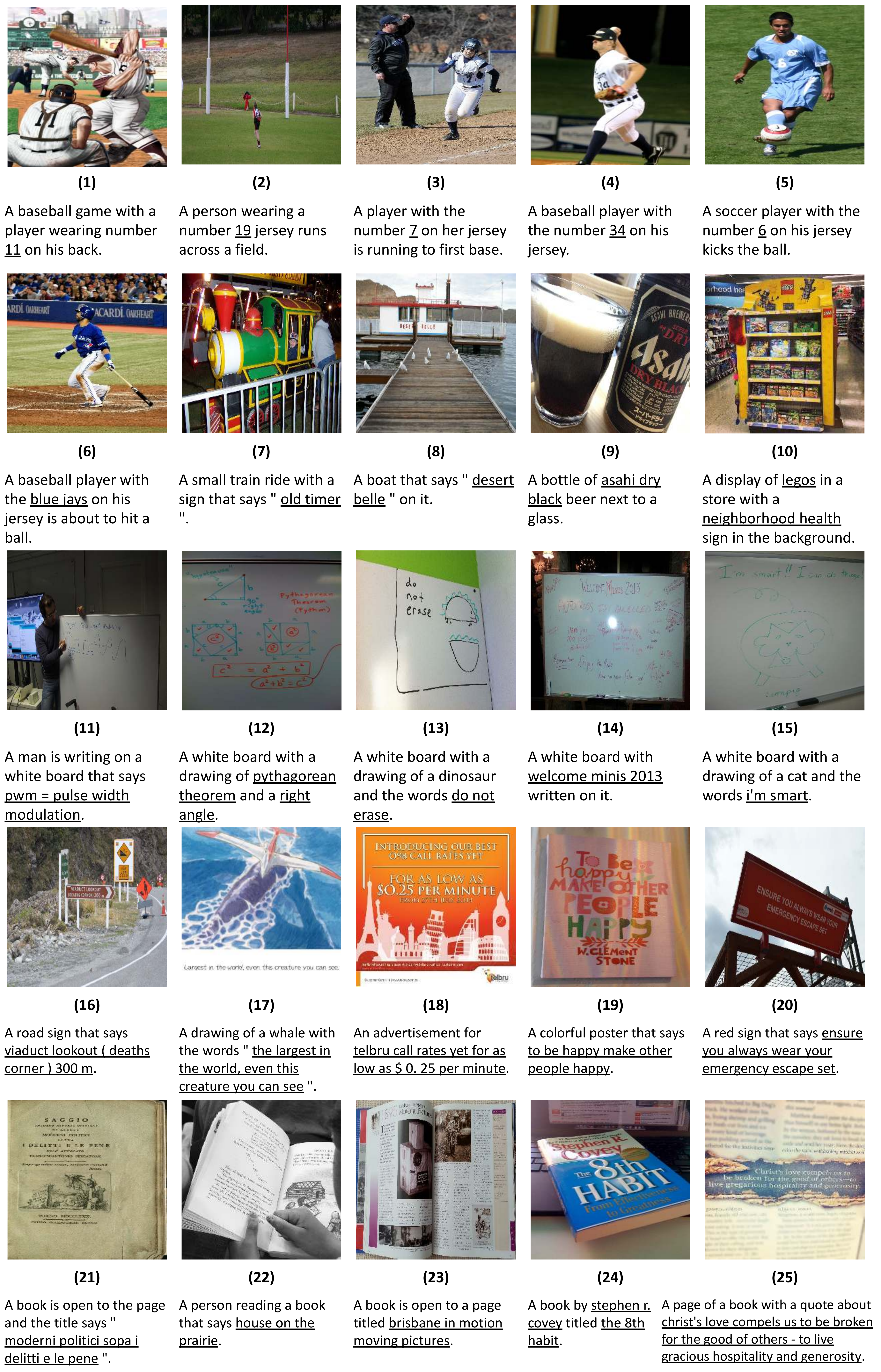}
\caption{
Grouped caption predictions from TextCaps. 
(1-5) Numbers on jerseys. (6-10) Occluded scene text. (11-15) Hand-written scene text. (16-20) Long pieces of scene texts. (21-25) Bookpages.
}
\label{fig:visuB}
\end{figure}

\begin{figure}[]
\centering
\includegraphics[width=0.87\textwidth]{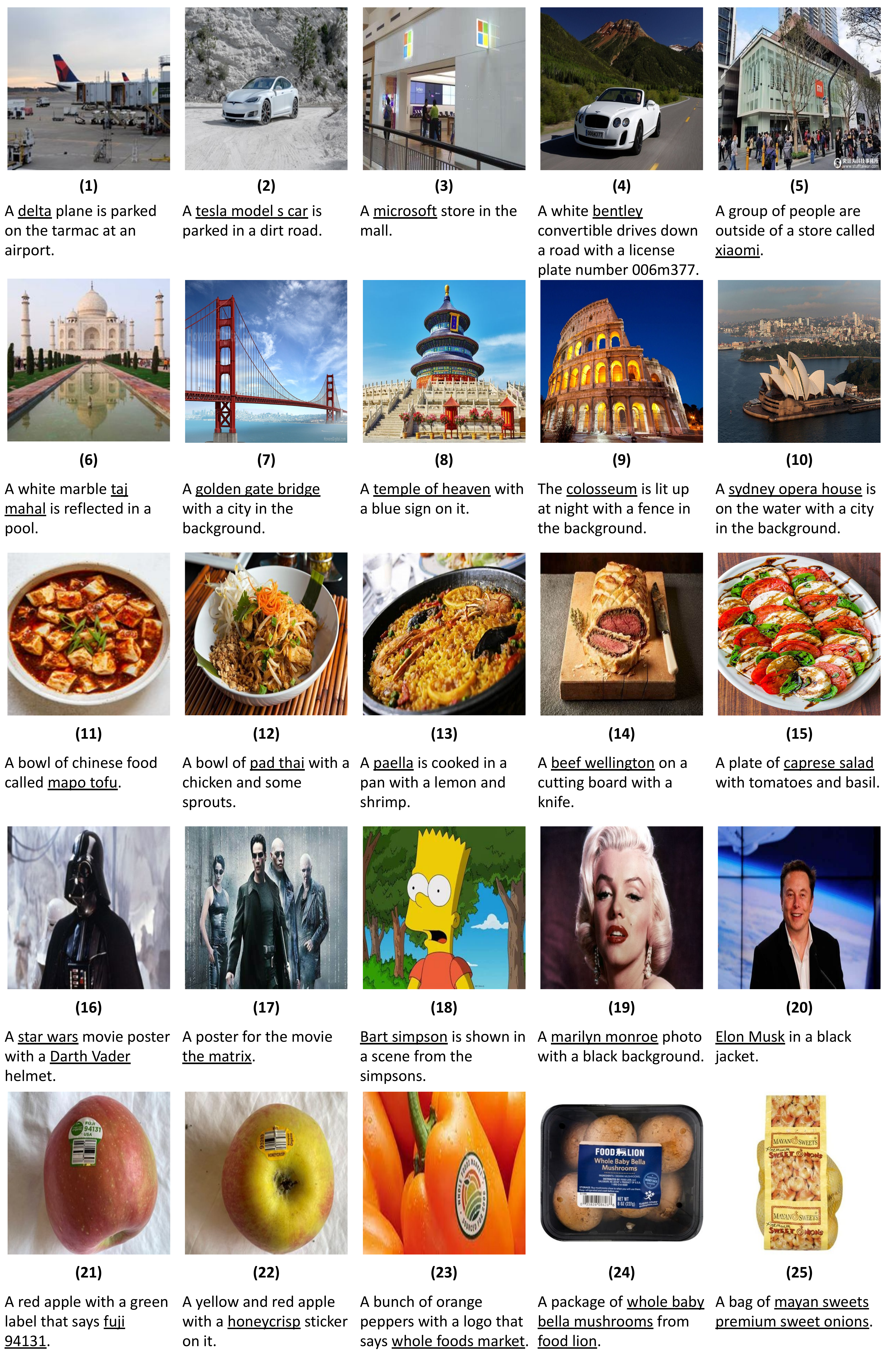}
\caption{
Grouped caption predictions on web images generated by TextCaps-fine-tuned \modelname. 
(1-5) Logos. (6-10) Landmarks. (11-15) Foods. (16-20) Characters and celebrities. (21-25) Products. 
}
\label{fig:visuE}
\end{figure}

\noindent\textbf{TextCaps.}
No SCST~\citep{scst} is performed.
Table~\ref{tab:full_textcaps} shows full results.
Fig.~\ref{fig:TextCaption_random}
shows predictions on random validation images.
We also manually group the predictions according to
different scenarios, as illustrated in
Fig.~\ref{fig:visuA}~and~\ref{fig:visuB}.
In Fig.~\ref{fig:visuA},  (1-5) show examples on which the model  describes the digital time displayed on screens,
which is correct most of the time.
(6-10) provide examples of reading scene text in Latin (Romance) languages such as French and Spanish. 
(11-15) show \modelnamewospace's ability in recognizing scene text in languages such as Arabic, Japanese, Korean, and Chinese. 
(16-20) provide examples of recognizing scene text in stylized fonts. 
As shown in (21-25), \modelname also performs well in reading curved scene text, which is generally considered a challenging case in scene text recognition studies.
In Fig.~\ref{fig:visuB}, samples (1-5) show examples of reading numbers on jerseys. 
As shown in (6-10), we observe that \modelname has a strong ability in inferring occluded scene text, based on both visual and text context information. For example, ``blue jays'' is a baseball team name in sample (6), ``asahi'' is a beer brand in sample (9), and the occluded letter could be letter ``t'' in sample (8). 
(11-15) provide examples of reading hand-written scene text. (16-20) demonstrate \modelnamewospace's ability in reading long pieces of scene texts. \modelname works well in organizing scene text words into a fluent and informative sentence. (21-25) show the challenging case of describing a book page, where the model needs to recognize and select the key information to describe. For example in sample (24), \modelname covers the name and author of the book in the image.

In addition to the scene text captioning ability, 
we observe that the TextCaps-fine-tuned \modelname is knowledgeable and can 
produce diverse and informative captions. 
We group the representative captions in Fig.~\ref{fig:visuE}. Samples (1-5) contain the descriptions of logos, such as ``delta,'' ``tesla,'' ``oneplus,'' \emph{etc.} \modelname also shows the capability of describing landmarks, \emph{e.g.}, ``taj mahal,'' ``golden gate bridge,'' ``temple of heaven,'' ``Colosseum,'' and ``Sydney opera house'' in (6-10). Samples (11-15) show examples on food images, such as ``mapo tofu,'' ``pad thai,'' ``paella,'' ``beef wellington,'' and ``caprese salad.'' (16-20) provide more examples of recognizing movie/cartoon characters and celebrities. Samples (21-25) describe products based on the tag or packaging information.

\begin{figure}
\centering
\small
\renewcommand{\arraystretch}{0.8}
\begin{tabular}{p{\fivewidth\linewidth}@{~~}p{\fivewidth\linewidth}@{~~}p{\fivewidth\linewidth}@{~~}p{\fivewidth\linewidth}@{~~}p{\fivewidth\linewidth}}
\includegraphics[width=1\linewidth,height=1\linewidth]{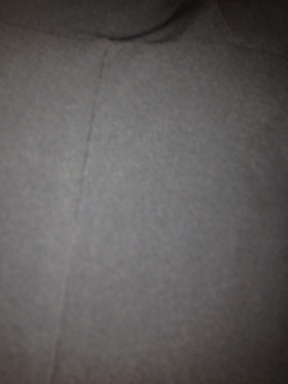} & \includegraphics[width=1\linewidth,height=1\linewidth]{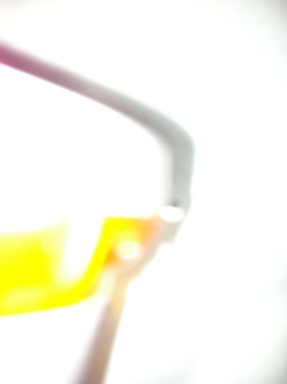} & \includegraphics[width=1\linewidth,height=1\linewidth]{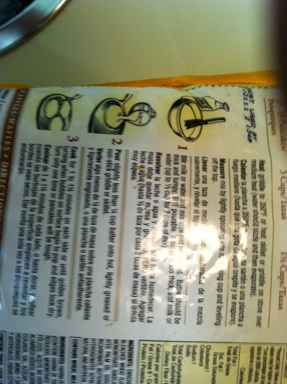} & \includegraphics[width=1\linewidth,height=1\linewidth]{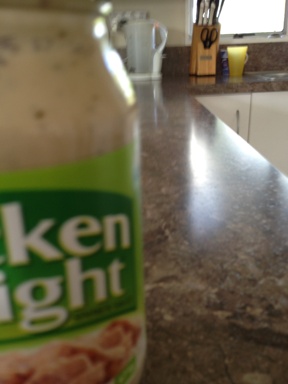} & \includegraphics[width=1\linewidth,height=1\linewidth]{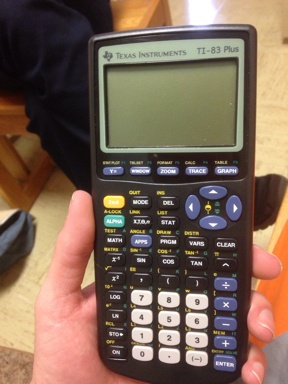}\\
\textbf{Pred}: a close up of a grey piece of fabric with a seam. & \textbf{Pred}: a close up of a yellow object on a white background. & \textbf{Pred}: the back of a package of food with the cooking instructions. & \textbf{Pred}: the front of a jar of chicken light salad dressing on a kitchen counter. & \textbf{Pred}: a hand holding a black calculator with a screen.\\
\includegraphics[width=1\linewidth,height=1\linewidth]{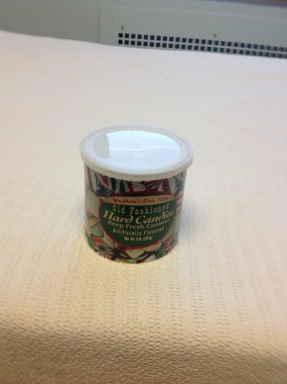} & \includegraphics[width=1\linewidth,height=1\linewidth]{fig/image/TaxVizWizCaption/32019.jpg} & \includegraphics[width=1\linewidth,height=1\linewidth]{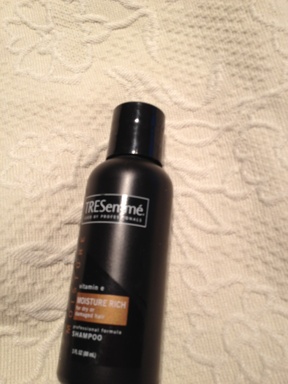} & \includegraphics[width=1\linewidth,height=1\linewidth]{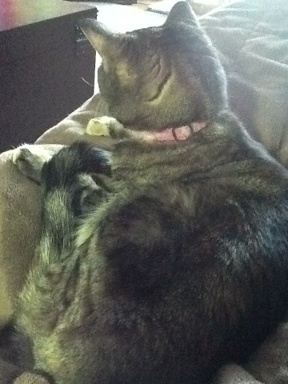} & \includegraphics[width=1\linewidth,height=1\linewidth]{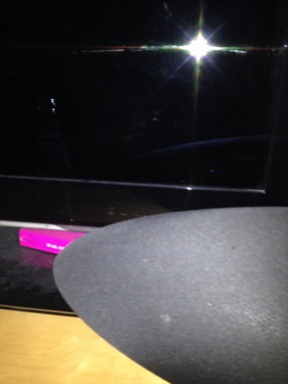}\\
\textbf{Pred}: a container of old fashion hard candies on a table. & \textbf{Pred}: the top of a microwave with buttons on it. & \textbf{Pred}: a black bottle of moisture rich shampoo on a white blanket. & \textbf{Pred}: a grey and black cat with a pink collar laying on a couch. & \textbf{Pred}: a black television screen on a wooden table with a grey object.\\
\includegraphics[width=1\linewidth,height=1\linewidth]{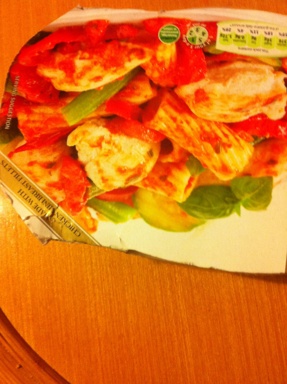} & \includegraphics[width=1\linewidth,height=1\linewidth]{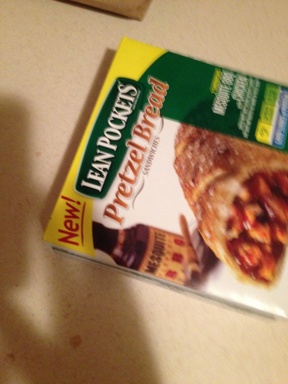} & \includegraphics[width=1\linewidth,height=1\linewidth]{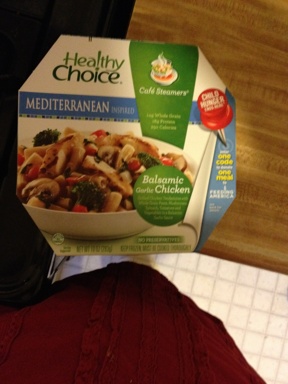} & \includegraphics[width=1\linewidth,height=1\linewidth]{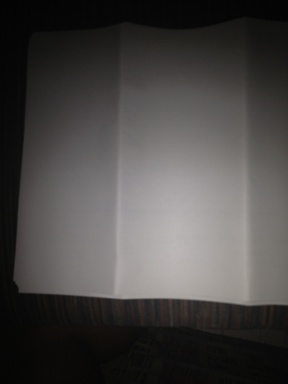} & \includegraphics[width=1\linewidth,height=1\linewidth]{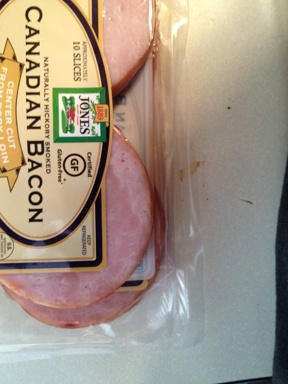}\\
\textbf{Pred}: the top of a box of frozen dinner on a wooden table. & \textbf{Pred}: the top of a box of pretzel bread on a counter. & \textbf{Pred}: the top of a box of healthy choice mediterranean balsamic garlic chicken frozen dinner. & \textbf{Pred}: a blank white piece of paper on a couch. & \textbf{Pred}: the top of a package of canadian bacon.\\
\includegraphics[width=1\linewidth,height=1\linewidth]{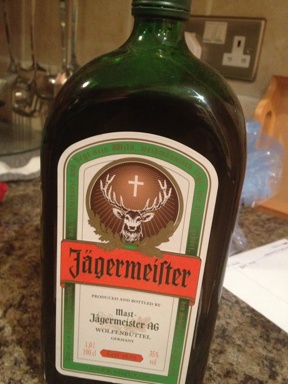} & \includegraphics[width=1\linewidth,height=1\linewidth]{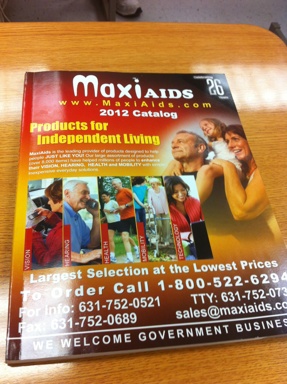} & \includegraphics[width=1\linewidth,height=1\linewidth]{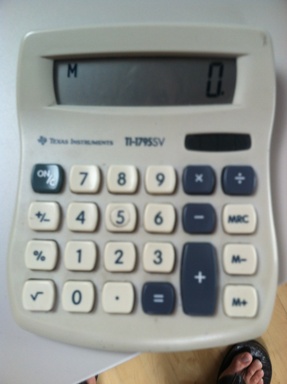} & \includegraphics[width=1\linewidth,height=1\linewidth]{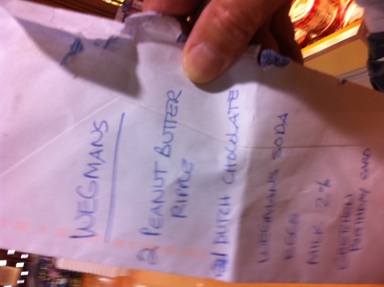} & \includegraphics[width=1\linewidth,height=1\linewidth]{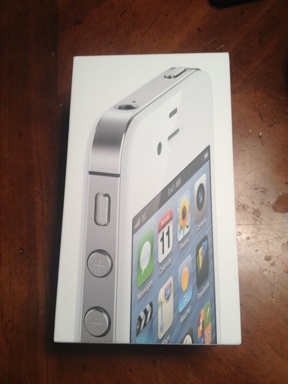}\\
\textbf{Pred}: the top of a green bottle of liquor with a label. & \textbf{Pred}: the front cover of a catalog for 2012 catalog. & \textbf{Pred}: the top of a calculator with white buttons on a table. & \textbf{Pred}: a hand holding a piece of paper with a grocery list. & \textbf{Pred}: the front of a white box for a cell phone.\\
\includegraphics[width=1\linewidth,height=1\linewidth]{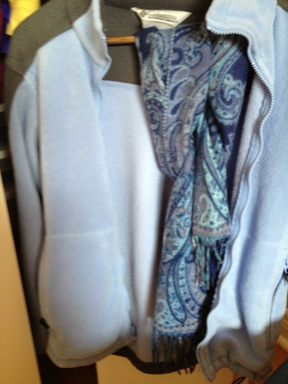} & \includegraphics[width=1\linewidth,height=1\linewidth]{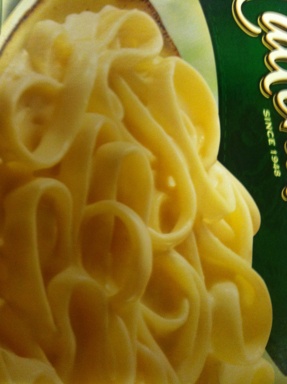} & \includegraphics[width=1\linewidth,height=1\linewidth]{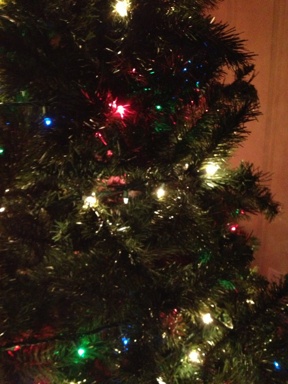} & \includegraphics[width=1\linewidth,height=1\linewidth]{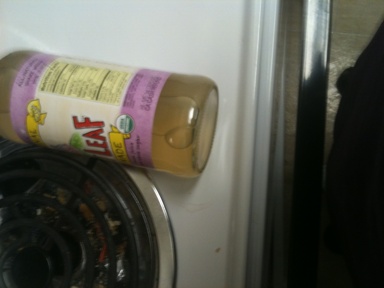} & \includegraphics[width=1\linewidth,height=1\linewidth]{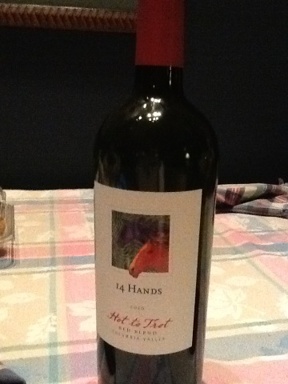}\\
\textbf{Pred}: a blue sweater with a blue scarf hanging on a hanger. & \textbf{Pred}: the top of a box of fettuccine alfredo. & \textbf{Pred}: the top of a christmas tree with lights on it. & \textbf{Pred}: a bottle of organic apple cider tea sitting on top of a stove. & \textbf{Pred}: a bottle of 14 hands red wine on a table.\\
\end{tabular}
\caption{Visualization of our model on random test images of VizWiz-Captions.}
\label{fig:TaxVizWizCaption_random}
\end{figure}

\begin{figure*}[]
\centering
\includegraphics[width=0.87\textwidth]{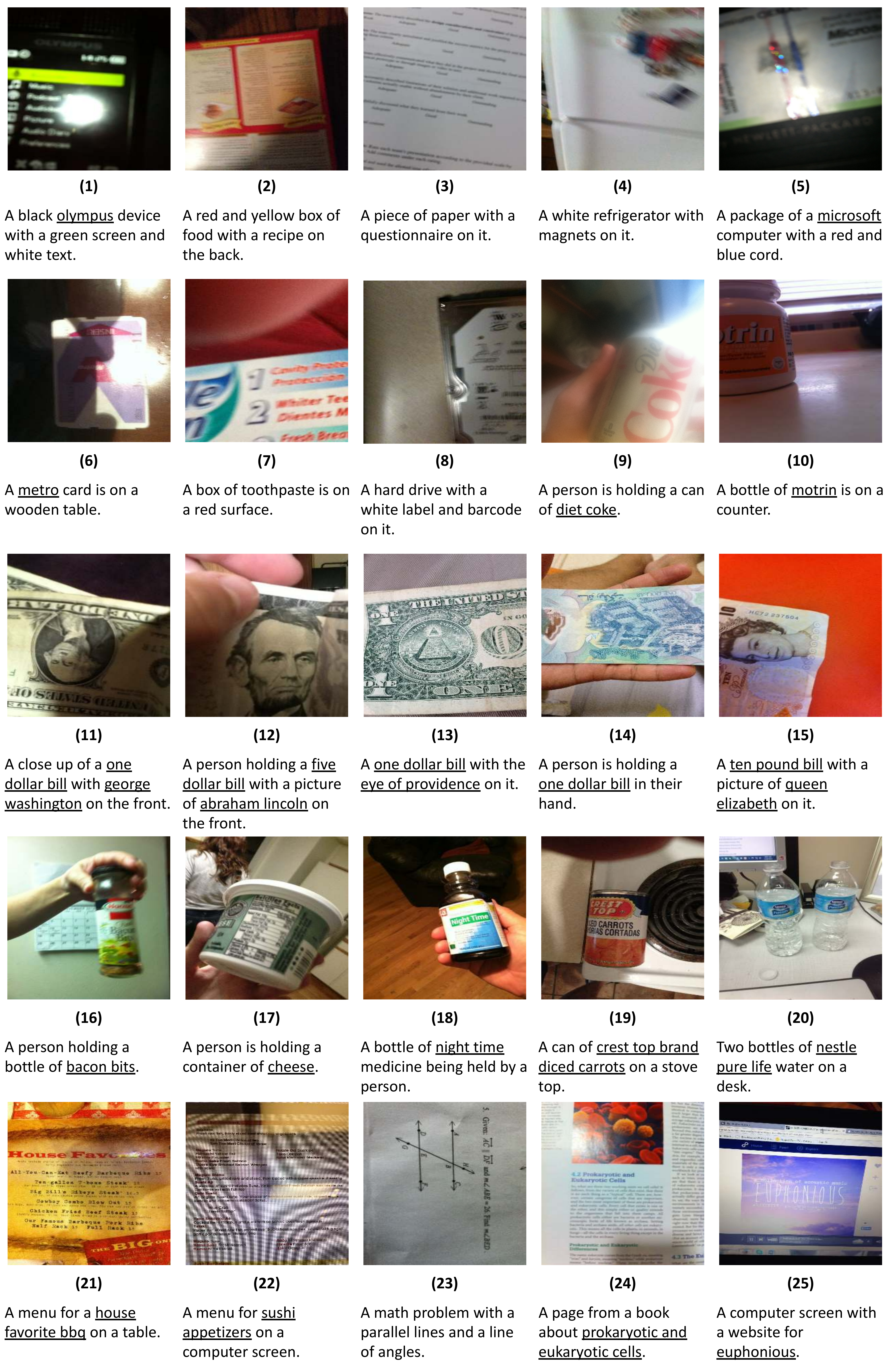}
\caption{Grouped caption predictions from Vizwiz-Captions. 
(1-5) Blurry images. 
(6-10) Low-quality or occluded images. 
(11-15) Banknotes. (16-20) Bottles and cans. (21-25) Menus, pages, and screens.
}
\label{fig:visuC}
\end{figure*}

\noindent\textbf{VizWiz-Captions.}
SCST is performed except \modelnamewospace2, and the full results are shown in Table~\ref{tab:full_vizwiz}. 
Fig.~\ref{fig:TaxVizWizCaption_random} visualizes
the predictions on random test images.
Fig.~\ref{fig:visuC} groups the results 
by different scenarios.
The model can well recognize the banknotes, scene text on bottles/cans, menus, screens, \emph{etc.}, and can better help
vision-impaired people in real use cases.
The first row (1-5) 
of Fig.~\ref{fig:visuC} shows the generated captions on blurry images. 
The second row (6-10) shows images with low image quality or key information partially occluded. For example, \modelname reads the scene text ``metro,'' ``diet coke,'' and ``mortrin'' in samples (6,9,10), and infers the object ``toothpaste'' and ``hard drive'' in samples (7,8). 
Samples (11-15) recognize banknotes in different currencies and denominations. 
(16-20) describe scene text on bottles and cans, thus providing more informative captions such as the ``bacon bits'' in (16) and the ``nestle water'' in (20). \modelname also works well in summarizing menus, pages, and screens, as shown in the bottom row (21-25).

\begin{table}[]
    \centering
    \caption{Results on VizWiz-Captions. Both test-dev and test-std are evaluated on the server.
    $^\#$: winner entry of 2021 VizWiz Grand Challenge\protect\footnotemark.
    B@4: BLEU@4; M: METEOR; R: ROUGE-L; C: CIDEr-D; S: SPICE.
    MTMA: ~\cite{GongZWCZSL21}.
    }
    \label{tab:full_vizwiz}
    \begin{tabular}{l@{~~~~}c@{~~~~}c@{~~~~}c@{~~~~}c@{~~~~}c@{~~~~}c@{~~~~}c@{~~~~}c@{~~~~}c@{~~~~}c}
    \toprule
     \multirow{2}{*}{Method}     &  \multicolumn{5}{c}{test-dev}          & \multicolumn{5}{c}{test-std} \\
                                 \cmidrule(lr){2-6} \cmidrule(lr){7-11}
                                 & B@4  & M    & R    & C     & S         & B@4  & M    & R    & C     & S   \\
    \midrule    
    MTMA$^\#$     & 30.8 & 23.7 & 51.9 & 94.9  & 19.9 & 30.7 & 23.6 & 51.6 & 94.1  & 19.9 \\ \midrule
    \modelnamewospace$_\text{B}$      & 25.1 & 21.7 & 49.4 & 71.5  & 17.8 &  -   &  -   & -    & -     & -    \\
    \modelnamewospace$_\text{L}$      & 29.4 & 23.5 & 50.0 & 96.1  & 20.1      & -    & -    &  -   & -     & -  \\  
    \modelname                        & 33.1 & 25.5 & 53.1 & 113.1 & 22.2      & 33.4 & 25.6 & 53.2 & 114.4 & 22.3 \\ 
    \midrule
    \modelnamewospace2                & \bf36.7 & \bf26.0 & \bf54.6 & \bf119.4 & \bf22.7      & \bf37.1 & \bf26.2 & \bf54.9 & \bf120.8 & \bf22.8 \\ 
    \bottomrule
    \end{tabular}
\end{table}
\footnotetext{https://vizwiz.org/workshops/2021-workshop/}


{
\newpage
\noindent\textbf{Flickr30K.}
Table~\ref{tab:flickr_few_full} shows the full results.
SCST is not applied. 
For the 16/32-shot setting, the batch size is reduced to 16, and the number of
iterations is 100.
}

\begin{table}[h]
    \centering
    \caption{
    Zero/Few/Full-shot evaluation on Flickr30K with Karpathy split.
    }
    \label{tab:flickr_few_full}
    \begin{tabular}{cccccc}
    \toprule
    Shot       &  0   & 16   & 32   & 290 (1\%) & full \\
    \midrule
\cite{ZhouPZHCG20} & - & -   & -    & -         & 68.5  \\
    Flamingo   & 67.2 & 78.9 & 75.4 &  -        &    -  \\
    \midrule
   \modelnamewospace$_B$ & 35.2 & 65.8 & 66.4 &   71.8 & 81.8 \\
   \modelnamewospace$_L$ & 39.2 & 64.4 & 68.5 &  75.4  & 92.4 \\
\modelname     & 49.6 & 78.0 & 80.5 &  86.6     & 98.5 \\
   \modelnamewospace2 & 50.7 & 79.6 & 82.0 & 88.2     & 98.5 \\
\bottomrule
    \end{tabular}
\end{table}

\section{Results on Visual Question Answering}
\begin{figure}
\centering
\small
\begin{tabular}{p{\fivewidth\linewidth}@{~~}p{\fivewidth\linewidth}@{~~}p{\fivewidth\linewidth}@{~~}p{\fivewidth\linewidth}@{~~}p{\fivewidth\linewidth}}
\includegraphics[width=1\linewidth,height=1\linewidth]{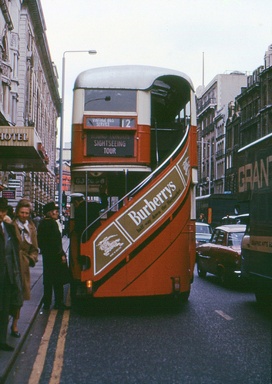} & \includegraphics[width=1\linewidth,height=1\linewidth]{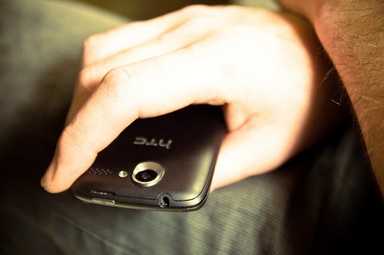} & \includegraphics[width=1\linewidth,height=1\linewidth]{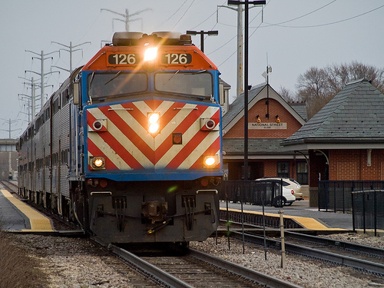} & \includegraphics[width=1\linewidth,height=1\linewidth]{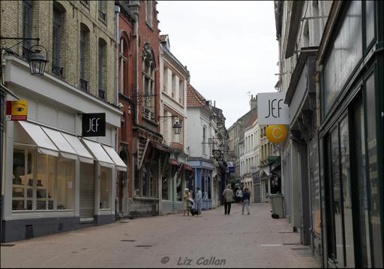} & \includegraphics[width=1\linewidth,height=1\linewidth]{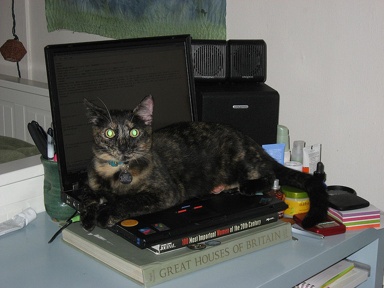}\\
\textbf{Q}: What is the number on the bus? & \textbf{Q}: What is this phone manufacturer's name? & \textbf{Q}: What is the number on the train? & \textbf{Q}: What letters are printed on the white sign? & \textbf{Q}: How many women are featured in the black book\\
{\textbf{Pred}}: 12 & {\textbf{Pred}}: htc & {\textbf{Pred}}: 126 & {\textbf{Pred}}: jef & {\textbf{Pred}}: 100\\
\includegraphics[width=1\linewidth,height=1\linewidth]{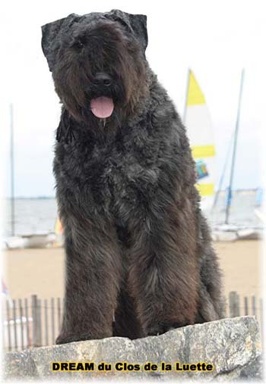} & \includegraphics[width=1\linewidth,height=1\linewidth]{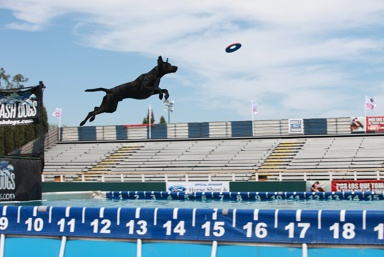} & \includegraphics[width=1\linewidth,height=1\linewidth]{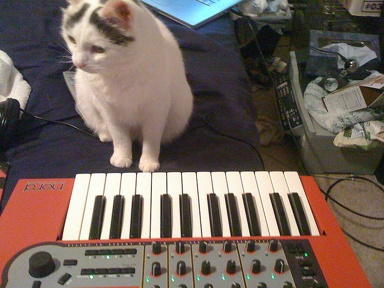} & \includegraphics[width=1\linewidth,height=1\linewidth]{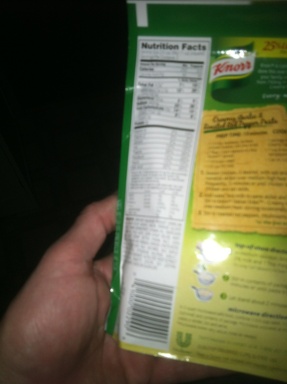} & \includegraphics[width=1\linewidth,height=1\linewidth]{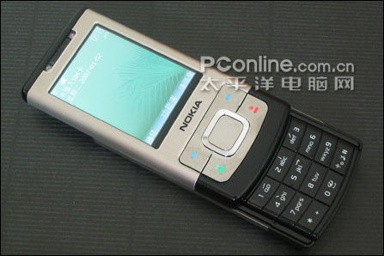}\\
\textbf{Q}: What animal is pictured? & \textbf{Q}: What car company is an official sponsor? & \textbf{Q}: What type of keyboard is pictured? & \textbf{Q}: What brand name is on the bag? & \textbf{Q}: What is the brand name of the mobile phone?\\
{\textbf{Pred}}: dog & {\textbf{Pred}}: ford & {\textbf{Pred}}: nord & {\textbf{Pred}}: knorr & {\textbf{Pred}}: nokia\\
\includegraphics[width=1\linewidth,height=1\linewidth]{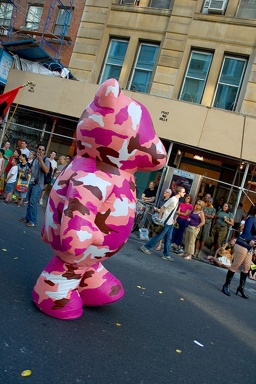} & \includegraphics[width=1\linewidth,height=1\linewidth]{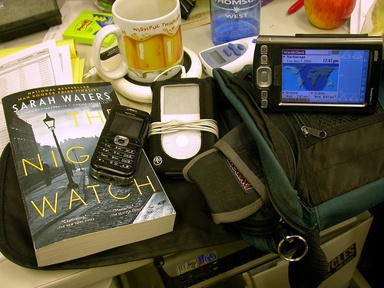} & \includegraphics[width=1\linewidth,height=1\linewidth]{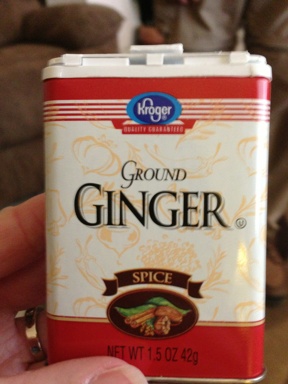} & \includegraphics[width=1\linewidth,height=1\linewidth]{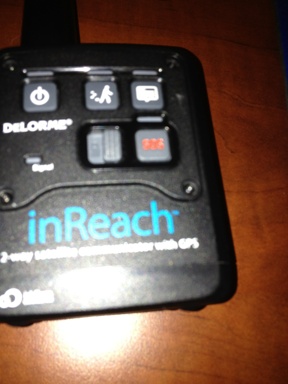} & \includegraphics[width=1\linewidth,height=1\linewidth]{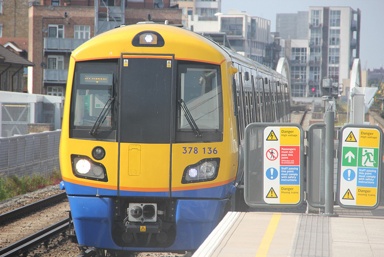}\\
\textbf{Q}: What type of booth is behind the people? & \textbf{Q}: What is the person's first name at the top of the book? & \textbf{Q}: WHICH BRAND IS IT & \textbf{Q}: what is written in blue color? & \textbf{Q}: what are the numbers on train\\
{\textbf{Pred}}: phone & {\textbf{Pred}}: sarah & {\textbf{Pred}}: kroger & {\textbf{Pred}}: inreach & {\textbf{Pred}}: 378 136\\
\includegraphics[width=1\linewidth,height=1\linewidth]{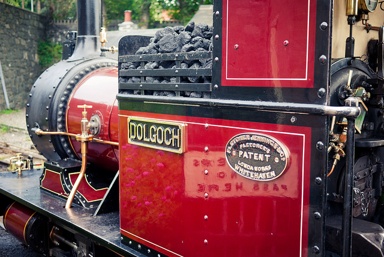} & \includegraphics[width=1\linewidth,height=1\linewidth]{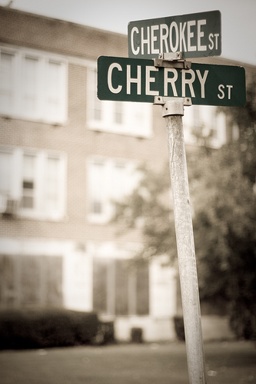} & \includegraphics[width=1\linewidth,height=1\linewidth]{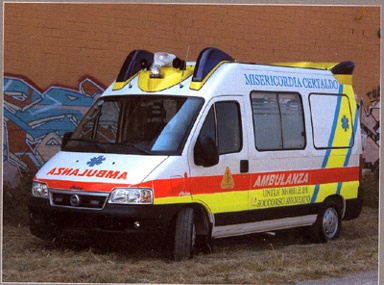} & \includegraphics[width=1\linewidth,height=1\linewidth]{fig/image/TaxSTVQA/good/VisualGenome/1/2357226_jpg.jpg} & \includegraphics[width=1\linewidth,height=1\linewidth]{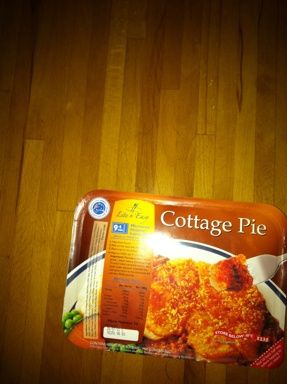}\\
\textbf{Q}: What is the brand of train? & \textbf{Q}: What does the bottom sign say? & \textbf{Q}: What is written in white on the red-orange line on the side of this vehicle? & \textbf{Q}: What is the destination of the white bus? & \textbf{Q}: what food is shown in this picture?\\
{\textbf{Pred}}: dolgoch & {\textbf{Pred}}: cherry st & {\textbf{Pred}}: ambulanza & {\textbf{Pred}}: crosstown & {\textbf{Pred}}: cottage pie\\
\end{tabular}
\caption{Correct predictions on random validation images of ST-VQA.}
\label{fig:stvqa_good}
\end{figure}

\begin{figure}
\centering
\small
\renewcommand{\arraystretch}{0.8}
\begin{tabular}{p{\fivewidth\linewidth}@{~~}p{\fivewidth\linewidth}@{~~}p{\fivewidth\linewidth}@{~~}p{\fivewidth\linewidth}@{~~}p{\fivewidth\linewidth}}
\includegraphics[width=1\linewidth,height=1\linewidth]{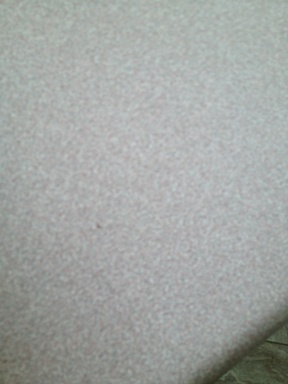} & \includegraphics[width=1\linewidth,height=1\linewidth]{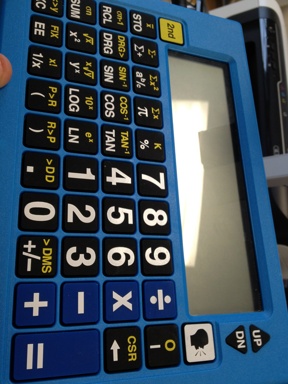} & \includegraphics[width=1\linewidth,height=1\linewidth]{fig/image/TaxVizWizVQA/good/1474.jpg} & \includegraphics[width=1\linewidth,height=1\linewidth]{fig/image/TaxVizWizVQA/good/402.jpg} & \includegraphics[width=1\linewidth,height=1\linewidth]{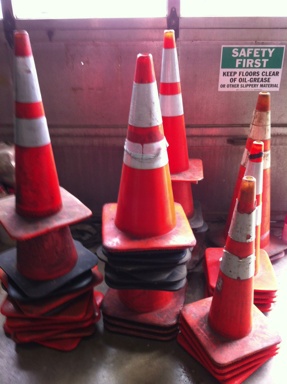}\\
\textbf{Q}: What is this? & \textbf{Q}: What is this? & \textbf{Q}: What is this? & \textbf{Q}: Is this valium? & \textbf{Q}: What are these?\\
{\textbf{Pred}}: unanswerable & {\textbf{Pred}}: calculator & {\textbf{Pred}}: keyboard & {\textbf{Pred}}: unanswerable & {\textbf{Pred}}: traffic cones\\
\includegraphics[width=1\linewidth,height=1\linewidth]{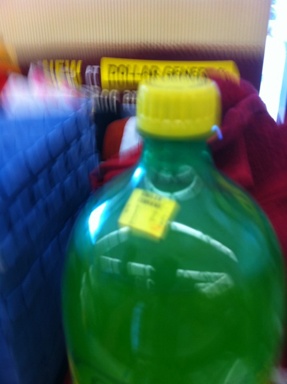} & \includegraphics[width=1\linewidth,height=1\linewidth]{fig/image/TaxVizWizVQA/good/2514.jpg} & \includegraphics[width=1\linewidth,height=1\linewidth]{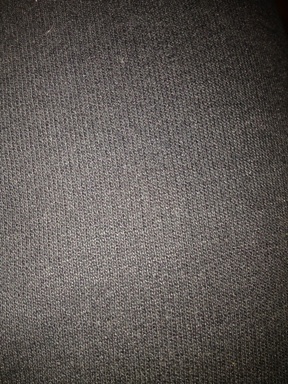} & \includegraphics[width=1\linewidth,height=1\linewidth]{fig/image/TaxVizWizVQA/good/1108.jpg} & \includegraphics[width=1\linewidth,height=1\linewidth]{fig/image/TaxVizWizVQA/good/2401.jpg}\\
\textbf{Q}: In this bottle. & \textbf{Q}: What is this Kane? & \textbf{Q}: What color is this? & \textbf{Q}: Help with question. & \textbf{Q}: What is in this tin?\\
{\textbf{Pred}}: unanswerable & {\textbf{Pred}}: unanswerable & {\textbf{Pred}}: grey & {\textbf{Pred}}: unanswerable & {\textbf{Pred}}: corn\\
\includegraphics[width=1\linewidth,height=1\linewidth]{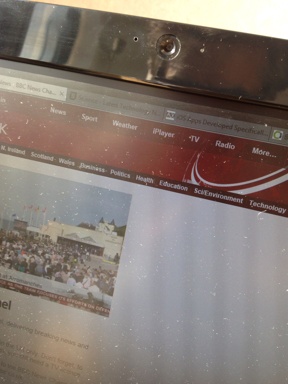} & \includegraphics[width=1\linewidth,height=1\linewidth]{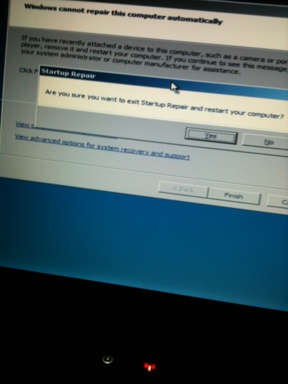} & \includegraphics[width=1\linewidth,height=1\linewidth]{fig/image/TaxVizWizVQA/good/1748.jpg} & \includegraphics[width=1\linewidth,height=1\linewidth]{fig/image/TaxVizWizVQA/good/2669.jpg} & \includegraphics[width=1\linewidth,height=1\linewidth]{fig/image/TaxVizWizVQA/good/1188.jpg}\\
\textbf{Q}: Who is this picture? & \textbf{Q}: What's on the screen? & \textbf{Q}: What about the scenery? & \textbf{Q}: What does my screen say? & \textbf{Q}: What color is the castle?\\
{\textbf{Pred}}: unanswerable & {\textbf{Pred}}: startup repair & {\textbf{Pred}}: unanswerable & {\textbf{Pred}}: unsuitable & {\textbf{Pred}}: unanswerable\\
\includegraphics[width=1\linewidth,height=1\linewidth]{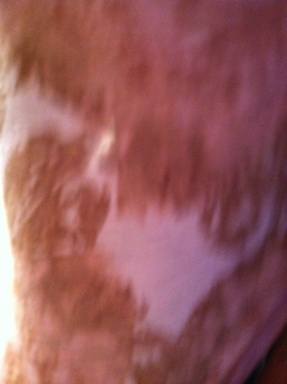} & \includegraphics[width=1\linewidth,height=1\linewidth]{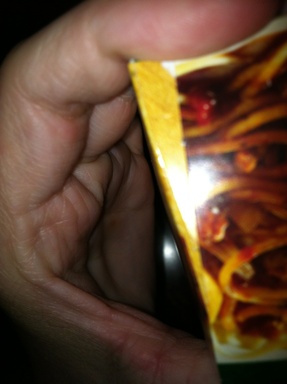} & \includegraphics[width=1\linewidth,height=1\linewidth]{fig/image/TaxVizWizVQA/good/2508.jpg} & \includegraphics[width=1\linewidth,height=1\linewidth]{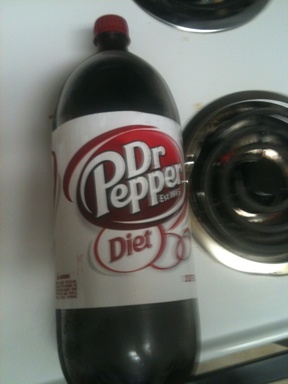} & \includegraphics[width=1\linewidth,height=1\linewidth]{fig/image/TaxVizWizVQA/good/945.jpg}\\
\textbf{Q}: What color is this shirt? & \textbf{Q}: What is the expiration date? & \textbf{Q}: Which country does this belong to? & \textbf{Q}: Can you tell me what flavor this is?  & \textbf{Q}: what company is this my hotspot from?\\
{\textbf{Pred}}: unsuitable & {\textbf{Pred}}: unanswerable & {\textbf{Pred}}: unanswerable & {\textbf{Pred}}: diet dr pepper & {\textbf{Pred}}: verizon\\
\includegraphics[width=1\linewidth,height=1\linewidth]{fig/image/TaxVizWizVQA/good/2305.jpg} & \includegraphics[width=1\linewidth,height=1\linewidth]{fig/image/TaxVizWizVQA/good/2687.jpg} & \includegraphics[width=1\linewidth,height=1\linewidth]{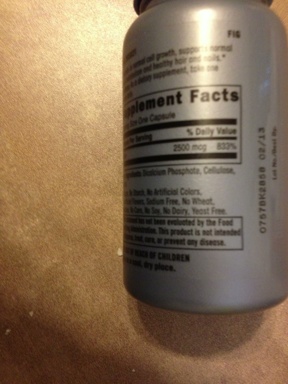} & \includegraphics[width=1\linewidth,height=1\linewidth]{fig/image/TaxVizWizVQA/good/3109.jpg} & \includegraphics[width=1\linewidth,height=1\linewidth]{fig/image/TaxVizWizVQA/good/1532.jpg}\\
\textbf{Q}: What flavor of take five gum is this? & \textbf{Q}: Could you tell me what brand this is? Thank you. & \textbf{Q}: Is the label showing? And what is in this bottle? & \textbf{Q}: How much milk and water do I add to my hamburger helper? & \textbf{Q}: Who is the fifth president and who is on the six dollar bill?\\
{\textbf{Pred}}: unanswerable & {\textbf{Pred}}: grace & {\textbf{Pred}}: unanswerable & {\textbf{Pred}}: unsuitable & {\textbf{Pred}}: abraham lincoln\\
\end{tabular}
\caption{Visualization of correct predictions for the validation set on VizWiz-VQA.}
\label{fig:vis_TaxVizWizVQA_good}
\end{figure}

\begin{figure}
\centering
\small
\begin{tabular}{p{\fivewidth\linewidth}@{~~}p{\fivewidth\linewidth}@{~~}p{\fivewidth\linewidth}@{~~}p{\fivewidth\linewidth}@{~~}p{\fivewidth\linewidth}}
\includegraphics[width=1\linewidth,height=1\linewidth]{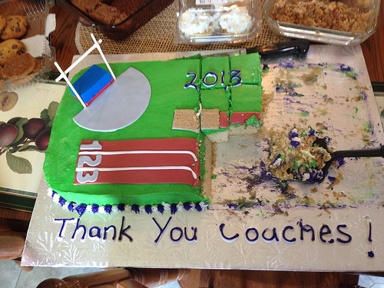} & \includegraphics[width=1\linewidth,height=1\linewidth]{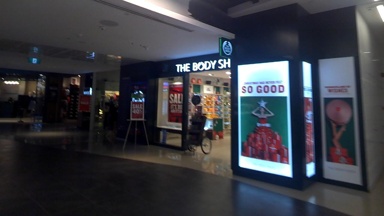} & \includegraphics[width=1\linewidth,height=1\linewidth]{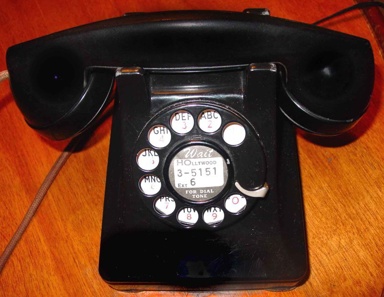} & \includegraphics[width=1\linewidth,height=1\linewidth]{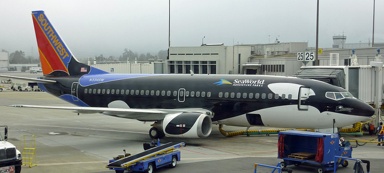} & \includegraphics[width=1\linewidth,height=1\linewidth]{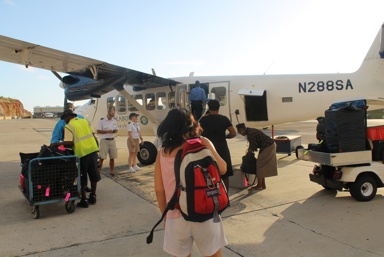}\\
\textbf{Q}: what are the numbers on the track on the cake? & \textbf{Q}: What sale is advertised for the store in the image? & \textbf{Q}: How does one get a dial tone? & \textbf{Q}: What numbers are on the plane? & \textbf{Q}: What number is the plane?\\
{\textbf{Pred}}: 13 & {\textbf{Pred}}: sale & {\textbf{Pred}}: 6 & {\textbf{Pred}}: 25 & {\textbf{Pred}}: n2889sa\\
\textbf{GT}: 123 & \textbf{GT}: 40\% & \textbf{GT}: wait & \textbf{GT}: n334sw & \textbf{GT}: n288sa\\
\includegraphics[width=1\linewidth,height=1\linewidth]{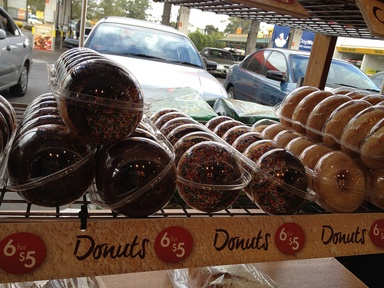} & \includegraphics[width=1\linewidth,height=1\linewidth]{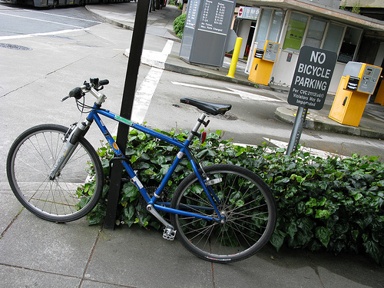} & \includegraphics[width=1\linewidth,height=1\linewidth]{fig/image/TaxSTVQA/bad/VisualGenome/1/2344298_jpg.jpg} & \includegraphics[width=1\linewidth,height=1\linewidth]{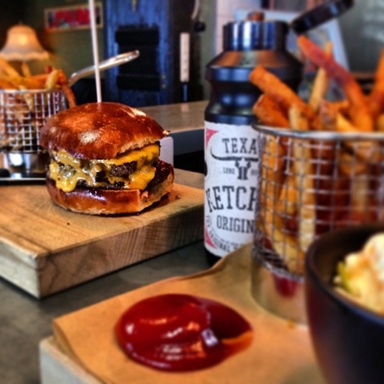} & \includegraphics[width=1\linewidth,height=1\linewidth]{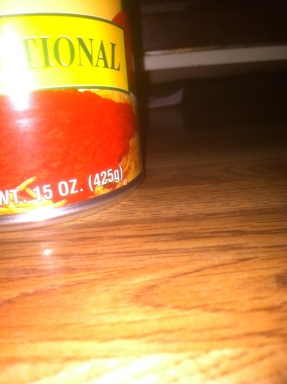}\\
\textbf{Q}: What can you get 6 of for \$5? & \textbf{Q}: What word is on the third line of the sign? & \textbf{Q}: How many cups in 3 gallons? & \textbf{Q}: What flavor is the ketchup? & \textbf{Q}: What is the weight in ounces?\\
{\textbf{Pred}}: \$ 5 & {\textbf{Pred}}: bicycle & {\textbf{Pred}}: 48 & {\textbf{Pred}}: texas & {\textbf{Pred}}: 425g\\
\textbf{GT}: donuts & \textbf{GT}: parking & \textbf{GT}: 48 cups & \textbf{GT}: original & \textbf{GT}: 15, 15 oz.\\
\includegraphics[width=1\linewidth,height=1\linewidth]{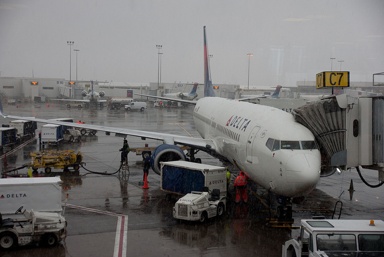} & \includegraphics[width=1\linewidth,height=1\linewidth]{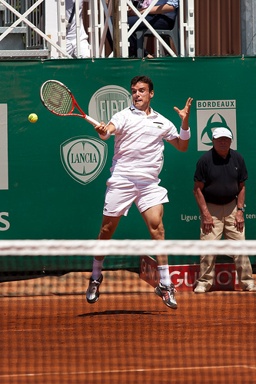} & \includegraphics[width=1\linewidth,height=1\linewidth]{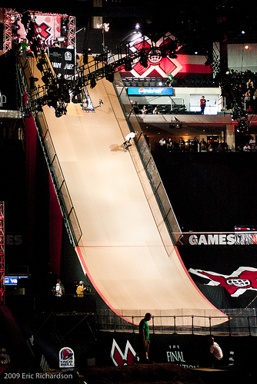} & \includegraphics[width=1\linewidth,height=1\linewidth]{fig/image/TaxSTVQA/bad/VisualGenome/2/2409731_jpg.jpg} & \includegraphics[width=1\linewidth,height=1\linewidth]{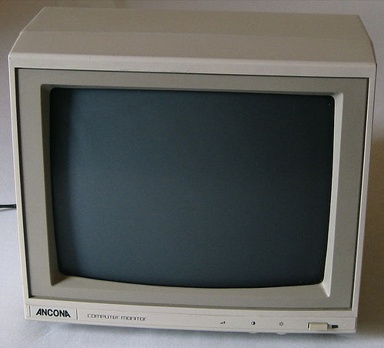}\\
\textbf{Q}: What airline and gate number? & \textbf{Q}: Which word is shown above the man with the white hat? & \textbf{Q}: What restaurant is advertised at the bottom of this picture? & \textbf{Q}: WHAT IS WRITTEN ON THE WALL? & \textbf{Q}: WHAT IS THAT\\
{\textbf{Pred}}: 7 & {\textbf{Pred}}: lancia & {\textbf{Pred}}: games & {\textbf{Pred}}: allews & {\textbf{Pred}}: ancona computer monitor\\
\textbf{GT}: delta c7 & \textbf{GT}: bordeaux & \textbf{GT}: taco bell & \textbf{GT}: dallus, allus & \textbf{GT}: computer monitor\\
\includegraphics[width=1\linewidth,height=1\linewidth]{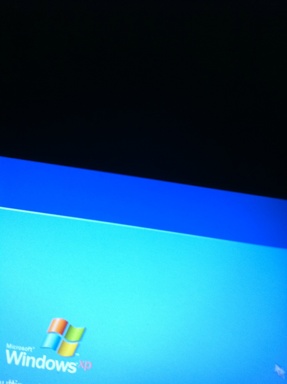} & \includegraphics[width=1\linewidth,height=1\linewidth]{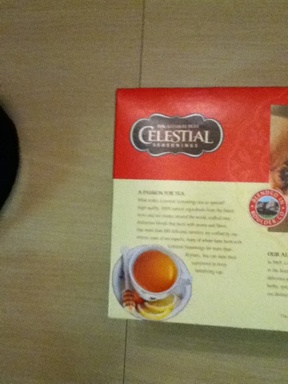} & \includegraphics[width=1\linewidth,height=1\linewidth]{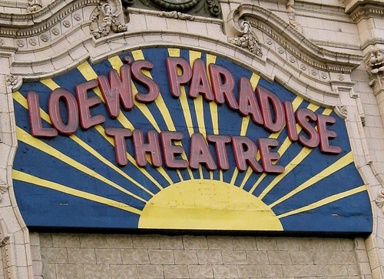} & \includegraphics[width=1\linewidth,height=1\linewidth]{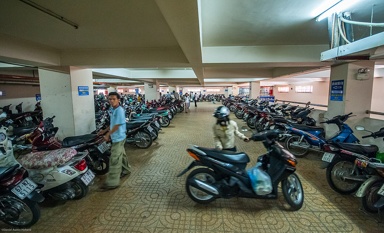} & \includegraphics[width=1\linewidth,height=1\linewidth]{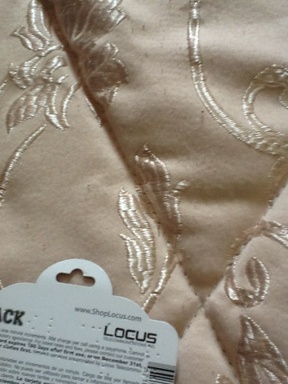}\\
\textbf{Q}: What is written in this picture? & \textbf{Q}: What is the Brand name? & \textbf{Q}: What is the name displayed on the board? & \textbf{Q}: What are the license numbers on white motor bike to the left? & \textbf{Q}: Where could this product be purchased online?\\
{\textbf{Pred}}: windows xp & {\textbf{Pred}}: celestial & {\textbf{Pred}}: loews theatre & {\textbf{Pred}}: 61 - 12 & {\textbf{Pred}}: locus\\
\textbf{GT}: microsoft windows & \textbf{GT}: celestial seasonings & \textbf{GT}: loew's paradise theatre & \textbf{GT}: 67-n9 67 1024 1024, 67-n9 1024 & \textbf{GT}: www.shoplocus.com, shoplocus.com\\
\end{tabular}
\caption{Incorrect predictions on random validation images of ST-VQA.}
\label{fig:stvqa_bad}
\end{figure}

\begin{figure}
\centering
\small
\renewcommand{\arraystretch}{0.8}
\begin{tabular}{p{\fivewidth\linewidth}@{~~}p{\fivewidth\linewidth}@{~~}p{\fivewidth\linewidth}@{~~}p{\fivewidth\linewidth}@{~~}p{\fivewidth\linewidth}}
\includegraphics[width=1\linewidth,height=1\linewidth]{fig/image/TaxVizWizVQA/bad/3721.jpg} & \includegraphics[width=1\linewidth,height=1\linewidth]{fig/image/TaxVizWizVQA/bad/787.jpg} & \includegraphics[width=1\linewidth,height=1\linewidth]{fig/image/TaxVizWizVQA/bad/4154.jpg} & \includegraphics[width=1\linewidth,height=1\linewidth]{fig/image/TaxVizWizVQA/bad/56.jpg} & \includegraphics[width=1\linewidth,height=1\linewidth]{fig/image/TaxVizWizVQA/bad/1362.jpg}\\
\textbf{Q}: What scent is this? & \textbf{Q}: What's on the screen? & \textbf{Q}: What is this question? & \textbf{Q}: Specials on this menu. & \textbf{Q}: How much fat does it contain?\\
{\textbf{Pred}}: lavender & {\textbf{Pred}}: unsuitable & {\textbf{Pred}}: cleaning brush & {\textbf{Pred}}: beef & {\textbf{Pred}}: 180\\
\textbf{GT}: french lavendar, french lavender & \textbf{GT}: web links, web search results, unsuitable image, search results, words, search & \textbf{GT}: pipe cleaner, i dont know, dish cleaner, lint brush, unanswerable, brush, bottle cleaner & \textbf{GT}: shrimp, unsuitable image, dont see any, unanswerable, beef seafood & \textbf{GT}: 12\%, g, 8 grams, 8 g, g per serving 3 servings per container\\
\includegraphics[width=1\linewidth,height=1\linewidth]{fig/image/TaxVizWizVQA/bad/3561.jpg} & \includegraphics[width=1\linewidth,height=1\linewidth]{fig/image/TaxVizWizVQA/bad/1947.jpg} & \includegraphics[width=1\linewidth,height=1\linewidth]{fig/image/TaxVizWizVQA/bad/650.jpg} & \includegraphics[width=1\linewidth,height=1\linewidth]{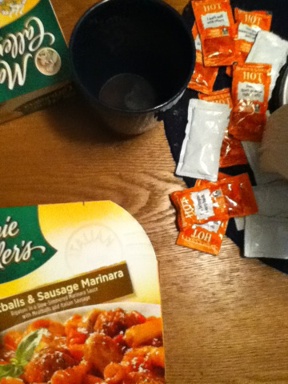} & \includegraphics[width=1\linewidth,height=1\linewidth]{fig/image/TaxVizWizVQA/bad/1969.jpg}\\
\textbf{Q}: What kind of can food is this? & \textbf{Q}: What is the title of the book? & \textbf{Q}: what can i hear on the screen. & \textbf{Q}: What kind of dinner bowl is this? & \textbf{Q}: Please just describe what this is.\\
{\textbf{Pred}}: unsuitable & {\textbf{Pred}}: unanswerable & {\textbf{Pred}}: windows logo & {\textbf{Pred}}: unanswerable & {\textbf{Pred}}: candy\\
\textbf{GT}: baby corn, vegetables, bamboo shoots, salmon, soup, noodles, unanswerable & \textbf{GT}: auf die lange, auf die lange kommt es, auf die länge kommt es & \textbf{GT}: nothing, desktop screen, computer screen to login, just shows normal desk top home screen, unanswerable, windows desktop & \textbf{GT}: meatballs adn sausage marinara, meatballs sausage marinara, food item, cup, ceramic & \textbf{GT}: candy bag, unsuitable, package, candy in clear plastic bag, bag candy, bag candy setting on kitchen table\\
\end{tabular}
\caption{Visualization of incorrect predictions for the validation set on VizWiz-VQA.}
\label{fig:vis_TaxVizWizVQA_bad}
\end{figure}

Except on VizWiz-QA, the number of fine-tuning epochs is 20 and the learning rate is $1e^{-5}$.
On VizWiz-QA, the number of epochs is 40 and the learning rate is $2e^{-5}$.
The input size is 384 and 576 for intermediate fine-tuning and the 
final fine-tuning, respectively.
No intermediate fine-tuning is conducted for
\modelnamewospace$_\text{B}$
and \modelnamewospace$_\text{L}$.
Full results are shown in Table~\ref{tab:full_vqa}.
Fig.~\ref{fig:vis_TaxVizWizVQA_good} 
and Fig.~\ref{fig:stvqa_good}
show correct prediction on randomly selected images of VizWiz-VQA and 
ST-VQA, respectively.
Fig.~\ref{fig:vis_TaxVizWizVQA_bad} and 
Fig.~\ref{fig:stvqa_bad} show the randomly selected incorrect predictions.

\begin{table}[]
    \centering
    \small
    \caption{Results on visual question answering. 
    (a): for VQAv2, approaches are divided according to whether the answer vocabulary is pre-defined (Closed) or not (Open) during inference. The model with closed vocabulary can be 
        a classification model or generation model with constrained outputs, \emph{e.g.},~\cite{abs-2202-03052,abs-2201-12086}.
    (b): for TextVQA, Mia$^\#$ is the winner entry of TextVQA Challenge 2021 with a fine-tuned  T5-3B~\citep{RaffelSRLNMZLL20} model.
    (c): $^{\#\#}$: winner entry of 2021 VizWiz Grand Challenge Workshop.
    ALBEF: ~\cite{li2021align},
    BLOCK+CNN+W2V: ~\cite{mishra2019ocr},
    BLIP: ~\cite{abs-2201-12086},
    CLIP-ViL: ~\cite{abs-2107-06383},
    CoCa: ~\cite{yu2022coca},
    Florence: ~\cite{abs-2111-11432},
    Flamingo: ~\cite{alayrac2022flamingo},
    LaAP-Net: ~\cite{HanHH20},
    LaTr: ~\cite{biten2022latr},
    mPlug: ~\cite{li2022mplug},
    M4C: ~\cite{HuSDR20},
    METER: ~\cite{abs-2111-02387},
    Mia: ~\cite{abs-2106-15332},
    OSCAR: ~\cite{Li0LZHZWH0WCG20},
    OFA: ~\cite{abs-2202-03052},
    PixelBERT: ~\cite{abs-2004-00849},
    UFO: ~\cite{ufo},
    UNITER: ~\cite{ChenLYK0G0020},
    UNIMO: ~\cite{li2020unimo},
    Visual Parsing: ~\cite{abs-2106-13488},
    VILLA: ~\cite{Gan0LZ0020},
    VinVL: ~\cite{abs-2101-00529},
    SA-M4C: ~\cite{KantBASPLA20},
    SMA: ~\cite{abs-2006-00753},
    SimVLM: ~\cite{wang2021simvlm},
    TAP: ~\cite{abs-2012-04638}.
    }
    \label{tab:full_vqa}
    \begin{tabular}{cc}
        \begin{tabular}{c}
                    \begin{tabular}{llcc} \toprule
                        Vocabulary                             &  Model                                 & test-dev            & test-std \\ \midrule
                                                               & OSCAR           & 73.61               & 73.82 \\
                        \multirow{11}{*}{Closed}               & UNITER            & 73.82               & 74.02 \\
                                                               & Visual Parsing   & 74.00               & 74.17 \\
                                                              & PixelBERT         & 74.45               & 74.55 \\
                                                               & VILLA                & 74.69               & 74.87 \\
                                                               & UNIMO               & 75.06               & 75.27 \\
                                                               & ALBEF               & 75.84               & 76.04 \\
                                                               & VinVL            & 76.52               & 76.60 \\
                                                               & UFO                         & 76.64               & 76.76 \\
                                                               & CLIP-ViL         & 76.48               & 76.70 \\
                                                               & METER            & 77.68               & 77.64 \\
                                                               & BLIP             & 78.25               & 78.32 \\
                                                               & OFA              & 79.87               & 80.02  \\ 
                                                               & SimVLM           & 80.03               & 80.34 \\
                                                               & Florence         & 80.16               & 80.36 \\ 
                                                               & mPlug               & 81.27               & 81.26 \\
                                                               & CoCa                 & \bf82.3                & \bf82.3  \\
                                                               \midrule
                         \multirow{4}{*}{Open}                 & Flamingo (80B)    & \bf82.0                & \bf82.1  \\
                                                               \cmidrule(lr){2-4}
                                                               & \modelnamewospace$_\text{B}$ (0.1B)           & 72.72               &   -         \\ 
                                                               & \modelnamewospace$_\text{L}$ (0.3B)          & 75.51               &   -        \\ 
                                                               & \modelname (0.7B)                            & 78.56               &   78.81    \\ 
                                                               \cmidrule(lr){2-4}
                                                               & \modelnamewospace2 (5.1B)  & 81.74 & 81.92 \\ 
                        \bottomrule
                    \end{tabular}
              \\
              (a) VQAv2
        \end{tabular}
    &
        \begin{tabular}{@{}c@{}}
                    \begin{tabular}{lcc}
                    \toprule
                    Model                          &  validation & test       \\
                    \midrule
                    M4C             & 40.55       & 40.46      \\
                    LaAP-Net        & 41.02       & 41.41      \\
                    SA-M4C     & 45.4        & 44.6         \\
                    SMA      & 44.58       & 45.51    \\
                    TAP      & 54.71       & 53.97     \\
               Flamingo & 57.1       & 54.1      \\
               LaTr           & 61.05      & 61.60  \\
                    Mia$^\#$ &   -         & \bf73.67       \\
                    \midrule
                    \modelnamewospace$_\text{B}$   & 18.81       &  -         \\ 
                    \modelnamewospace$_\text{L}$   & 37.47       &  -         \\ 
                    \modelname                     & 59.93       & 59.75     \\ 
                    \midrule
                    \modelnamewospace2             & 68.38        & 67.27    \\ 
                    \bottomrule
                    \end{tabular}
            \\
            (b) TextVQA
             \\
                        \begin{tabular}{lcc} \toprule
                            Model                             & test-dev & test \\ \midrule
                            \cite{vizwiz2021winner}$^{\#\#}$      & 61.8     & 60.6 \\
                        Flamingo   & 65.7     &  65.4 \\
                        \midrule
                            \modelnamewospace$_\text{B}$      & 54.6     &  -       \\ 
                            \modelnamewospace$_\text{L}$      & 62.5     &  -       \\ 
                            \modelname                        & 68.0  & 67.5 \\ 
                            \midrule
                            \modelnamewospace2          & \bf70.97    & \bf70.1 \\ 
                            \bottomrule
                        \end{tabular}
                        \\
                        (c) VizWiz-QA
        \end{tabular}
     \\
     \\
            \begin{tabular}{lccc} \toprule
                Model                           & Val Acc. & Val ANLS & Test ANLS\\
                \hline
                M4C               & 38.1     & 47.2     & 46.2 \\
                LaAP-Net          & 39.7     & 49.7     & 48.5 \\
                SA-M4C       & 42.2     & 51.2     & 50.4 \\
                TAP        & 50.8     & 59.8     & 59.7 \\     
                LaTr        & 61.64    & 70.2     & 69.6 \\ 
                \midrule
                \modelnamewospace$_\text{B}$    &  14.7   & 20.7       &    -         \\  
                \modelnamewospace$_\text{L}$    &  32.3   & 44.6       &    -         \\  
                \modelname                      & 59.2      & 69.1     & 69.6     \\  
                \midrule
                \modelnamewospace2              & \bf66.6   & \bf75.1  & \bf75.8       \\ 
                \bottomrule
            \end{tabular}
        &
            \begin{tabular}{@{}lcc@{}} \toprule
                model                            & val          & test \\ \midrule
             BLOCK+CNN+W2V  & -            & 48.3 \\
                M4C               & 63.5         & 63.9 \\
                LaAP-Net          & 63.8         & 64.1 \\
                LaTr        & 67.5         & 67.9 \\
                \midrule
                \modelnamewospace$_\text{B}$     & 57.3         &  57.5  \\  
                \modelnamewospace$_\text{L}$     & 62.4         & 62.9   \\  
                \modelname                       & 67.8      & 68.1 \\  
                \midrule
                \modelnamewospace2               & \bf69.9     & \bf70.3 \\ 
                \bottomrule 
            \end{tabular}
        \\
        (d) ST-VQA & (e) OCR-VQA
    \end{tabular}
\end{table}



\begin{table}[!]
\centering
\small
\caption{Results on video captioning. 
$^E$: model ensemble; 
$^T$: with the subtitle as additional input.
YouCook2 is on the validation set. 
ActBERT: ~\cite{zhu2020actbert},
CLIP4Caption++: ~\cite{abs-2110-05204},
Flamingo: ~\cite{alayrac2022flamingo},
HERO: ~\cite{li2020hero},
MMT: ~\cite{lei2020tvr},
MV-GPT: ~\cite{abs-2201-08264},
SibNet: ~\cite{liu2020sibnet},
OA-BTG: ~\cite{zhang2019object},
OpenBook: ~\cite{zhang2021open},
ORG-TRL: ~\cite{Zhang_2020_CVPR},
PMI-CAP: ~\cite{chen2020learning},
POS+CG: ~\cite{wang2019controllable},
UniVL: ~\cite{Luo2020UniVL},
VaTeX: ~\cite{wang2019vatex},
VALUE: ~\cite{VALUE},
VideoBERT: ~\cite{sun2019videobert},
Support-set: ~\cite{patrick2020support},
STG-KD: ~\cite{pan2020spatio},
SwinBERT: ~\cite{abs-2111-13196},
X-L.+T.: ~\cite{zhu2019vatex}.
}
\label{tab:full_video_caption}
\begin{tabular}{c}
        \begin{tabular}{@{}c@{~}c@{~}c@{}}
            \begin{tabular}{c}
                \begin{tabular}{@{}l@{~}c@{~}c@{~}c@{~}c@{}} \toprule
                    Method                              &B@4    & M    & R    & C     \\ \midrule
                    SibNet         & 54.2 & 34.8 & 71.7 & 88.2  \\
                    POS+CG  & 52.5 & 34.1 & 71.3 & 88.7  \\
                    OA-BTG       & 56.9 & 36.2 & -    & 90.6  \\
                    STG-KD         & 52.2 & 36.9 & 73.9 & 93.0  \\
                    PMI-CAP     & 54.6 & 36.4 & -    & 95.1  \\
                    ORG-TRL      & 54.3 & 36.4 & 73.9 & 95.2  \\
                    SwinBERT      & 58.2 & 41.3 & 77.5 & 120.6 \\ \midrule
                    \modelnamewospace$_\text{B}$        & 69.3 & 44.5 & 81.4 & 142.6 \\  
                    \modelnamewospace$_\text{L}$        & 75.8 & 48.7 & 85.5 & 162.9 \\  
                    \modelname                          & 79.5 & 51.1 & 87.3 & 180.2 \\  
                    \midrule
                    \modelnamewospace2                  & \bf82.2 & \bf52.3 & \bf88.7 & \bf 185.4 \\  
                    \bottomrule
                \end{tabular}
                \\
                (a) MSVD
            \end{tabular}
            &
            \begin{tabular}{c}
                \begin{tabular}{@{}l@{~}c@{~}c@{~}c@{~}c@{}} \toprule
                     Method                              &B@4    & M    & R    & C     \\ \midrule
                    STG-KD          & 40.5 & 28.3 & 60.9 & 47.1 \\
                    Support-set & 38.9 & 28.2 & 59.8 & 48.6 \\
                    PMI-CAP      & 42.1 & 28.7 & -    & 49.4 \\
                    ORG-TRL       & 43.6 & 28.8 & 62.1 & 50.9 \\
                    OpenBook        & 33.9 & 23.7 & 50.2 & 52.9 \\
                    SwinBERT       & 41.9 & 29.9 & 62.1 & 53.8 \\ 
                    MV-GPT$^T$         & 48.9 & 38.7 & 64.0 & 60   \\
                    \midrule
                    \modelnamewospace$_\text{B}$         & 46.6 & 29.6 & 63.2 & 57.8 \\  
                    \modelnamewospace$_\text{L}$         & 48.7 & 30.9 & 64.9 & 64.1 \\  
                    \modelname                           & 53.8 & 32.9 & 67.7 & 73.9 \\  
                    \midrule
                    \modelnamewospace2                   & \bf54.8 & \bf33.1 & \bf68.2 & \bf75.9 \\  
                    \bottomrule
                \end{tabular}
                \\
               (b) MSRVTT
            \end{tabular}
            &
            \begin{tabular}{c}
                \begin{tabular}{@{}l@{~}c@{~}c@{~}c@{~}c@{}} \toprule
                    Method                                  & B@4       & M        & R        & C    \\ \midrule
                    VideoBERT       & 4.3      & 11.9     & -        & 55.0 \\ 
                    ActBERT           & 5.4      & 13.3     & -        & 65.0 \\ 
                    SwinBERT           & 9.0      & 15.6     & 37.3     & 109.0 \\
                    Flamingo     & -        &  -       & -        & 118.6 \\
                    VALUE$^{T}$                      & 12.4     & 18.8     & 40.4     & 130.3 \\ 
                    UniVL$^T$              & 17.4     & 22.4     & 46.5     & 181 \\
                    MV-GPT$^T$            & \bf21.9  & \bf27.1  & \bf49.4  & \bf221 \\
                    \midrule
                    \modelnamewospace$_\text{B}$            &  5.8    & 12.2    & 31.5    & 80.3         \\  
                    \modelnamewospace$_\text{L}$            &  7.5    & 14.4    & 34.9    & 98.3    \\  
                    \modelname                              & 10.3      & 17.3     & 39.8     & 129.8 \\ 
                    \midrule
                    \modelnamewospace2                      & 9.4     & 15.6      & 37.5      & 131.2 \\ 
                    \bottomrule
                \end{tabular}
                \\
                (c) YouCook2
            \end{tabular}
        \end{tabular}
        \begin{tabular}{c}
    \end{tabular}
    \\
    \\
    \begin{tabular}{@{}p{0.4\linewidth}@{}p{0.4\linewidth}@{}}
            \vtop{\vskip 0pt \vskip -\ht\strutbox
        \begin{tabular}{c}
            \begin{tabular}{@{}l@{~}c@{~}c@{~}c@{~}c@{}} \toprule
                     Method                              & B@4    & R     & M     & C \\ \midrule
                    VaTeX          & 28.4  & 47.0  & 21.7  & 45.1 \\
                 OpenBook          & 33.9  & 50.2  & 23.7  & 57.5 \\
                VALUE$^{T}$                      & -     & -     & -     & 58.1 \\
                SwinBERT          & 38.7  & 53.2  & 26.2  & 73.0 \\ 
                CLIP4Caption++$^{ET}$           & 40.6  & 54.5  & -     & 85.7  \\
                \midrule
                   \modelnamewospace$_\text{B}$         & 37.9    & 51.9    & 24.4    & 60.0   \\  
                   \modelnamewospace$_\text{L}$         &  41.6   & 54.3    & 26.2    & 72.5    \\  
                \modelname                              & 41.6  & 55.4  &  28.1  & 91.5  \\ 
                \midrule
                \modelnamewospace2                      & \bf42.7  & \bf56.5  &  \bf28.8  & \bf94.5  \\ 
                \bottomrule
            \end{tabular}
            \\
            (d) VATEX public test 
            \\
            \\
            \begin{tabular}{@{}l@{~}c@{}} \toprule
                     Method                              & C \\ \midrule
  X-L.+T.$^E$                        & 81.4 \\
                     Flamingo & 84.2 \\
            CLIP4Caption++$^{ET}$            &  86.5     \\
                \midrule
                \modelname                              & 93.8  \\ 
                \midrule
                 \modelnamewospace2                     & \bf96.6  \\ 
                \bottomrule
            \end{tabular}
            \\
           (e) VATEX private test  
        \end{tabular}
            \vskip -\dp\strutbox }
         &
         \vtop{\vskip 0pt \vskip -\ht\strutbox
         \begin{tabular}{c}
              \begin{tabular}{@{}l@{~}c@{~}c@{~}c@{~}c@{}} \toprule
                           Method                  & B@4    & R     & M     & C \\ \midrule
                          MMT$^{T}$   & 10.8  & 32.8  & 16.9  & 45.3 \\
                          HERO$^{T}$  & 12.3  & 34.1  & 17.6  & 49.9 \\
                          VALUE$^{T}$      & 11.6  & 33.9  & 17.6  & 50.5 \\
                  SwinBERT  & 14.5  & 36.1  & 18.5  & 55.4 \\ 
           CLIP4Caption++$^{ET}$      & 15.0  & 36.9  &  -    & 66.0 \\    
                  \midrule
                  \modelnamewospace$_\text{B}$         & 13.0    & 33.2    & 16.6    & 47.3         \\  
                  \modelnamewospace$_\text{L}$         & 14.9    & 35.4    & 18.0    &    55.7      \\  
                  \modelname                           & 16.2  & 36.7  & 18.9  & 63.0 \\ 
                  \midrule
                  \modelnamewospace2                   & \bf16.9  & \bf37.2  & \bf19.4  & \bf66.1 \\ 
                  \bottomrule
              \end{tabular}
              \\
              (f) TVC~\cite{lei2020tvr} validation
              \\
              \\
                      \begin{tabular}{lccc}
                           \toprule
                           Method                                 & C     \\
                           \midrule
                          CLIP4Caption++$^{T}$     & 59.41 \\
                          CLIP4Caption++$^{ET}$    & 64.49  \\
                          \midrule
                          \modelname                              & 61.19 \\
                          \modelnamewospace2                      & \bf65.02  \\
                          \bottomrule
                       \end{tabular}
            \\
            (g) TVC private test
          \end{tabular}
                      \vskip -\dp\strutbox }
         \end{tabular}
    \end{tabular}
\end{table}

\begin{table}[]
    \small
    \centering
    \caption{Results on video question answering. All are open-ended question answering tasks. 
    All-in-one: ~\cite{wang2022all},
    ClipBERT: \cite{LeiLZGBB021},
    CoMVT: ~\cite{SeoNS21},
    Flamingo: ~\cite{alayrac2022flamingo},
    JustAsk: ~\cite{YangMSLS21},
    MERLOT: \cite{ZellersLHYPCFC21},
    MV-GPT: ~\cite{abs-2201-08264},
    QueST: ~\cite{JiangCLZG20},
    HCRN: ~\cite{LeLVT21},
    VIOLET: ~\cite{abs-2111-12681}.
    }
    \label{tab:full_video_qa}
    \begin{tabular}{ccc}
                \begin{tabular}{lc}
                \toprule
                Method          & Accuracy \\
                            \midrule
            QueST        & 34.6 \\
            HCRN             & 36.1 \\
            CoMVT            & 42.6  \\
            JustAsk       & 46.3  \\
            VIOLET    &    47.9 \\
            All-in-one   & 48.3 \\
                \midrule
    \modelnamewospace$_\text{B}$            & 51.2    \\ 
    \modelnamewospace$_\text{L}$            & 55.1     \\ 
                \modelname                  & 56.8   \\ 
                \midrule
                \modelnamewospace2          & \bf58.2 \\ 
                \bottomrule
                \end{tabular}
         & 
                \begin{tabular}{lc}
                \toprule
                Method          & Accuracy \\
                            \midrule
            JustAsk       & 41.5  \\
            MV-GPT    & 41.7 \\
            MERLOT   & 43.1 \\
            VIOLET    & 43.9 \\
         All-in-one      & 46.8 \\
        Flamingo & \bf47.4 \\
                \midrule
    \modelnamewospace$_\text{B}$            & 41.0    \\ 
    \modelnamewospace$_\text{L}$            & 42.7    \\ 
                \modelname                  & 43.2 \\ 
                \midrule
        \modelnamewospace2                  & 45.6 \\ 
                \bottomrule
                \end{tabular}
         &
            \begin{tabular}{lc}
            \toprule
            Method          &  Accuracy \\
            \midrule
        HCRN             & 55.9 \\
        QueST        & 59.7 \\
        ClipBERT      &   60.3 \\
           All-in-one     & 66.3 \\
        VIOLET    &    68.9 \\
        MERLOT   & 69.5 \\
            \midrule
    \modelnamewospace$_\text{B}$        & 69.1    \\ 
    \modelnamewospace$_\text{L}$        & 71.9    \\ 
            \modelname                  & 72.8 \\ 
            \midrule
            \modelnamewospace2          & \bf74.9 \\ 
            \bottomrule
            \end{tabular}
            \\
            (a) MSVD-QA & (b) MSRVTT-QA & (c) TGIF-Frame
    \end{tabular}
\end{table}

\begin{table}[t]
    \centering
    \small
    \caption{Fine-tuning epochs and learning rate of \modelname on video tasks. SCST~\citep{scst} is performed for VATEX with the same hyperprameters.}
    \label{tab:video_hyper}
    \begin{tabular}{c@{~~}c@{~~}c@{~~}c@{~~}c@{~~}c@{~~}c@{~~}c@{~~}c}
    \toprule
                    & \multicolumn{5}{c}{Video Captioning}  & \multicolumn{3}{c}{Video Question Answering} \\
                    \cmidrule(lr){2-6} \cmidrule(lr){7-9}
                    &  MSVD     & MSRVTT        & YouCook2  & VATEX       & TVC       & MSVD       & MSRVTT    & TGIF-Frame \\
                    \midrule
    epochs          &  10       & 20            & 20        & 10          & 20        & 40         & 20        & 20  \\
    learning rate   & $1e^{-6}$ & $2.5e^{-6}$   & $1e^{-5}$ & $2.5e^{-6}$ & $5e^{-6}$ & $1e^{-5}$  & $1e^{-5}$ & $1e^{-5}$ \\
    \bottomrule
    \end{tabular}
\end{table}

\section{Results on Video Captioning and Question Answering}
Table~\ref{tab:video_hyper} shows the fine-tuning hyperparameters on video tasks for \modelnamewospace. 
Table~\ref{tab:full_video_caption} and Table~\ref{tab:full_video_qa}
show the complete results on video captioning and video question answering, respectively.
{
During training, we randomly sample 6 frames with equal interval,
and apply the same random crop on these frames. 
During inference, we uniformly sample 6 frames  with center crop.
}


\begin{figure}
    \centering
\small
\begin{python}
from nltk.corpus import wordnet as wn
def get_name(offset):
    white_list = {
      2012849: 'crane bird',
      3126707: 'crane machine',
      2113186: 'cardigan dog',
      2963159: 'cardigan jacket', 
      3710637: 'maillot tights',
      3710721: 'maillot bathing suit',
    }
    if offset in white_list:
        return white_list[offset]
    name = wn.synset_from_pos_and_offset('n', offset).name()
    return name[:-5].replace('_', ' ')      
\end{python}
    \caption{Python script to generate a unique name for each offset in ImageNet-1K categories.}
    \label{fig:python_script_imagenet_name}
\end{figure}

\begin{figure}
\centering
\small
\renewcommand{\arraystretch}{0.8}
\begin{tabular}{p{\fivewidth\linewidth}@{~~}p{\fivewidth\linewidth}@{~~}p{\fivewidth\linewidth}@{~~}p{\fivewidth\linewidth}@{~~}p{\fivewidth\linewidth}}
\includegraphics[width=1\linewidth,height=1\linewidth]{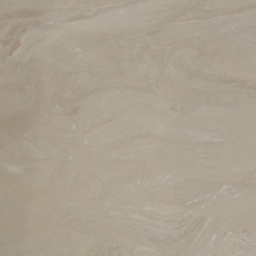} & \includegraphics[width=1\linewidth,height=1\linewidth]{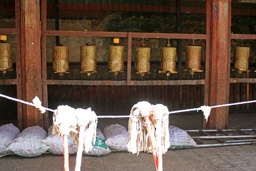} & \includegraphics[width=1\linewidth,height=1\linewidth]{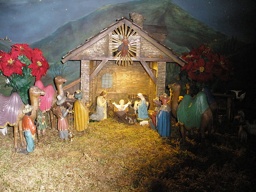} & \includegraphics[width=1\linewidth,height=1\linewidth]{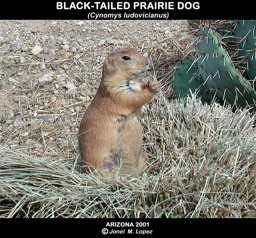} & \includegraphics[width=1\linewidth,height=1\linewidth]{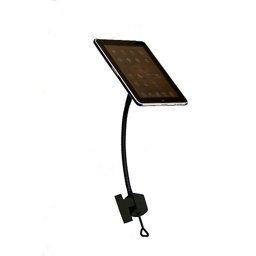}\\
\hfil (a) & \hfil (b) & \hfil (c) & \hfil (d) & \hfil (e)\\
{\textbf{Pred}}: tile & {\textbf{Pred}}: prayer wheel & {\textbf{Pred}}: nativity & {\textbf{Pred}}: prairie dog & {\textbf{Pred}}: ipad\\
\textbf{GT}: sandbar & \textbf{GT}: swab & \textbf{GT}: cradle & \textbf{GT}: wood rabbit & \textbf{GT}: hand-held computer\\
\includegraphics[width=1\linewidth,height=1\linewidth]{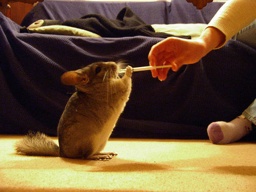} & \includegraphics[width=1\linewidth,height=1\linewidth]{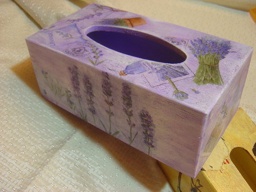} & \includegraphics[width=1\linewidth,height=1\linewidth]{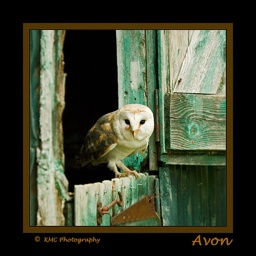} & \includegraphics[width=1\linewidth,height=1\linewidth]{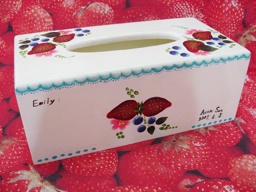} & \includegraphics[width=1\linewidth,height=1\linewidth]{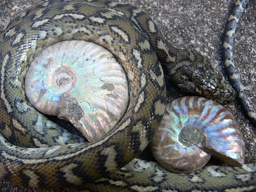}\\
\hfil (f) & \hfil (g) & \hfil (h) & \hfil (i) & \hfil (j)\\
{\textbf{Pred}}: chinchilla & {\textbf{Pred}}: tissue box & {\textbf{Pred}}: barn owl & {\textbf{Pred}}: tissue box & {\textbf{Pred}}: chambered snake\\
\textbf{GT}: syringe & \textbf{GT}: carton & \textbf{GT}: barn & \textbf{GT}: carton & \textbf{GT}: chambered nautilus\\
\end{tabular}
\caption{Image samples on which the prediction of our GIT is out of the 1K categories with the whitespace removed on ImageNet-1K.}
\label{fig:space_exclude_outlier}
\end{figure}

\begin{table}[]
    \centering
    \caption{Results on ImageNet-1k classification task.
    Our approach takes the class name as the caption and predict 
    the label in an auto-regressive way without pre-defining the vocabulary.}
    \label{tab:imagenet}
    \begin{tabular}{llc}
    \toprule
    Vocabulary          &  Method                        & Top-1 \\
    \midrule
\multirow{2}{*}{Closed} & ALIGN~\citep{JiaYXCPPLSLD21}    & 88.64 \\
                        & Florence~\citep{abs-2111-11432} & 90.05 \\
                        & CoCa~\citep{yu2022coca}         & \bf91.0  \\
    \midrule
    \multirow{4}{*}{Open}& \modelnamewospace$_\text{B}$  & 78.86 \\ 
                         & \modelnamewospace$_\text{L}$  & 84.05 \\ 
                         & \modelname                    & 88.79 \\ 
                         \cmidrule(lr){2-3}
                         & \modelnamewospace2              & 89.22 \\ 
    \bottomrule
    \end{tabular}
\end{table}

\begin{table}[h]
    \centering
    \caption{
    Zero/Few-shot evaluation on ImageNet with 3 metrics. 
    \textit{equal}: the unrestricted prediction should be exactly matched to the ground-truth.
    \textit{in}: the unrestricted prediction should contain the ground-truth label name.
    \textit{voc-prior}: the vocabulary is pre-defined as a prior. For our \modelname, a trie structure is constructed motivated from \cite{abs-2202-03052} 
    to limit the candidate tokens during each token prediction, such that the predicted result is guaranteed to be within the vocabulary. 
    }
    \label{tab:imagenet_few_shot_full}
    \begin{tabular}{cccccccccc}
    \toprule
                 & \multicolumn{3}{c}{Zero-shot}    & \multicolumn{3}{c}{1-shot per class}  & \multicolumn{3}{c}{5-shot per class} \\
                 \cmidrule(lr){2-4}   \cmidrule(lr){5-7} \cmidrule(lr){8-10}
Accuracy type    & equal & in     & voc-prior       & equal & in    & voc-prior   & equal & in    & voc-prior   \\
\midrule          
Flamingo         &   -   & -      &  -              & -     & -     & 71.7        & -     & -     & 77.3        \\
\midrule
\modelnamewospace$_\text{B}$ & 0 & 11.49 & 9.34            & 29.8  & 30.97 & 35.41       & 51.99 & 52.39 & 53.4  \\
\modelnamewospace$_\text{L}$ & 0 & 15.28 & 10.35           & 43.24 & 45.75 & 53.06       & 68.35 & 68.85 & 69.91 \\
\modelname       & 1.93  & 40.88  & 33.48           & 64.54 & 66.76 & 72.45       & 79.79 & 80.15 & 80.95  \\
\modelnamewospace2      & 1.91  & 41.92  & 40.57           & 67.01 & 69.07 & 74.93       & 80.06 & 80.57 & 82.29  \\
\bottomrule
    \end{tabular}
\end{table}

\section{Results on Image Classification}
On ImageNet-1K~\citep{imagenet_cvpr09}, 
we map each label to a 
unique name.
Each label belongs to an entry of WordNet hierarchy and is represented with a unique offset, \emph{e.g.}, 2012849.
Fig.~\ref{fig:python_script_imagenet_name} illustrates the python script to generate a readable
unique name given the offset. 
The model is fine-tuned with 10 epochs and the learning rate is $1e^{-5}$.
The batch size is 4096 for the full fine-tuning and 16 for the few-shot setting.
No beam search is performed during inference.

Table~\ref{tab:imagenet} and Table~\ref{tab:imagenet_few_shot_full} shows
the full results with other model variants.

\begin{figure}
\centering
\small
\renewcommand{\arraystretch}{0.8}
\begin{tabular}{p{\fivewidth\linewidth}@{~~}p{\fivewidth\linewidth}@{~~}p{\fivewidth\linewidth}@{~~}p{\fivewidth\linewidth}@{~~}p{\fivewidth\linewidth}}
\includegraphics[width=1\linewidth,height=1\linewidth]{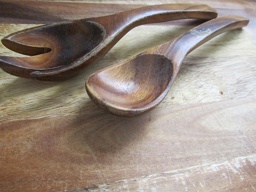} & \includegraphics[width=1\linewidth,height=1\linewidth]{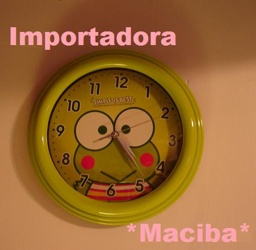} & \includegraphics[width=1\linewidth,height=1\linewidth]{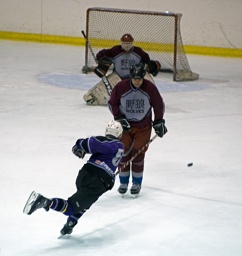} & \includegraphics[width=1\linewidth,height=1\linewidth]{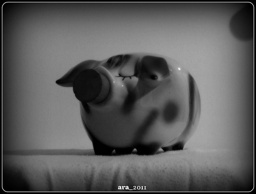} & \includegraphics[width=1\linewidth,height=1\linewidth]{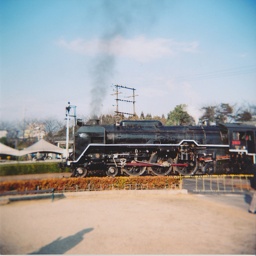}\\
{\textbf{Pred}}: wooden spoon & {\textbf{Pred}}: wall clock & {\textbf{Pred}}: puck & {\textbf{Pred}}: piggy bank & {\textbf{Pred}}: steam locomotive\\
\includegraphics[width=1\linewidth,height=1\linewidth]{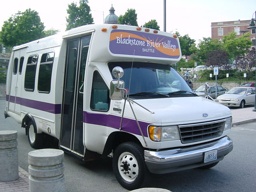} & \includegraphics[width=1\linewidth,height=1\linewidth]{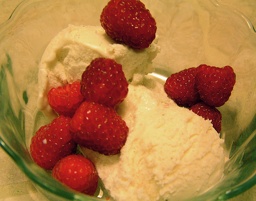} & \includegraphics[width=1\linewidth,height=1\linewidth]{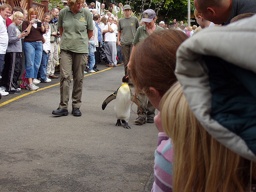} & \includegraphics[width=1\linewidth,height=1\linewidth]{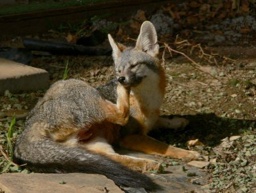} & \includegraphics[width=1\linewidth,height=1\linewidth]{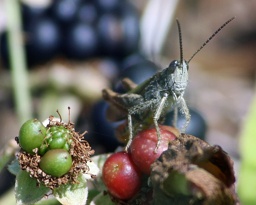}\\
{\textbf{Pred}}: minibus & {\textbf{Pred}}: ice cream & {\textbf{Pred}}: king penguin & {\textbf{Pred}}: grey fox & {\textbf{Pred}}: grasshopper\\
\includegraphics[width=1\linewidth,height=1\linewidth]{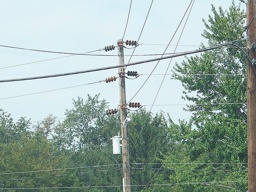} & \includegraphics[width=1\linewidth,height=1\linewidth]{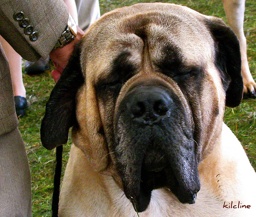} & \includegraphics[width=1\linewidth,height=1\linewidth]{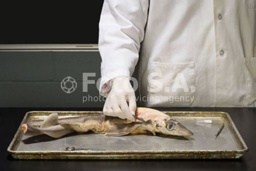} & \includegraphics[width=1\linewidth,height=1\linewidth]{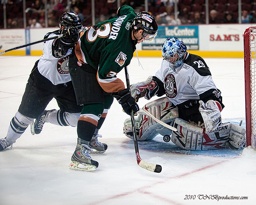} & \includegraphics[width=1\linewidth,height=1\linewidth]{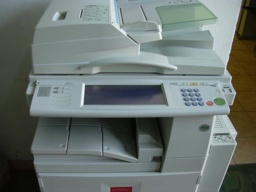}\\
{\textbf{Pred}}: pole & {\textbf{Pred}}: bull mastiff & {\textbf{Pred}}: lab coat & {\textbf{Pred}}: puck & {\textbf{Pred}}: photocopier\\
\includegraphics[width=1\linewidth,height=1\linewidth]{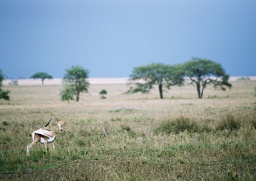} & \includegraphics[width=1\linewidth,height=1\linewidth]{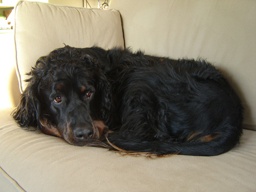} & \includegraphics[width=1\linewidth,height=1\linewidth]{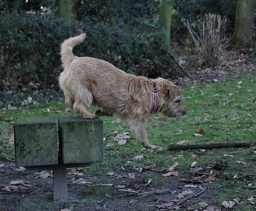} & \includegraphics[width=1\linewidth,height=1\linewidth]{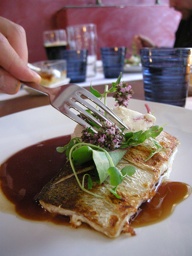} & \includegraphics[width=1\linewidth,height=1\linewidth]{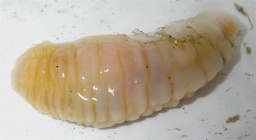}\\
{\textbf{Pred}}: gazelle & {\textbf{Pred}}: gordon setter & {\textbf{Pred}}: norfolk terrier & {\textbf{Pred}}: plate & {\textbf{Pred}}: sea cucumber\\
\includegraphics[width=1\linewidth,height=1\linewidth]{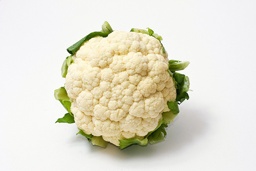} & \includegraphics[width=1\linewidth,height=1\linewidth]{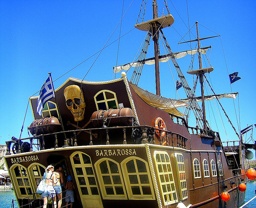} & \includegraphics[width=1\linewidth,height=1\linewidth]{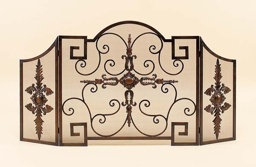} & \includegraphics[width=1\linewidth,height=1\linewidth]{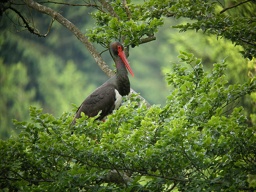} & \includegraphics[width=1\linewidth,height=1\linewidth]{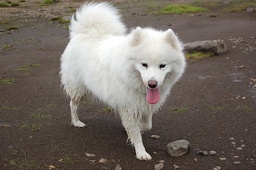}\\
{\textbf{Pred}}: cauliflower & {\textbf{Pred}}: pirate & {\textbf{Pred}}: fire screen & {\textbf{Pred}}: black stork & {\textbf{Pred}}: samoyed\\
\includegraphics[width=1\linewidth,height=1\linewidth]{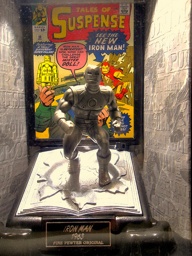} & \includegraphics[width=1\linewidth,height=1\linewidth]{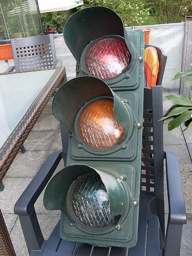} & \includegraphics[width=1\linewidth,height=1\linewidth]{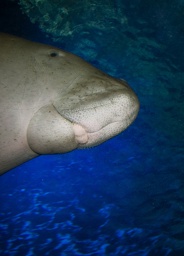} & \includegraphics[width=1\linewidth,height=1\linewidth]{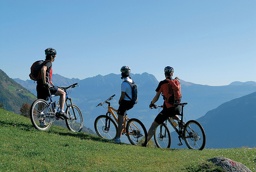} & \includegraphics[width=1\linewidth,height=1\linewidth]{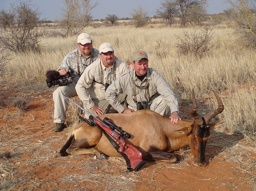}\\
{\textbf{Pred}}: comic book & {\textbf{Pred}}: traffic light & {\textbf{Pred}}: dugong & {\textbf{Pred}}: mountain bike & {\textbf{Pred}}: hartebeest\\
\end{tabular}
\caption{Visualization of correct predictions of our generative model on ImageNet-1K.}
\label{fig:vis_TaxImageNet2012GenV1_good}
\end{figure}

\begin{figure}
\centering
\small
\renewcommand{\arraystretch}{0.8}
\begin{tabular}{p{\fivewidth\linewidth}@{~~}p{\fivewidth\linewidth}@{~~}p{\fivewidth\linewidth}@{~~}p{\fivewidth\linewidth}@{~~}p{\fivewidth\linewidth}}
\includegraphics[width=1\linewidth,height=1\linewidth]{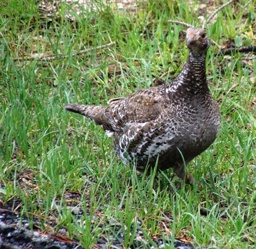} & \includegraphics[width=1\linewidth,height=1\linewidth]{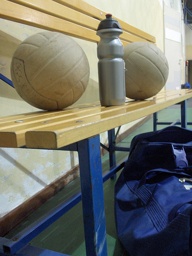} & \includegraphics[width=1\linewidth,height=1\linewidth]{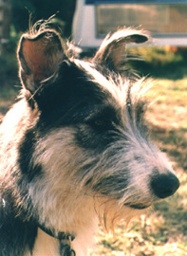} & \includegraphics[width=1\linewidth,height=1\linewidth]{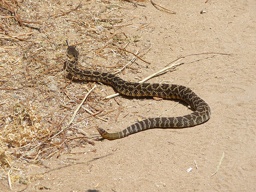} & \includegraphics[width=1\linewidth,height=1\linewidth]{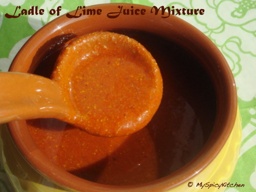}\\
{\textbf{Pred}}: black grouse & {\textbf{Pred}}: volleyball & {\textbf{Pred}}: scottish deerhound & {\textbf{Pred}}: diamondback & {\textbf{Pred}}: wooden spoon\\
\textbf{GT}: ruffed grouse & \textbf{GT}: water bottle & \textbf{GT}: australian terrier & \textbf{GT}: night snake & \textbf{GT}: ladle\\
\includegraphics[width=1\linewidth,height=1\linewidth]{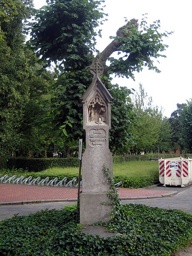} & \includegraphics[width=1\linewidth,height=1\linewidth]{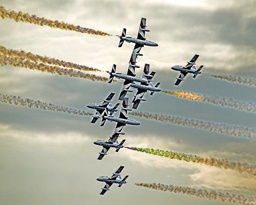} & \includegraphics[width=1\linewidth,height=1\linewidth]{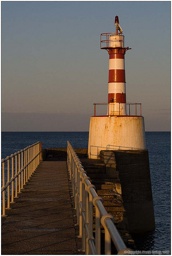} & \includegraphics[width=1\linewidth,height=1\linewidth]{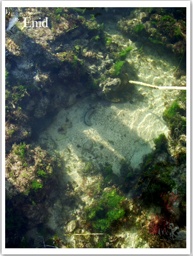} & \includegraphics[width=1\linewidth,height=1\linewidth]{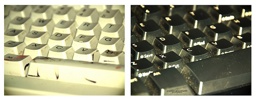}\\
{\textbf{Pred}}: obelisk & {\textbf{Pred}}: warplane & {\textbf{Pred}}: beacon & {\textbf{Pred}}: coral reef & {\textbf{Pred}}: computer keyboard\\
\textbf{GT}: pedestal & \textbf{GT}: wing & \textbf{GT}: breakwater & \textbf{GT}: sea cucumber & \textbf{GT}: space bar\\
\includegraphics[width=1\linewidth,height=1\linewidth]{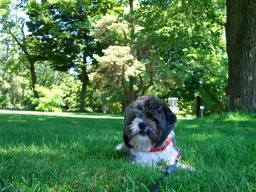} & \includegraphics[width=1\linewidth,height=1\linewidth]{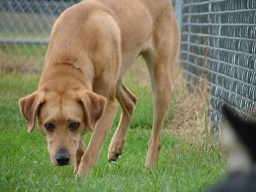} & \includegraphics[width=1\linewidth,height=1\linewidth]{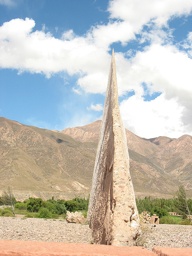} & \includegraphics[width=1\linewidth,height=1\linewidth]{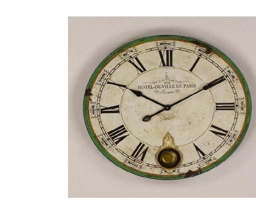} & \includegraphics[width=1\linewidth,height=1\linewidth]{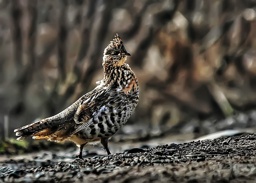}\\
{\textbf{Pred}}: lhasa & {\textbf{Pred}}: rhodesian ridgeback & {\textbf{Pred}}: megalith & {\textbf{Pred}}: wall clock & {\textbf{Pred}}: ruffed grouse\\
\textbf{GT}: shih-tzu & \textbf{GT}: vizsla & \textbf{GT}: sundial & \textbf{GT}: analog clock & \textbf{GT}: partridge\\
\includegraphics[width=1\linewidth,height=1\linewidth]{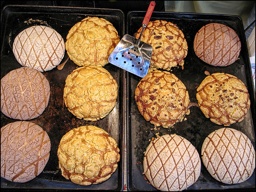} & \includegraphics[width=1\linewidth,height=1\linewidth]{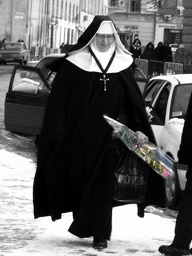} & \includegraphics[width=1\linewidth,height=1\linewidth]{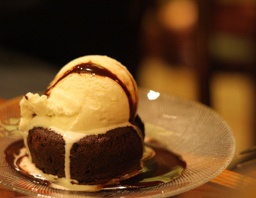} & \includegraphics[width=1\linewidth,height=1\linewidth]{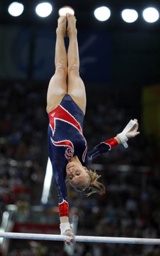} & \includegraphics[width=1\linewidth,height=1\linewidth]{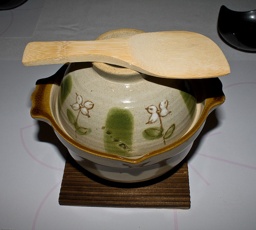}\\
{\textbf{Pred}}: bakery & {\textbf{Pred}}: vestment & {\textbf{Pred}}: ice cream & {\textbf{Pred}}: horizontal bar & {\textbf{Pred}}: mixing bowl\\
\textbf{GT}: spatula & \textbf{GT}: cloak & \textbf{GT}: chocolate sauce & \textbf{GT}: parallel bars & \textbf{GT}: wooden spoon\\
\includegraphics[width=1\linewidth,height=1\linewidth]{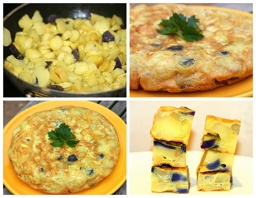} & \includegraphics[width=1\linewidth,height=1\linewidth]{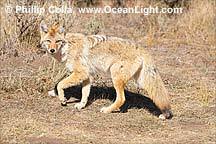} & \includegraphics[width=1\linewidth,height=1\linewidth]{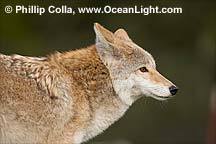} & \includegraphics[width=1\linewidth,height=1\linewidth]{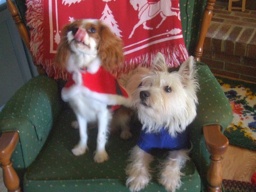} & \includegraphics[width=1\linewidth,height=1\linewidth]{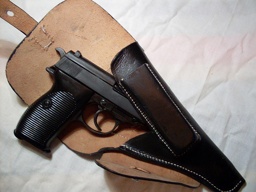}\\
{\textbf{Pred}}: frying pan & {\textbf{Pred}}: grey fox & {\textbf{Pred}}: coyote & {\textbf{Pred}}: blenheim spaniel & {\textbf{Pred}}: holster\\
\textbf{GT}: plate & \textbf{GT}: coyote & \textbf{GT}: red wolf & \textbf{GT}: west highland white terrier & \textbf{GT}: revolver\\
\end{tabular}
\caption{Visualization of incorrect predictions of our generative model on ImageNet-1K.}
\label{fig:vis_TaxImageNet2012GenV1_bad}
\end{figure}

In the main paper, we demonstrated a decent 
accuracy of 88.79\% top-1 on ImageNet-1k with
our generative model in the full fine-tuning setting.
As no constraint is on the 
output, 
we find that only 13 or (or $0.026\%$)
predictions are outside of the 1K category.
Fig.~\ref{fig:space_exclude_outlier} illustrates 10 samples. Although deemed as incorrect,
some predictions are reasonable.
For example, the prediction of 
Fig.~\ref{fig:space_exclude_outlier} (e) is \textit{ipad} and is reasonable,
although the ground-truth label is \textit{hand-held computer}.
These observations also imply that the generation model can quickly adapt to the 
classification task without pre-defining the vocabulary.
Fig.~\ref{fig:vis_TaxImageNet2012GenV1_good}
and Fig.~\ref{fig:vis_TaxImageNet2012GenV1_bad}
show the correct and incorrect predictions, respectively.


\newpage
\section{Results on Scene Text Recognition}
\begin{figure*}[]
\centering
\includegraphics[width=0.83\textwidth]{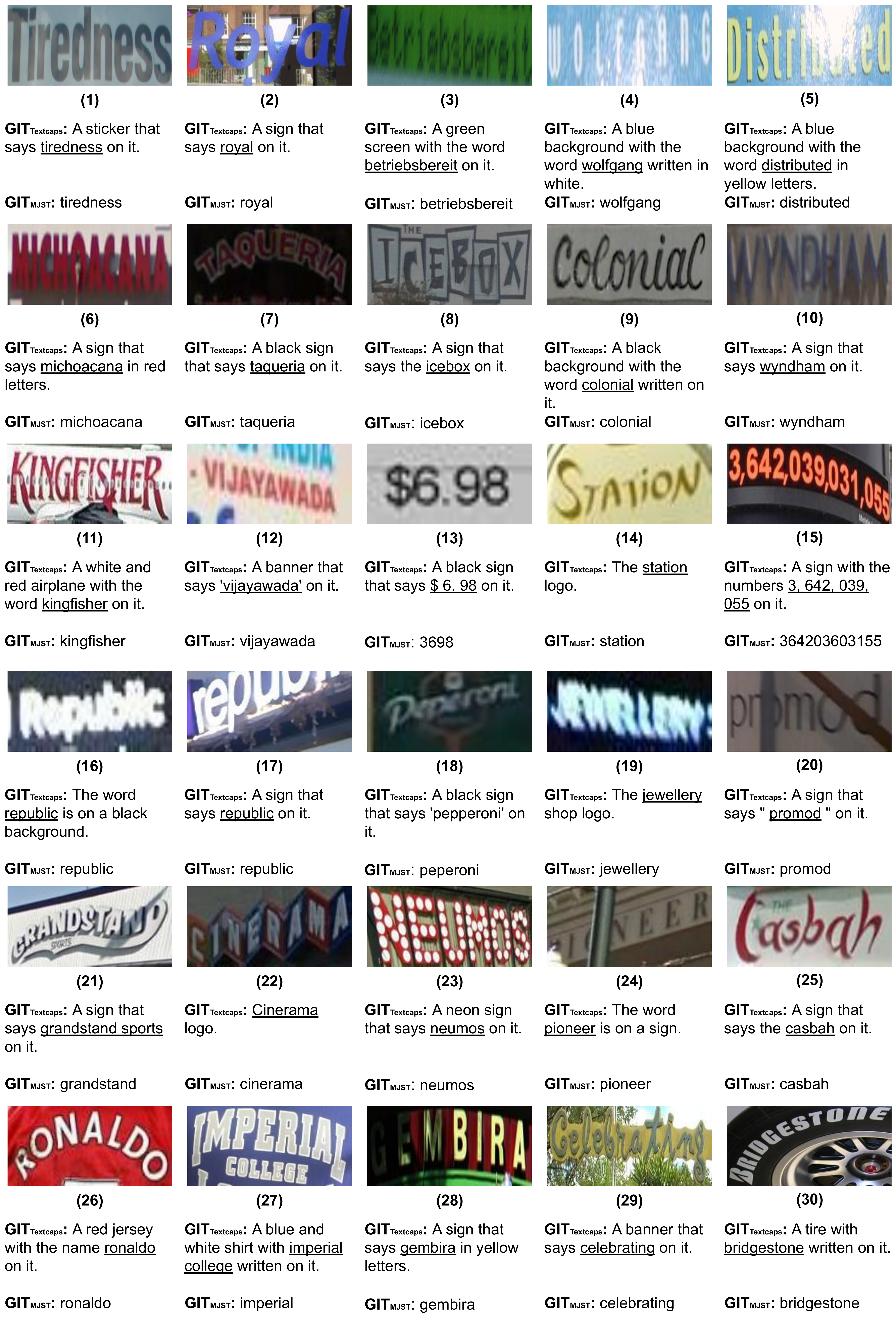}
\caption{Grouped scene text recognition predictions on all six experimented benchmarks. \modelnamewospace$_\text{Textcaps}$ and \modelnamewospace$_\text{MJST}$ are model variants finetuned with TextCaps caption data~\citep{abs-2003-12462}, and scene text recognition data MJ+ST~\citep{jaderberg2014synthetic,jaderberg2016reading,gupta2016synthetic}, respectively. (1-5) IC13~\citep{karatzas2013icdar}. (6-10) SVT~\citep{wang2011end}. (11-15) IIIT~\citep{mishra2012scene}. (16-20) IC15~\citep{karatzas2015icdar}. (21-25) SVTP~\citep{phan2013recognizing}. (26-30) CUTE~\citep{risnumawan2014robust}.}
\label{fig:visuD}
\end{figure*}

Table~\ref{table:str} shows the performance on six individual evaluation sets.
Fig.~\ref{fig:visuD} shows the visualization samples with our TextCaps-fine-tuned \modelname (denoted as
\modelnamewospace$_\text{TextCaps}$)
and 
with the MJ+ST-fine-tuned \modelname (denoted as \modelnamewospace$_\text{MJSJ}$). For scene text recognition, we resize the image with the longer edge to 384 and pad the image to a square.
Visually, \modelnamewospace$_\text{TextCaps}$ can well recognize the scene text, almost as good as 
\modelnamewospace$_\text{MJSJ}$, but in the natural language form. 
\modelnamewospace$_\text{MJSJ}$ can adapt well to the task and predict a clean result of the scene text. 
Fig.~\ref{fig:visuD} shows the visualizations on all six experimented benchmarks, \emph{i.e.}, IC13~\citep{karatzas2013icdar}, SVT~\citep{wang2011end}, IIIT~\citep{mishra2012scene}, IC15~\citep{karatzas2015icdar}, SVTP~\citep{phan2013recognizing}, CUTE~\citep{risnumawan2014robust} from the top to the bottom row, respectively. \modelname performs especially well on testing images visually similar to natural images, such as the CUTE dataset shown in the bottom row. Quantitatively, \modelname achieves an even larger performance improvement of $3.9\%$ absolute accuracy on Irregular-Text CUTE80.

{
We also finetune the pretrained \modelname on the TextOCR~\citep{singh2021textocr}
benchmarks.
As the test annotations are not publicly available, we evaluate the performance
on the validation set and achieve 81.27\% accuracy. 
}

\begin{table}[]
    \small
    \centering
    \caption{Results on scene text recognition. MJ and ST indicate the MJSynth (MJ)~\citep{jaderberg2014synthetic,jaderberg2016reading} and SynthText (ST)~\citep{gupta2016synthetic} datasets used for training scene text recognition models.
    SAM: ~\cite{liao2019mask},
    Ro.Scanner: ~\cite{yue2020robustscanner}
    SRN: ~\cite{yu2020towards}
    ABINet: ~\cite{fang2021read}
    S-GTR: ~\cite{he2021visual}
    MaskOCR: ~\cite{lyu2022maskocr}
    }
\begin{tabular}{l @{~~~} c@{~~~}c@{~~~}c@{~~~}c@{~~~}c@{~~~}c@{~~~}c@{~~~}c}
    \toprule
   \multirow{2}{*}{Method} & Fine-tuning & \multicolumn{3}{c}{Regular Text} & \multicolumn{3}{c}{Irregular Text} & \multirow{2}{*}{Average} \\
    \cmidrule(lr){3-5} \cmidrule(lr){6-8} & Data     & IC13 & SVT & IIIT & IC15 & SVTP & CUTE & \\
   \midrule
    SAM~\citep{liao2019mask}                     & MJ+ST     & 95.3    & 90.6 & 93.9 & 77.3 & 82.2 & 87.8 & 87.8 \\
    Ro.Scanner~\citep{yue2020robustscanner}   & MJ+ST     & 94.8    & 88.1 & 95.3 & 77.1 & 79.5 & 90.3 & 87.5 \\
    SRN~\citep{yu2020towards}                    & MJ+ST     & 95.5    & 91.5 & 94.8 & 82.7 & 85.1 & 87.8 & 89.6 \\
    ABINet~\citep{fang2021read}                  & MJ+ST     & 97.4 & 93.5 & 96.2 & 86.0 & 89.3 & 89.2 & 91.9 \\
    S-GTR~\citep{he2021visual}                   & MJ+ST     & 96.8    & 94.1 & 95.8 & 84.6 & 87.9 & 92.3 & 91.9 \\
    MaskOCR~\citep{lyu2022maskocr}               & MJ+ST     & \bf97.8 & 94.1 & 96.5 & \bf88.7 & 90.2 & 92.7 & 93.8 \\
    \midrule
    \multirow{2}{*}{\modelname}                 & TextCaps  & 94.2    & 91.5 & 92.9 & 78.2 & 87.1 & 95.5 & 89.9 \\
                                                & MJ+ST     & 97.3    & 95.2 & 95.3 & 83.7 & 89.9 & 96.2 & 92.9 \\
                                                \midrule
    \modelnamewospace2                          & MJ+ST     & \bf97.8 & \bf95.5 & \bf97.6 & 85.6 & \bf91.3 & \bf99.0 & \bf94.5 \\ 
    \bottomrule
    \end{tabular}
    \label{table:str}
\end{table}






\begin{figure}[t!]
    \centering
    \small
    \begin{tabular}{@{}c@{}c@{}c@{}}
    \includegraphics[width=0.33\linewidth]{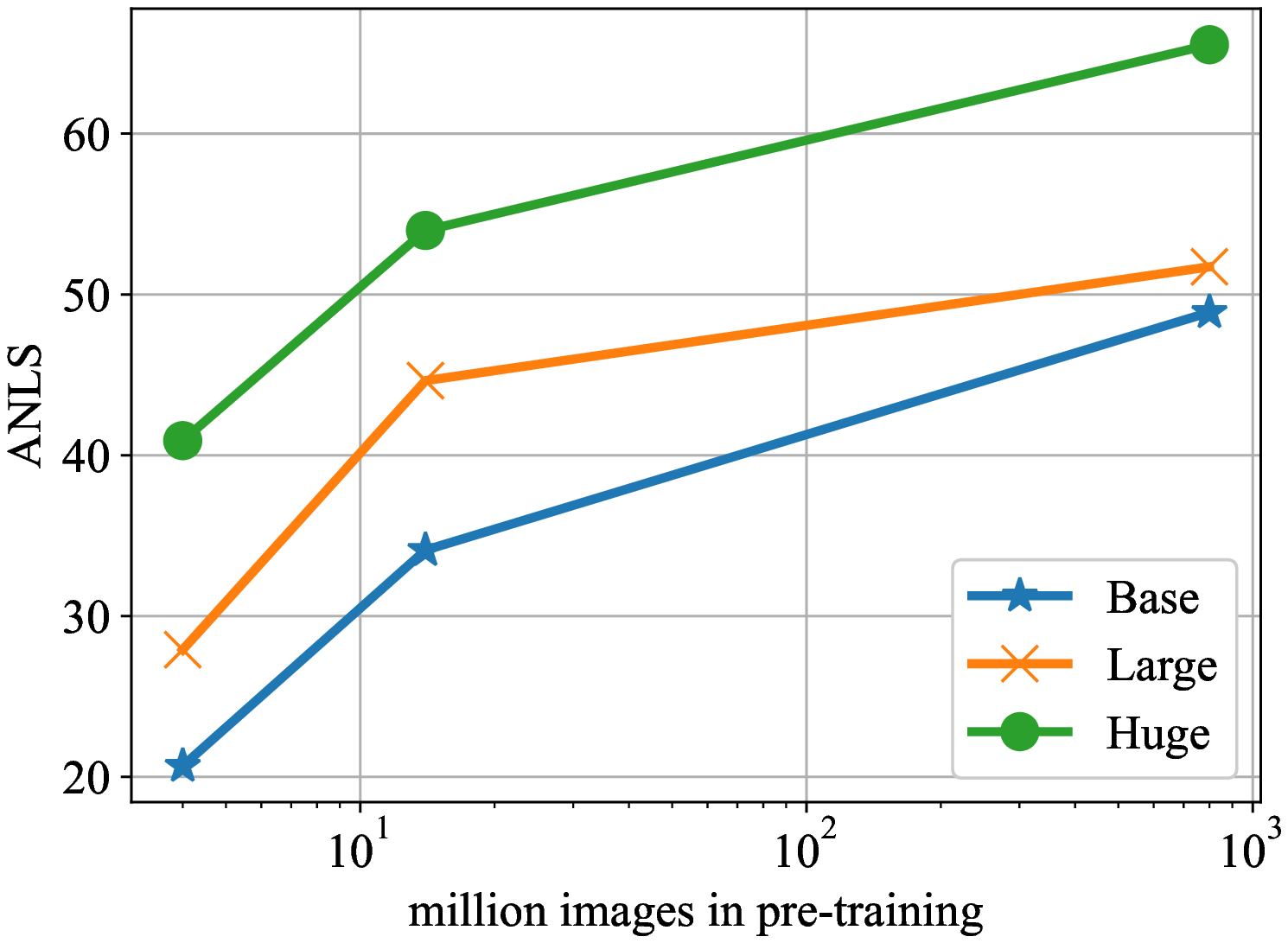}         &      
    \includegraphics[width=0.33\linewidth]{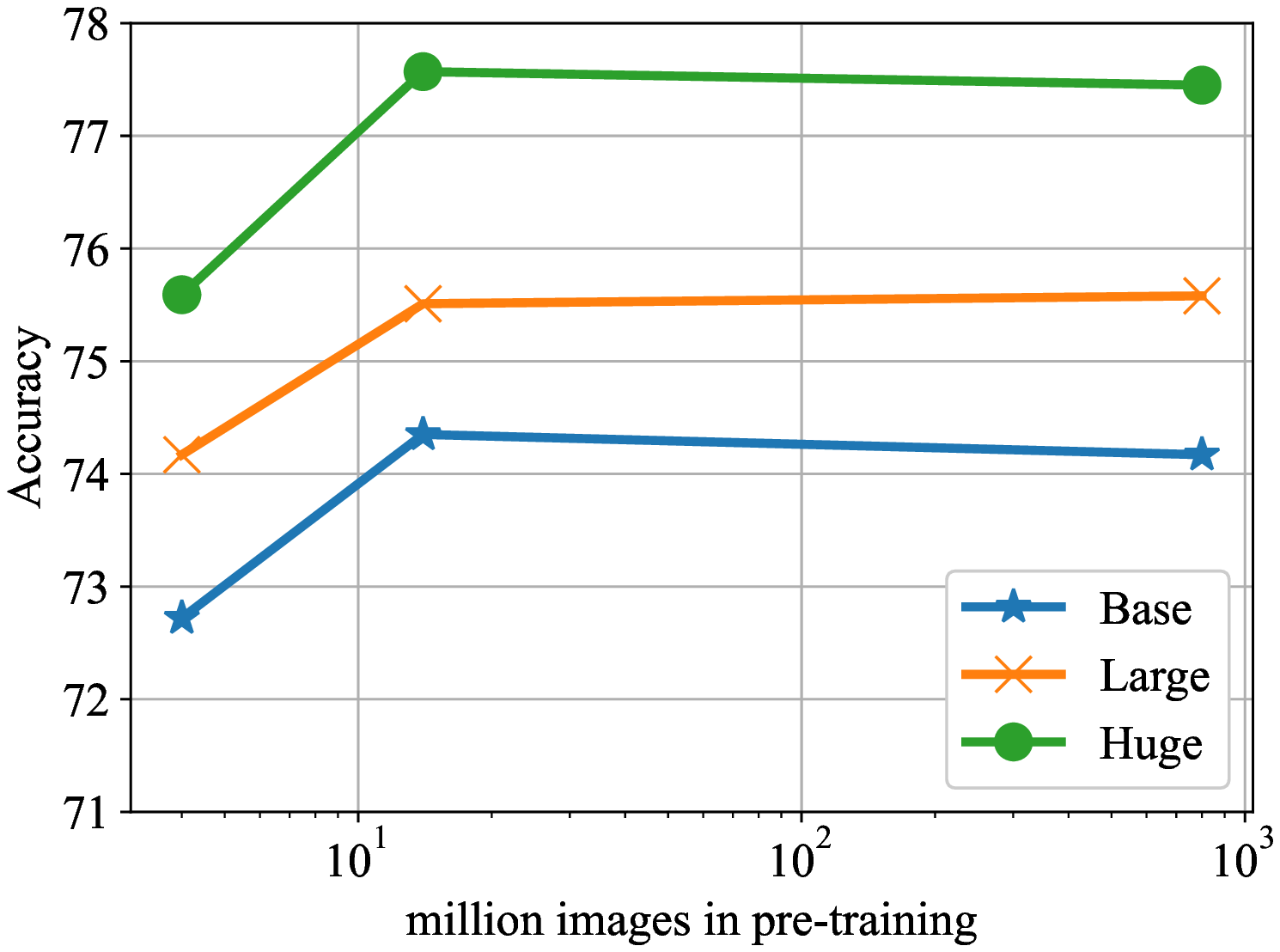}     & 
    \includegraphics[width=0.33\linewidth]{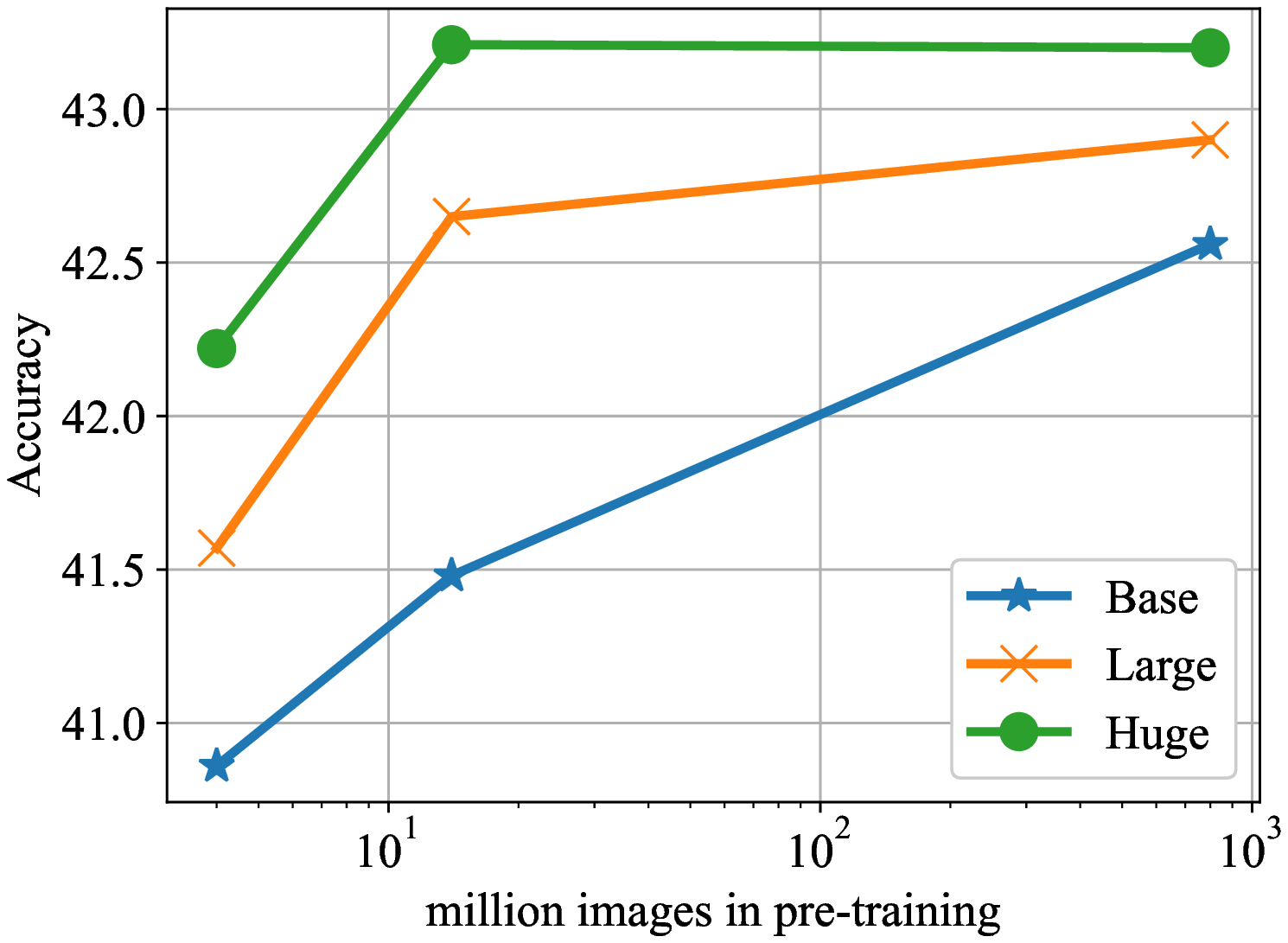}         \\
            (a) ST-VQA    &   (b) VQAv2  & (c) MSRVTT-QA  \\
    \includegraphics[width=0.33\linewidth]{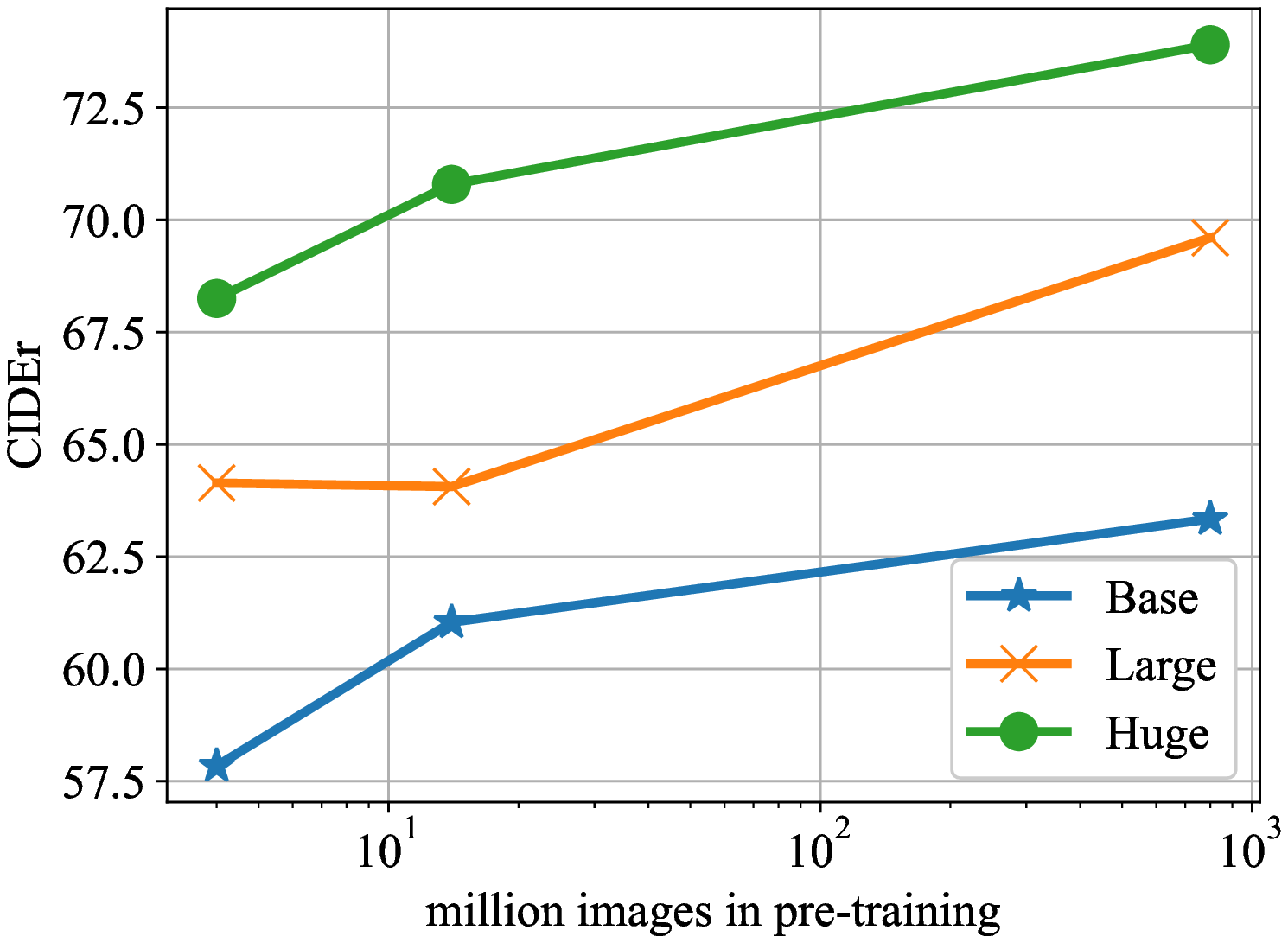}         &      
    \includegraphics[width=0.33\linewidth]{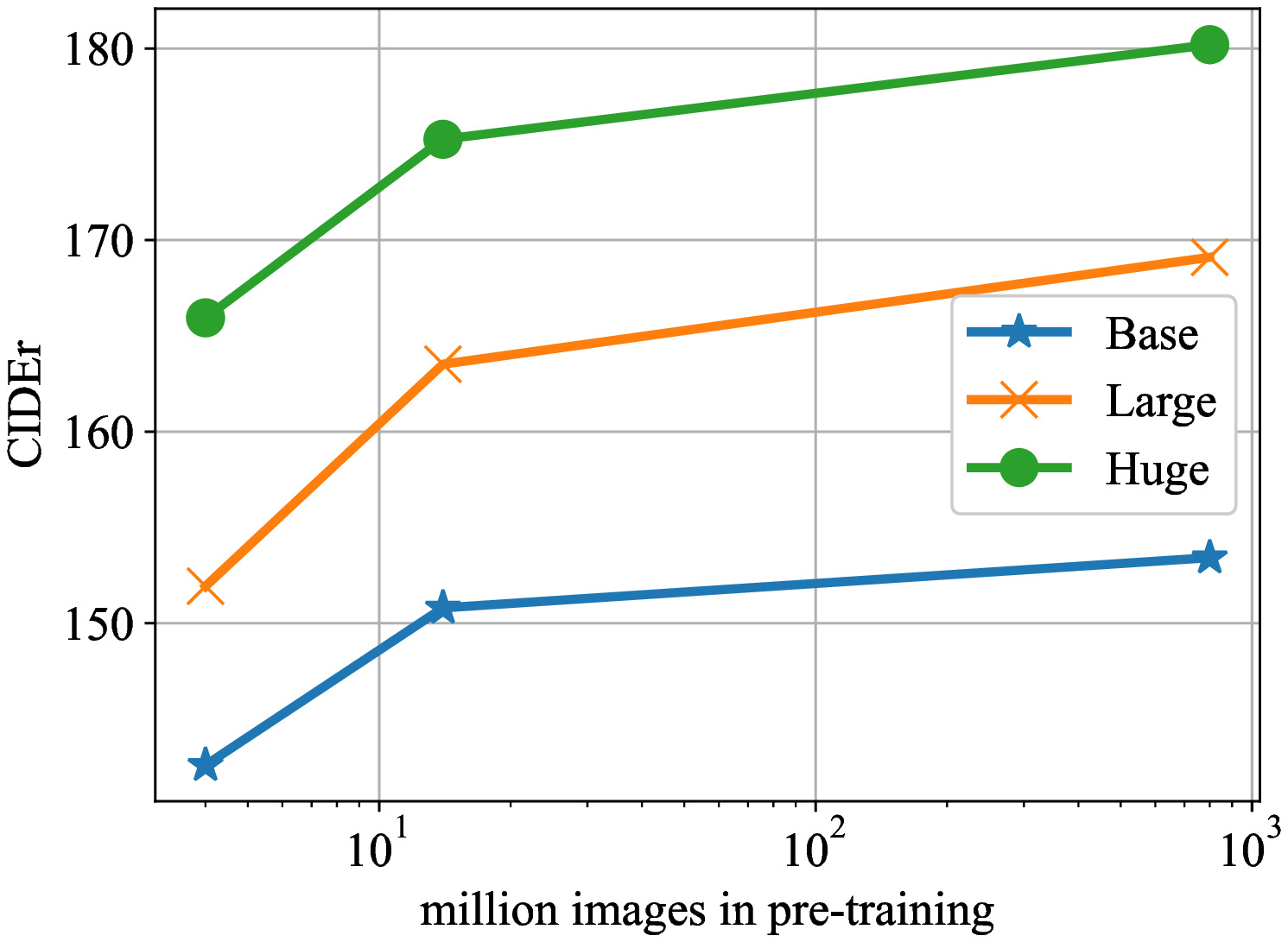}     & 
    \includegraphics[width=0.33\linewidth]{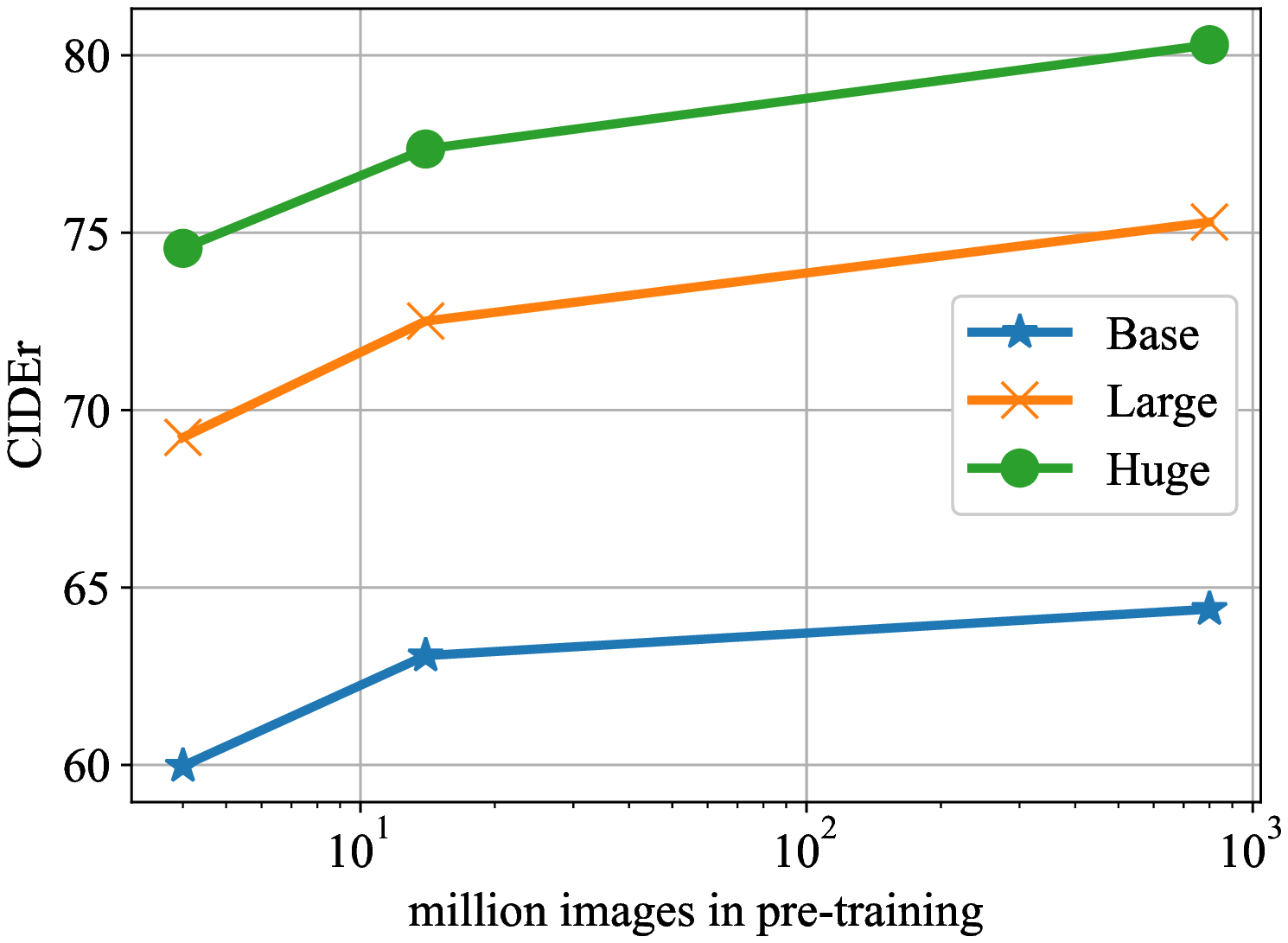}     \\
    (d) MSRVTT   & (e) MSVD & (f) VATEX
    \end{tabular}
    \caption{
    Performance with different pre-training data scales 
    and different model sizes. 
    }
    \label{fig:scale_more}
\end{figure}

\section{Analysis}
\subsection{Model and data scaling}
In the main paper, we present the impact of scaling on COCO, TextCaps and VizWiz-QA.
Fig.~\ref{fig:scale_more} shows results on other tasks.
On scene-text-related QA tasks (a) and video captioning (d)/(e)/(f),
both larger model sizes and more pre-training data boost the performance significantly. 
For VQAv2 (b), the 0.8B data help little or even worsen the performance slightly. 
The task data~\citep{GoyalKSBP16} are from COCO, and 
the first 20M image-text pairs are more similar to COCO images than 
the majority of the web crawled 0.8B data.
This may indicate the first 20M image-text pairs are enough for VQAv2.
For video QA (c), the improvement on more pre-training data is mild.
The reason might be the domain gap between the image-text pairs and the video-question-answer triplets,
which reduces the benefit of more image-text data.



\begin{table}[]
    \centering
    \caption{
    Comparison between pure self-attention-based decoder and the cross-attention-based decoder
    under different amounts of pre-training data.
    No SCST is applied on captioning. No intermediate fine-tuning is applied for VQA.
    }
    \label{tab:cross_attention}
    \begin{tabular}{c@{~~}c@{~~}c@{~~}c@{~~}c@{~~}c@{~~}c@{~~}c@{~~}c}
    \toprule
              &          &  \multicolumn{4}{c}{Captioning}                       & \multicolumn{3}{c}{Visual Question Answering} \\
                        \cmidrule(lr){3-6} \cmidrule(lr){7-9}
data                     & Cross-Att. & COCO       & nocaps    & TextCaps  & VizWiz & ST-VQA   & TextVQA &   VizWiz\\
    \midrule
\multirow{2}{*}{0.8B}    &    w/o     & \bf144.2  & \bf120.3  & \bf143.7   & \bf107.2     & \bf65.3 & \bf58.5 & \bf59.1         \\
                         & w/         & 143.2     & 118.2     & 139.3      & 103.0  & 63.1    & 55.6    & 58.9            \\ 
    \midrule
\multirow{2}{*}{10M}     & w/o        & \bf139.1  & 75.4      & 92.7       &  \bf89.3 & 40.9    & 33.0    & 51.8            \\
                         &    w/      & 138.1     & \bf86.2   & \bf93.9    &    88.5  & \bf42.7 & \bf34.7 & \bf54.8         \\ 
    \bottomrule
    \end{tabular}
    \label{tab:my_label}
\end{table}

\subsection{Cross-attention-based decoder}
We concatenate the representations of the image and the text as the input 
to the transformer. 
An alternative way is to use a cross-attention module to incorporate the 
image representations, as in~\cite{alayrac2022flamingo, abs-2201-12086}. 
The former  
allows the image tokens to attend each other, which may refine the representation for better performance; while
the latter isolates each image token.
However, the former uses a shared set of projections for both the
image tokens and the text tokens, while the latter uses separate projection matrices. 
A shared projection may be hard to learn effectively. 
Table~\ref{tab:cross_attention} shows the comparison with different sets of pre-training image-text pairs. 
With smaller dataset, the latter with cross-attention outperforms, while with large-scale data, the former wins.
A plausible explanation is that with more pre-training data, the parameters are well optimized
such that the shared projection can adapt to both the image and the text domains,
which mitigates the drawback of the shared parameters.
With the self-attention, the image token can be attended with each other 
for a better representation
during decoding.
In all experiments, we use the former architecture.

{
\subsection{Initialization of the text decoder}
We randomly initialized the decoder. One question is whether the pre-trained weights from text corpus
can give a better performance.
Table~\ref{tab:text_decoder_init} shows the comparison results. 
We study two decoder sizes: Base (following BERT$_B$~\cite{devlin2018bert} with 12 layers) 
and Large (following BERT$_L$ with 24 layers).
As we can see, the text corpus pretrained checkpoint shows no or even worse improvement than the random initialization in both Base and Large sizes. 
This observation is also in consistent with Table 9 of ~\citep{abs-2012-06946}. 
The reason could be that the pretrained weights has no knowledge of the image signals, 
but this is the key for the VL model.
Another observation is that larger decoder also exhibits no benefit, which may be attributed to the fact of more difficulty with a larger transformer. 
Here, the discussion is based on the fact that the weights are learnable, and thus may not be applicable 
to the case with frozen parameters, 
where the initial weight is important, e.g.~\citep{alayrac2022flamingo,zeng2022socratic,xie2022visual,wang2022language}.
This is also not applicable to the case where the input is not image features, but
the text, e.g.~\citep{lin2021vx2text}.

 

\begin{table}[]
    \centering
    \caption{
    Ablation study of different initializations for the text decoder.
    The model structure and the pretraining dataset follow \modelnamewospace$_\text{B}$,
    except that the decoder transformer follows BERT$_B$~\citep{devlin2018bert} for Base and 
    BERT$_L$ for Large.
    }
    \label{tab:text_decoder_init}
    \begin{tabular}{cccccc}
    \toprule
     Decoder Size     & Initialization &  COCO  & TextCaps & VizWiz-Captions\\
     \midrule
\multirow{2}{*}{Base} &  Random         & 127.5  & 61.7     & 68.0  \\
                      &  BERT$_B$       & 126.9  & 59.7     & 67.0   \\
                      \midrule
\multirow{2}{*}{Large} & Random         & 126.6  & 54.1     & 65.0  \\
                       & BERT$_L$       & 123.9  & 52.3     & 63.6  \\
 \bottomrule
    \end{tabular}
\end{table}

}

{
\subsection{Initialization of the image encoder}
In our design, the image encoder is initialized from the contrastive pretraining.
Table~\ref{tab:initialization_image_encoder} 
shows the comparison with other initialization methods.
The setting follows \modelnamewospace$_B$.
The image encoder is the base-sized version of ViT, which
is initialized from the CLIP model,
from the supervised pretraining 
(classification task on ImageNet),
from the self-supervised pretraining (MAE~\citep{he2022masked} on ImageNet),
or randomly initialized.
From the results, we can clearly observe the higher performance with
the CLIP pretrained weights.
Compared with the supervised/self-supervised, we note that
the pretraining datasets for the image encoder are different here 
due to the availability of these weights.
Although it is unclear whether the pretraining dataset is more important
than the task or vice versa, we choose the contrastive pretraining
as the pretraining dataset is also easy to scale up, and the result is better.
For randomly initialization, we observe significant lower performance.
The reason could be the small scale of the pretraining set (10M image-text pairs in the set-up of \modelnamewospace$_B$).
A larger dataset may reduce the gap, but 
it may require longer training iterations. We leave how to effectively train the model from scratch as future work. 

\begin{table}[]
    \centering
    \caption{
    Ablation study with different initialization strategies for the image encoder under the setting of \modelnamewospace$_B$. \textit{Supervised} is pretrained
    with the classification task on ImageNet; \textit{self-supervised}
    is MAE on ImageNet.
    }
    \label{tab:initialization_image_encoder}
    \begin{tabular}{cccc}
    \toprule
                       &  COCO   & TextCaps & VizWiz-Captions \\
    \midrule
CLIP                   & 131.4   & 64.9     & 61.2 \\
Supervised             & 122.1   & 47.0     & 58.3 \\
self-supervised        & 123.4   & 44.9     & 51.6 \\        
Random                 & 89.0    & 36.5     & 38.1  \\
\bottomrule
    \end{tabular}
\end{table}

}

\begin{table}[th]
    \centering
    \small
    \caption{Effectiveness of the intermediate fine-tuning on VQA tasks. The gain is large when the target training data scale is small. }
    \label{tab:inter_ft_vqa}
    \begin{tabular}{cccccc}\toprule
                  & OCR-VQA &  VQAv2 & TextVQA & VizWiz-QA & ST-VQA \\ \midrule
    w/o inter. FT & 67.6    & 78.0   & 59.0    & 66.6      & 66.9   \\
    w/  inter. FT & 67.8    & 78.6   & 59.9    & 68.0      & 69.1   \\ \midrule
    $\Delta$      & 0.2     & 0.6    & 0.9     & 1.4       & 2.2     \\ \midrule
    Train data    & 166K    & 122K   & 22K     & 20K       & 17K    \\
    \bottomrule
    \end{tabular}
    \label{tab:my_label}
\end{table}
\subsection{Intermediate fine-tuning on VQA}
For VQA, we conduct the intermediate fine-tuning by combining multiple VQA datasets. Table~\ref{tab:inter_ft_vqa} shows the performance comparison with direct fine-tuning without the intermediate fine-tuning. From the results, we can see the intermediate fine-tuning improves the performance for all tasks, and the improvement is more if the target training data scale is small.



{
\subsection{Bias study over gender and skin}
Motivated by~\cite{zhao2021understanding}, we investigate the bias of our captioning model
as follows. 
\cite{zhao2021understanding} provides the gender type (male or female)
and the skin type (light or dark) for 
the COCO 2014 test images containing people.
As we use the Kapathy split, 
we first collect the overlapped images between the Kapathy test and  
the images with well-defined gender and skin annotations in \cite{zhao2021understanding}.
Then, we evaluate the performance on the subset images of each category.
To measure the bias, we calculate the normalized performance difference (NPD).
For example of the gender, we first obtain the metric (e.g. CIDEr) on the images annotated with
\textit{male} ($C_1$) and on the images with \textit{female} ($C_2$). Then, NPD is $|C_1 - C_2| / (C_1 + C_2)$. 
With no bias, $C_1$ equals $C_2$ and NPD is 0. If the model performs well on one group but totally fails on the other group (metric is 0), 
NPD is 1. 
Table~\ref{tab:npd_bias} 
shows the result, and we can see that the bias ranges only from 0.7\% to 5.3\% across all metrics.

\begin{table}[]
    \centering
    \caption{
    Normalized performance difference for gender (male vs female) and skin (light vs dark), the annotation of which is provided in \cite{zhao2021understanding}.
    The value ranges from 0 to 1, and 0 means no bias at all. The lower, the better.
    }
    \label{tab:npd_bias}
    \begin{tabular}{ccccc}
    \toprule
                & B4     & M     & C     & S    \\
    \midrule
    Gender      & 0.7\%  & 0.9\% & 2.0\% & 2.1\% \\
    Skin        & 4.2\%  & 2.3\% & 5.3\% & 2.2\% \\
    \bottomrule
    \end{tabular}
\end{table}
}

\subsection{Scene text in pre-training data}
\begin{figure*}[]
\centering
\includegraphics[width=0.82\textwidth]{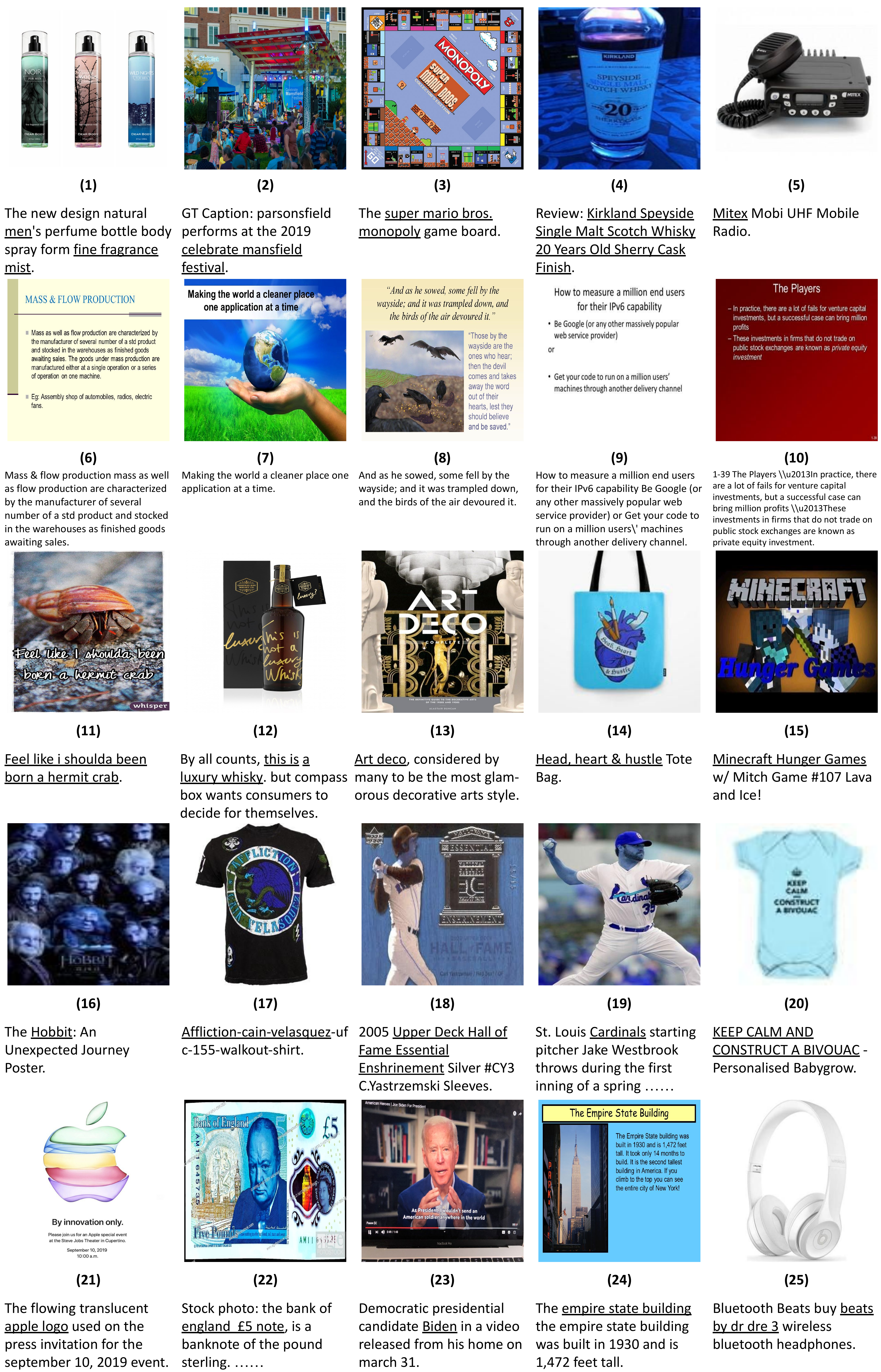}
\caption{Grouped representative pre-training samples from CC12M and the downloaded images. We highlight the informative descriptions in underline. (1-5) Pre-training samples with scene text descriptions.  (6-10) Long pieces of scene texts.  (11-15) Scene texts in stylized fonts.  (16-20) Hard samples of blurry, occluded, or curved scene texts. (21-25) Pre-training data also contains diverse knowledge on logos, banknotes, celebrities, landmarks, products, \emph{etc.}.}
\label{fig:visuF}
\end{figure*}

We show in the main paper that a considerable amount of pre-training samples contain scene text descriptions.
Fig.~\ref{fig:visuF} groups the pre-training samples with scene text from CC12M and the downloaded by different scenarios. 
Samples (1-5) show the associated text which contains the scene text in a natural language way.
This is in line with the requirement of scene-text related tasks, \emph{TextCaps}.
(6-10) show examples of long pieces of texts. The pre-training samples also describe scene text in stylized fonts (11-15), leading to \modelnamewospace's ability in robust scene text recognition. 
(16-20) contain pre-training examples with low-quality images and occluded/blurry/curved scene texts. In addition to the scene text pre-training samples, pre-training datasets also contain descriptions of diverse entities, \emph{e.g.}, the ``apple logo'' in (21), banknotes in (22), celebrity ``Biden'' in (23), landmark ``empire state building'' in (24), and product ``beats headphone'' in (25). The pre-training data plays a critical role in \modelnamewospace's capability in scene text description and informative caption generation.


\bibliography{egbib}

\begin{thebibliography}{143}
\providecommand{\natexlab}[1]{#1}
\providecommand{\url}[1]{\texttt{#1}}
\expandafter\ifx\csname urlstyle\endcsname\relax
  \providecommand{\doi}[1]{doi: #1}\else
  \providecommand{\doi}{doi: \begingroup \urlstyle{rm}\Url}\fi

\bibitem[Aafaq et~al.(2019)Aafaq, Akhtar, Liu, Gilani, and
  Mian]{aafaq2019spatio}
Nayyer Aafaq, Naveed Akhtar, Wei Liu, Syed~Zulqarnain Gilani, and Ajmal Mian.
\newblock Spatio-temporal dynamics and semantic attribute enriched visual
  encoding for video captioning.
\newblock In \emph{CVPR}, 2019.

\bibitem[Agrawal et~al.(2019)Agrawal, Desai, Wang, Chen, Jain, Johnson, Batra,
  Parikh, Lee, and Anderson]{abs-1812-08658}
Harsh Agrawal, Karan Desai, Yufei Wang, Xinlei Chen, Rishabh Jain, Mark
  Johnson, Dhruv Batra, Devi Parikh, Stefan Lee, and Peter Anderson.
\newblock nocaps: novel object captioning at scale.
\newblock In \emph{ICCV}, 2019.

\bibitem[Alayrac et~al.(2022)Alayrac, Donahue, Luc, Miech, Barr, Hasson, Lenc,
  Mensch, Millican, Reynolds, et~al.]{alayrac2022flamingo}
Jean-Baptiste Alayrac, Jeff Donahue, Pauline Luc, Antoine Miech, Iain Barr,
  Yana Hasson, Karel Lenc, Arthur Mensch, Katie Millican, Malcolm Reynolds,
  et~al.
\newblock Flamingo: a visual language model for few-shot learning.
\newblock \emph{arXiv preprint arXiv:2204.14198}, 2022.

\bibitem[Anderson et~al.(2018)Anderson, He, Buehler, Teney, Johnson, Gould, and
  Zhang]{00010BT0GZ18}
Peter Anderson, Xiaodong He, Chris Buehler, Damien Teney, Mark Johnson, Stephen
  Gould, and Lei Zhang.
\newblock Bottom-up and top-down attention for image captioning and visual
  question answering.
\newblock In \emph{CVPR}, 2018.

\bibitem[Biten et~al.(2019)Biten, Tito, Mafla, Gomez, Rusinol, Valveny,
  Jawahar, and Karatzas]{biten2019scene}
Ali~Furkan Biten, Ruben Tito, Andres Mafla, Lluis Gomez, Mar{\c{c}}al Rusinol,
  Ernest Valveny, CV~Jawahar, and Dimosthenis Karatzas.
\newblock Scene text visual question answering.
\newblock In \emph{ICCV}, 2019.

\bibitem[Biten et~al.(2022)Biten, Litman, Xie, Appalaraju, and
  Manmatha]{biten2022latr}
Ali~Furkan Biten, Ron Litman, Yusheng Xie, Srikar Appalaraju, and R~Manmatha.
\newblock Latr: Layout-aware transformer for scene-text vqa.
\newblock In \emph{CVPR}, 2022.

\bibitem[Brown et~al.(2020)Brown, Mann, Ryder, Subbiah, Kaplan, Dhariwal,
  Neelakantan, Shyam, Sastry, Askell, Agarwal, Herbert{-}Voss, Krueger,
  Henighan, Child, Ramesh, Ziegler, Wu, Winter, Hesse, Chen, Sigler, Litwin,
  Gray, Chess, Clark, Berner, McCandlish, Radford, Sutskever, and
  Amodei]{abs-2005-14165}
Tom~B. Brown, Benjamin Mann, Nick Ryder, Melanie Subbiah, Jared Kaplan,
  Prafulla Dhariwal, Arvind Neelakantan, Pranav Shyam, Girish Sastry, Amanda
  Askell, Sandhini Agarwal, Ariel Herbert{-}Voss, Gretchen Krueger, Tom
  Henighan, Rewon Child, Aditya Ramesh, Daniel~M. Ziegler, Jeffrey Wu, Clemens
  Winter, Christopher Hesse, Mark Chen, Eric Sigler, Mateusz Litwin, Scott
  Gray, Benjamin Chess, Jack Clark, Christopher Berner, Sam McCandlish, Alec
  Radford, Ilya Sutskever, and Dario Amodei.
\newblock Language models are few-shot learners.
\newblock \emph{arXiv preprint arXiv:2005.14165}, 2020.

\bibitem[Changpinyo et~al.(2021)Changpinyo, Sharma, Ding, and
  Soricut]{changpinyo2021cc12m}
Soravit Changpinyo, Piyush Sharma, Nan Ding, and Radu Soricut.
\newblock {Conceptual 12M}: Pushing web-scale image-text pre-training to
  recognize long-tail visual concepts.
\newblock In \emph{CVPR}, 2021.

\bibitem[Chen \& Dolan(2011)Chen and Dolan]{chen-dolan-2011-collecting}
David Chen and William Dolan.
\newblock Collecting highly parallel data for paraphrase evaluation.
\newblock In \emph{ACL}, 2011.

\bibitem[Chen \& Jiang(2019)Chen and Jiang]{chen2019motion}
Shaoxiang Chen and Yu-Gang Jiang.
\newblock Motion guided spatial attention for video captioning.
\newblock In \emph{AAAI}, 2019.

\bibitem[Chen et~al.(2020{\natexlab{a}})Chen, Jiang, Liu, and
  Jiang]{chen2020learning}
Shaoxiang Chen, Wenhao Jiang, Wei Liu, and Yu-Gang Jiang.
\newblock Learning modality interaction for temporal sentence localization and
  event captioning in videos.
\newblock In \emph{ECCV}, 2020{\natexlab{a}}.

\bibitem[Chen et~al.(2018)Chen, Wang, Zhang, and Huang]{chen2018less}
Yangyu Chen, Shuhui Wang, Weigang Zhang, and Qingming Huang.
\newblock Less is more: Picking informative frames for video captioning.
\newblock In \emph{ECCV}, 2018.

\bibitem[Chen et~al.(2020{\natexlab{b}})Chen, Li, Yu, Kholy, Ahmed, Gan, Cheng,
  and Liu]{ChenLYK0G0020}
Yen{-}Chun Chen, Linjie Li, Licheng Yu, Ahmed~El Kholy, Faisal Ahmed, Zhe Gan,
  Yu~Cheng, and Jingjing Liu.
\newblock {UNITER:} universal image-text representation learning.
\newblock In \emph{ECCV}, 2020{\natexlab{b}}.

\bibitem[Cho et~al.(2021)Cho, Lei, Tan, and Bansal]{cho2021unifying}
Jaemin Cho, Jie Lei, Hao Tan, and Mohit Bansal.
\newblock Unifying vision-and-language tasks via text generation.
\newblock In \emph{ICML}, 2021.

\bibitem[Chowdhery et~al.(2022)Chowdhery, Narang, Devlin, Bosma, Mishra,
  Roberts, Barham, Chung, Sutton, Gehrmann, et~al.]{chowdhery2022palm}
Aakanksha Chowdhery, Sharan Narang, Jacob Devlin, Maarten Bosma, Gaurav Mishra,
  Adam Roberts, Paul Barham, Hyung~Won Chung, Charles Sutton, Sebastian
  Gehrmann, et~al.
\newblock Palm: Scaling language modeling with pathways.
\newblock \emph{arXiv preprint arXiv:2204.02311}, 2022.

\bibitem[Cornia et~al.(2021)Cornia, Baraldi, Fiameni, and
  Cucchiara]{abs-2111-12727}
Marcella Cornia, Lorenzo Baraldi, Giuseppe Fiameni, and Rita Cucchiara.
\newblock Universal captioner: Long-tail vision-and-language model training
  through content-style separation.
\newblock \emph{arXiv preprint arXiv:2111.12727}, 2021.

\bibitem[Deng et~al.(2009)Deng, Dong, Socher, Li, Li, and
  Fei-Fei]{imagenet_cvpr09}
J.~Deng, W.~Dong, R.~Socher, L.-J. Li, K.~Li, and L.~Fei-Fei.
\newblock {ImageNet: A Large-Scale Hierarchical Image Database}.
\newblock In \emph{CVPR09}, 2009.

\bibitem[Denkowski \& Lavie(2014)Denkowski and Lavie]{DenkowskiL14}
Michael~J. Denkowski and Alon Lavie.
\newblock Meteor universal: Language specific translation evaluation for any
  target language.
\newblock In \emph{WMT@ACL}, 2014.

\bibitem[Desai \& Johnson(2021)Desai and Johnson]{virtex}
Karan Desai and Justin Johnson.
\newblock Virtex: Learning visual representations from textual annotations.
\newblock In \emph{CVPR}, 2021.

\bibitem[Devlin et~al.(2018)Devlin, Chang, Lee, and Toutanova]{devlin2018bert}
Jacob Devlin, Ming-Wei Chang, Kenton Lee, and Kristina Toutanova.
\newblock Bert: Pre-training of deep bidirectional transformers for language
  understanding.
\newblock \emph{arXiv preprint arXiv:1810.04805}, 2018.

\bibitem[Ding et~al.(2022)Ding, Xiao, Codella, Luo, Wang, and
  Yuan]{ding2022davit}
Mingyu Ding, Bin Xiao, Noel Codella, Ping Luo, Jingdong Wang, and Lu~Yuan.
\newblock Davit: Dual attention vision transformers.
\newblock 2022.

\bibitem[Dosovitskiy et~al.(2021)Dosovitskiy, Beyer, Kolesnikov, Weissenborn,
  Zhai, Unterthiner, Dehghani, Minderer, Heigold, Gelly, Uszkoreit, and
  Houlsby]{DosovitskiyB0WZ21}
Alexey Dosovitskiy, Lucas Beyer, Alexander Kolesnikov, Dirk Weissenborn,
  Xiaohua Zhai, Thomas Unterthiner, Mostafa Dehghani, Matthias Minderer, Georg
  Heigold, Sylvain Gelly, Jakob Uszkoreit, and Neil Houlsby.
\newblock An image is worth 16x16 words: Transformers for image recognition at
  scale.
\newblock In \emph{ICLR}, 2021.

\bibitem[Dou et~al.(2021)Dou, Xu, Gan, Wang, Wang, Wang, Zhu, Zhang, Yuan,
  Peng, Liu, and Zeng]{abs-2111-02387}
Zi{-}Yi Dou, Yichong Xu, Zhe Gan, Jianfeng Wang, Shuohang Wang, Lijuan Wang,
  Chenguang Zhu, Pengchuan Zhang, Lu~Yuan, Nanyun Peng, Zicheng Liu, and
  Michael Zeng.
\newblock An empirical study of training end-to-end vision-and-language
  transformers.
\newblock \emph{arXiv preprint arXiv: 2111.02387}, 2021.

\bibitem[Fang et~al.(2021{\natexlab{a}})Fang, Xie, Wang, Mao, and
  Zhang]{fang2021read}
Shancheng Fang, Hongtao Xie, Yuxin Wang, Zhendong Mao, and Yongdong Zhang.
\newblock Read like humans: autonomous, bidirectional and iterative language
  modeling for scene text recognition.
\newblock In \emph{CVPR}, 2021{\natexlab{a}}.

\bibitem[Fang et~al.(2021{\natexlab{b}})Fang, Wang, Hu, Liang, Gan, Wang, Yang,
  and Liu]{abs-2112-05230}
Zhiyuan Fang, Jianfeng Wang, Xiaowei Hu, Lin Liang, Zhe Gan, Lijuan Wang,
  Yezhou Yang, and Zicheng Liu.
\newblock Injecting semantic concepts into end-to-end image captioning.
\newblock \emph{arXiv preprint arXiv:2112.05230}, 2021{\natexlab{b}}.

\bibitem[Fang et~al.(2021{\natexlab{c}})Fang, Wang, Hu, Wang, Yang, and
  Liu]{FangW0WY021}
Zhiyuan Fang, Jianfeng Wang, Xiaowei Hu, Lijuan Wang, Yezhou Yang, and Zicheng
  Liu.
\newblock Compressing visual-linguistic model via knowledge distillation.
\newblock In \emph{ICCV}, 2021{\natexlab{c}}.

\bibitem[Fu et~al.(2021)Fu, Li, Gan, Lin, Wang, Wang, and Liu]{abs-2111-12681}
Tsu{-}Jui Fu, Linjie Li, Zhe Gan, Kevin Lin, William~Yang Wang, Lijuan Wang,
  and Zicheng Liu.
\newblock {VIOLET} : End-to-end video-language transformers with masked
  visual-token modeling.
\newblock \emph{arXiv preprint arXiv:2111.12681}, 2021.

\bibitem[Gan et~al.(2020)Gan, Chen, Li, Zhu, Cheng, and Liu]{Gan0LZ0020}
Zhe Gan, Yen{-}Chun Chen, Linjie Li, Chen Zhu, Yu~Cheng, and Jingjing Liu.
\newblock Large-scale adversarial training for vision-and-language
  representation learning.
\newblock In \emph{NeurIPS}, 2020.

\bibitem[Gao et~al.(2020)Gao, Zhu, Wang, Li, Liu, van~den Hengel, and
  Wu]{abs-2006-00753}
Chenyu Gao, Qi~Zhu, Peng Wang, Hui Li, Yuliang Liu, Anton van~den Hengel, and
  Qi~Wu.
\newblock Structured multimodal attentions for textvqa.
\newblock \emph{arXiv preprint arXiv:2006.00753}, 2020.

\bibitem[Gong et~al.(2021)Gong, Zhu, Wang, Chen, Zhang, Shu, and
  Liu]{GongZWCZSL21}
Xuchao Gong, Hongji Zhu, Yongliang Wang, Biaolong Chen, Aixi Zhang, Fangxun
  Shu, and Si~Liu.
\newblock Multiple transformer mining for vizwiz image caption.
\newblock In \emph{2021 VizWiz Grand Challenge Workshop}, 2021.

\bibitem[Goyal et~al.(2017)Goyal, Khot, Summers{-}Stay, Batra, and
  Parikh]{GoyalKSBP16}
Yash Goyal, Tejas Khot, Douglas Summers{-}Stay, Dhruv Batra, and Devi Parikh.
\newblock Making the {V} in {VQA} matter: Elevating the role of image
  understanding in visual question answering.
\newblock In \emph{CVPR}, 2017.

\bibitem[Graves et~al.(2006)Graves, Fern{\'a}ndez, Gomez, and
  Schmidhuber]{graves2006connectionist}
Alex Graves, Santiago Fern{\'a}ndez, Faustino Gomez, and J{\"u}rgen
  Schmidhuber.
\newblock Connectionist temporal classification: labelling unsegmented sequence
  data with recurrent neural networks.
\newblock In \emph{ICML}, 2006.

\bibitem[Gupta et~al.(2016)Gupta, Vedaldi, and Zisserman]{gupta2016synthetic}
Ankush Gupta, Andrea Vedaldi, and Andrew Zisserman.
\newblock Synthetic data for text localisation in natural images.
\newblock In \emph{CVPR}, pp.\  2315--2324, 2016.

\bibitem[Gurari et~al.(2018)Gurari, Li, Stangl, Guo, Lin, Grauman, Luo, and
  Bigham]{Gurari0SGLGLB18}
Danna Gurari, Qing Li, Abigale~J. Stangl, Anhong Guo, Chi Lin, Kristen Grauman,
  Jiebo Luo, and Jeffrey~P. Bigham.
\newblock Vizwiz grand challenge: Answering visual questions from blind people.
\newblock In \emph{CVPR}, 2018.

\bibitem[Gurari et~al.(2020)Gurari, Zhao, Zhang, and
  Bhattacharya]{abs-2002-08565}
Danna Gurari, Yinan Zhao, Meng Zhang, and Nilavra Bhattacharya.
\newblock Captioning images taken by people who are blind.
\newblock \emph{arXiv preprint arXiv:2002.08565}, 2020.

\bibitem[Han et~al.(2020)Han, Huang, and Han]{HanHH20}
Wei Han, Hantao Huang, and Tao Han.
\newblock Finding the evidence: Localization-aware answer prediction for text
  visual question answering.
\newblock In \emph{COLING}, 2020.

\bibitem[He et~al.(2022{\natexlab{a}})He, Chen, Xie, Li, Doll{\'a}r, and
  Girshick]{he2022masked}
Kaiming He, Xinlei Chen, Saining Xie, Yanghao Li, Piotr Doll{\'a}r, and Ross
  Girshick.
\newblock Masked autoencoders are scalable vision learners.
\newblock In \emph{Proceedings of the IEEE/CVF Conference on Computer Vision
  and Pattern Recognition}, pp.\  16000--16009, 2022{\natexlab{a}}.

\bibitem[He et~al.(2022{\natexlab{b}})He, Chen, Zhang, Liu, He, Wang, and
  Du]{he2021visual}
Yue He, Chen Chen, Jing Zhang, Juhua Liu, Fengxiang He, Chaoyue Wang, and
  Bo~Du.
\newblock Visual semantics allow for textual reasoning better in scene text
  recognition.
\newblock In \emph{AAAI}, 2022{\natexlab{b}}.

\bibitem[Hou et~al.(2019)Hou, Wu, Zhao, Luo, and Jia]{hou2019joint}
Jingyi Hou, Xinxiao Wu, Wentian Zhao, Jiebo Luo, and Yunde Jia.
\newblock Joint syntax representation learning and visual cue translation for
  video captioning.
\newblock In \emph{ICCV}, 2019.

\bibitem[Hu et~al.(2020)Hu, Singh, Darrell, and Rohrbach]{HuSDR20}
Ronghang Hu, Amanpreet Singh, Trevor Darrell, and Marcus Rohrbach.
\newblock Iterative answer prediction with pointer-augmented multimodal
  transformers for textvqa.
\newblock In \emph{CVPR}, 2020.

\bibitem[Hu et~al.(2021{\natexlab{a}})Hu, Gan, Wang, Yang, Liu, Lu, and
  Wang]{abs-2111-12233}
Xiaowei Hu, Zhe Gan, Jianfeng Wang, Zhengyuan Yang, Zicheng Liu, Yumao Lu, and
  Lijuan Wang.
\newblock Scaling up vision-language pre-training for image captioning.
\newblock \emph{arXiv preprint arXiv:2111.12233}, 2021{\natexlab{a}}.

\bibitem[Hu et~al.(2021{\natexlab{b}})Hu, Yin, Lin, Wang, Zhang, Gao, and
  Liu]{abs-2009-13682}
Xiaowei Hu, Xi~Yin, Kevin Lin, Lijuan Wang, Lei Zhang, Jianfeng Gao, and
  Zicheng Liu.
\newblock {VIVO:} surpassing human performance in novel object captioning with
  visual vocabulary pre-training.
\newblock In \emph{AAAI}, 2021{\natexlab{b}}.

\bibitem[Huang et~al.(2019)Huang, Wang, Chen, and Wei]{huang2019attention}
Lun Huang, Wenmin Wang, Jie Chen, and Xiao-Yong Wei.
\newblock Attention on attention for image captioning.
\newblock In \emph{ICCV}, 2019.

\bibitem[Huang et~al.(2020)Huang, Zeng, Liu, Fu, and Fu]{abs-2004-00849}
Zhicheng Huang, Zhaoyang Zeng, Bei Liu, Dongmei Fu, and Jianlong Fu.
\newblock Pixel-bert: Aligning image pixels with text by deep multi-modal
  transformers.
\newblock \emph{arXiv preprint arXiv:2004.00849}, 2020.

\bibitem[Hudson \& Manning(2019)Hudson and Manning]{HudsonM19}
Drew~A. Hudson and Christopher~D. Manning.
\newblock {GQA:} {A} new dataset for real-world visual reasoning and
  compositional question answering.
\newblock In \emph{CVPR}, 2019.

\bibitem[Jaderberg et~al.(2014)Jaderberg, Simonyan, Vedaldi, and
  Zisserman]{jaderberg2014synthetic}
Max Jaderberg, Karen Simonyan, Andrea Vedaldi, and Andrew Zisserman.
\newblock Synthetic data and artificial neural networks for natural scene text
  recognition.
\newblock \emph{arXiv preprint arXiv:1406.2227}, 2014.

\bibitem[Jaderberg et~al.(2016)Jaderberg, Simonyan, Vedaldi, and
  Zisserman]{jaderberg2016reading}
Max Jaderberg, Karen Simonyan, Andrea Vedaldi, and Andrew Zisserman.
\newblock Reading text in the wild with convolutional neural networks.
\newblock \emph{IJCV}, 2016.

\bibitem[Jang et~al.(2017)Jang, Song, Yu, Kim, and Kim]{JangSYKK17}
Yunseok Jang, Yale Song, Youngjae Yu, Youngjin Kim, and Gunhee Kim.
\newblock {TGIF-QA:} toward spatio-temporal reasoning in visual question
  answering.
\newblock In \emph{CVPR}, 2017.

\bibitem[Jia et~al.(2021)Jia, Yang, Xia, Chen, Parekh, Pham, Le, Sung, Li, and
  Duerig]{JiaYXCPPLSLD21}
Chao Jia, Yinfei Yang, Ye~Xia, Yi{-}Ting Chen, Zarana Parekh, Hieu Pham,
  Quoc~V. Le, Yun{-}Hsuan Sung, Zhen Li, and Tom Duerig.
\newblock Scaling up visual and vision-language representation learning with
  noisy text supervision.
\newblock In \emph{ICML}, 2021.

\bibitem[Jiang et~al.(2020)Jiang, Chen, Lin, Zhao, and Gao]{JiangCLZG20}
Jianwen Jiang, Ziqiang Chen, Haojie Lin, Xibin Zhao, and Yue Gao.
\newblock Divide and conquer: Question-guided spatio-temporal contextual
  attention for video question answering.
\newblock In \emph{AAAI}, 2020.

\bibitem[Kant et~al.(2020)Kant, Batra, Anderson, Schwing, Parikh, Lu, and
  Agrawal]{KantBASPLA20}
Yash Kant, Dhruv Batra, Peter Anderson, Alexander~G. Schwing, Devi Parikh,
  Jiasen Lu, and Harsh Agrawal.
\newblock Spatially aware multimodal transformers for textvqa.
\newblock In \emph{ECCV}, 2020.

\bibitem[Karatzas et~al.(2013)Karatzas, Shafait, Uchida, Iwamura, i~Bigorda,
  Mestre, Mas, Mota, Almazan, and De~Las~Heras]{karatzas2013icdar}
Dimosthenis Karatzas, Faisal Shafait, Seiichi Uchida, Masakazu Iwamura,
  Lluis~Gomez i~Bigorda, Sergi~Robles Mestre, Joan Mas, David~Fernandez Mota,
  Jon~Almazan Almazan, and Lluis~Pere De~Las~Heras.
\newblock Icdar 2013 robust reading competition.
\newblock In \emph{ICDAR}, 2013.

\bibitem[Karatzas et~al.(2015)Karatzas, Gomez-Bigorda, Nicolaou, Ghosh,
  Bagdanov, Iwamura, Matas, Neumann, Chandrasekhar, Lu,
  et~al.]{karatzas2015icdar}
Dimosthenis Karatzas, Lluis Gomez-Bigorda, Anguelos Nicolaou, Suman Ghosh,
  Andrew Bagdanov, Masakazu Iwamura, Jiri Matas, Lukas Neumann,
  Vijay~Ramaseshan Chandrasekhar, Shijian Lu, et~al.
\newblock Icdar 2015 competition on robust reading.
\newblock In \emph{ICDAR}, 2015.

\bibitem[Karpathy \& Li(2015)Karpathy and Li]{KarpathyL15}
Andrej Karpathy and Fei{-}Fei Li.
\newblock Deep visual-semantic alignments for generating image descriptions.
\newblock In \emph{CVPR}, 2015.

\bibitem[Kim et~al.(2021)Kim, Son, and Kim]{KimSK21}
Wonjae Kim, Bokyung Son, and Ildoo Kim.
\newblock Vilt: Vision-and-language transformer without convolution or region
  supervision.
\newblock In \emph{ICML}, 2021.

\bibitem[Kolesnikov et~al.(2020)Kolesnikov, Beyer, Zhai, Puigcerver, Yung,
  Gelly, and Houlsby]{kolesnikov2020big}
Alexander Kolesnikov, Lucas Beyer, Xiaohua Zhai, Joan Puigcerver, Jessica Yung,
  Sylvain Gelly, and Neil Houlsby.
\newblock Big transfer (bit): General visual representation learning.
\newblock In \emph{ECCV}, 2020.

\bibitem[Krishna et~al.(2016)Krishna, Zhu, Groth, Johnson, Hata, Kravitz, Chen,
  Kalantidis, Li, Shamma, Bernstein, and Li]{KrishnaZGJHKCKL16}
Ranjay Krishna, Yuke Zhu, Oliver Groth, Justin Johnson, Kenji Hata, Joshua
  Kravitz, Stephanie Chen, Yannis Kalantidis, Li{-}Jia Li, David~A. Shamma,
  Michael~S. Bernstein, and Fei{-}Fei Li.
\newblock Visual genome: Connecting language and vision using crowdsourced
  dense image annotations.
\newblock \emph{arXiv preprint arXiv:1602.07332}, 2016.

\bibitem[Le et~al.(2021)Le, Le, Venkatesh, and Tran]{LeLVT21}
Thao~Minh Le, Vuong Le, Svetha Venkatesh, and Truyen Tran.
\newblock Hierarchical conditional relation networks for multimodal video
  question answering.
\newblock \emph{IJCV}, 2021.

\bibitem[Lei et~al.(2020)Lei, Yu, Berg, and Bansal]{lei2020tvr}
Jie Lei, Licheng Yu, Tamara~L Berg, and Mohit Bansal.
\newblock Tvr: A large-scale dataset for video-subtitle moment retrieval.
\newblock In \emph{ECCV}, 2020.

\bibitem[Lei et~al.(2021)Lei, Li, Zhou, Gan, Berg, Bansal, and
  Liu]{LeiLZGBB021}
Jie Lei, Linjie Li, Luowei Zhou, Zhe Gan, Tamara~L. Berg, Mohit Bansal, and
  Jingjing Liu.
\newblock Less is more: Clipbert for video-and-language learning via sparse
  sampling.
\newblock In \emph{CVPR}, 2021.

\bibitem[Li et~al.(2022{\natexlab{a}})Li, Xu, Tian, Wang, Yan, Bi, Ye, Chen,
  Xu, Cao, et~al.]{li2022mplug}
Chenliang Li, Haiyang Xu, Junfeng Tian, Wei Wang, Ming Yan, Bin Bi, Jiabo Ye,
  Hehong Chen, Guohai Xu, Zheng Cao, et~al.
\newblock mplug: Effective and efficient vision-language learning by
  cross-modal skip-connections.
\newblock \emph{arXiv preprint arXiv:2205.12005}, 2022{\natexlab{a}}.

\bibitem[Li et~al.(2021{\natexlab{a}})Li, Selvaraju, Gotmare, Joty, Xiong, and
  Hoi]{li2021align}
Junnan Li, Ramprasaath~R Selvaraju, Akhilesh~Deepak Gotmare, Shafiq Joty,
  Caiming Xiong, and Steven Hoi.
\newblock Align before fuse: Vision and language representation learning with
  momentum distillation.
\newblock In \emph{NeurIPS}, 2021{\natexlab{a}}.

\bibitem[Li et~al.(2022{\natexlab{b}})Li, Li, Xiong, and Hoi]{abs-2201-12086}
Junnan Li, Dongxu Li, Caiming Xiong, and Steven C.~H. Hoi.
\newblock {BLIP:} bootstrapping language-image pre-training for unified
  vision-language understanding and generation.
\newblock \emph{arXiv preprint arXiv:2201.12086}, 2022{\natexlab{b}}.

\bibitem[Li et~al.(2020{\natexlab{a}})Li, Chen, Cheng, Gan, Yu, and
  Liu]{li2020hero}
Linjie Li, Yen-Chun Chen, Yu~Cheng, Zhe Gan, Licheng Yu, and Jingjing Liu.
\newblock Hero: Hierarchical encoder for video+ language omni-representation
  pre-training.
\newblock In \emph{EMNLP}, 2020{\natexlab{a}}.

\bibitem[Li et~al.(2021{\natexlab{b}})Li, Lei, Gan, Yu, Chen, Pillai, Cheng,
  Zhou, Wang, Wang, et~al.]{VALUE}
Linjie Li, Jie Lei, Zhe Gan, Licheng Yu, Yen-Chun Chen, Rohit Pillai, Yu~Cheng,
  Luowei Zhou, Xin~Eric Wang, William~Yang Wang, et~al.
\newblock Value: A multi-task benchmark for video-and-language understanding
  evaluation.
\newblock In \emph{NeurIPS}, 2021{\natexlab{b}}.

\bibitem[Li et~al.(2021{\natexlab{c}})Li, Gao, Niu, Xiao, Liu, Liu, Wu, and
  Wang]{li2020unimo}
Wei Li, Can Gao, Guocheng Niu, Xinyan Xiao, Hao Liu, Jiachen Liu, Hua Wu, and
  Haifeng Wang.
\newblock Unimo: Towards unified-modal understanding and generation via
  cross-modal contrastive learning.
\newblock In \emph{ACL}, 2021{\natexlab{c}}.

\bibitem[Li et~al.(2020{\natexlab{b}})Li, Yin, Li, Zhang, Hu, Zhang, Wang, Hu,
  Dong, Wei, Choi, and Gao]{Li0LZHZWH0WCG20}
Xiujun Li, Xi~Yin, Chunyuan Li, Pengchuan Zhang, Xiaowei Hu, Lei Zhang, Lijuan
  Wang, Houdong Hu, Li~Dong, Furu Wei, Yejin Choi, and Jianfeng Gao.
\newblock Oscar: Object-semantics aligned pre-training for vision-language
  tasks.
\newblock In \emph{ECCV}, 2020{\natexlab{b}}.

\bibitem[Liao et~al.(2019)Liao, Lyu, He, Yao, Wu, and Bai]{liao2019mask}
Minghui Liao, Pengyuan Lyu, Minghang He, Cong Yao, Wenhao Wu, and Xiang Bai.
\newblock Mask textspotter: An end-to-end trainable neural network for spotting
  text with arbitrary shapes.
\newblock \emph{PAMI}, 2019.

\bibitem[Lin \& Och(2004)Lin and Och]{LinO04}
Chin{-}Yew Lin and Franz~Josef Och.
\newblock Automatic evaluation of machine translation quality using longest
  common subsequence and skip-bigram statistics.
\newblock In \emph{ACL}, 2004.

\bibitem[Lin et~al.(2021{\natexlab{a}})Lin, Li, Lin, Ahmed, Gan, Liu, Lu, and
  Wang]{abs-2111-13196}
Kevin Lin, Linjie Li, Chung{-}Ching Lin, Faisal Ahmed, Zhe Gan, Zicheng Liu,
  Yumao Lu, and Lijuan Wang.
\newblock Swinbert: End-to-end transformers with sparse attention for video
  captioning.
\newblock \emph{arXiv preprint arXiv:2111.13196}, 2021{\natexlab{a}}.

\bibitem[Lin et~al.(2014)Lin, Maire, Belongie, Bourdev, Girshick, Hays, Perona,
  Ramanan, Doll{\'{a}}r, and Zitnick]{LinMBHPRDZ14}
Tsung{-}Yi Lin, Michael Maire, Serge~J. Belongie, Lubomir~D. Bourdev, Ross~B.
  Girshick, James Hays, Pietro Perona, Deva Ramanan, Piotr Doll{\'{a}}r, and
  C.~Lawrence Zitnick.
\newblock Microsoft {COCO:} common objects in context.
\newblock \emph{arXiv preprint arXiv:1405.0312}, 2014.

\bibitem[Lin et~al.(2021{\natexlab{b}})Lin, Bertasius, Wang, Chang, Parikh, and
  Torresani]{lin2021vx2text}
Xudong Lin, Gedas Bertasius, Jue Wang, Shih-Fu Chang, Devi Parikh, and Lorenzo
  Torresani.
\newblock Vx2text: End-to-end learning of video-based text generation from
  multimodal inputs.
\newblock In \emph{CVPR}, 2021{\natexlab{b}}.

\bibitem[Liu et~al.(2020{\natexlab{a}})Liu, Xu, Wu, Du, Jia, and
  Tan]{LiuXWDJT20}
Fen Liu, Guanghui Xu, Qi~Wu, Qing Du, Wei Jia, and Mingkui Tan.
\newblock Cascade reasoning network for text-based visual question answering.
\newblock In Chang~Wen Chen, Rita Cucchiara, Xian{-}Sheng Hua, Guo{-}Jun Qi,
  Elisa Ricci, Zhengyou Zhang, and Roger Zimmermann (eds.), \emph{ACM MM},
  2020{\natexlab{a}}.

\bibitem[Liu et~al.(2020{\natexlab{b}})Liu, Ren, and Yuan]{liu2020sibnet}
Sheng Liu, Zhou Ren, and Junsong Yuan.
\newblock Sibnet: Sibling convolutional encoder for video captioning.
\newblock \emph{IEEE TPAMI}, 2020{\natexlab{b}}.

\bibitem[Liu et~al.(2019)Liu, Ott, Goyal, Du, Joshi, Chen, Levy, Lewis,
  Zettlemoyer, and Stoyanov]{liu2019roberta}
Yinhan Liu, Myle Ott, Naman Goyal, Jingfei Du, Mandar Joshi, Danqi Chen, Omer
  Levy, Mike Lewis, Luke Zettlemoyer, and Veselin Stoyanov.
\newblock Roberta: A robustly optimized bert pretraining approach.
\newblock \emph{arXiv preprint arXiv:1907.11692}, 2019.

\bibitem[Liu et~al.(2021)Liu, Huang, Song, Wang, Zhang, and
  Pan]{vizwiz2021winner}
Yu~Liu, Lianghua Huang, Liuyihang Song, Bin Wang, Yingya Zhang, and Pan Pan.
\newblock Enhancing textual cues in multi-modal transformers for vqa.
\newblock In \emph{2021 VizWiz Grand Challenge Workshop}, 2021.

\bibitem[Loshchilov \& Hutter(2019)Loshchilov and Hutter]{adamw}
Ilya Loshchilov and Frank Hutter.
\newblock Decoupled weight decay regularization.
\newblock In \emph{ICLR}, 2019.

\bibitem[Lu et~al.(2019)Lu, Batra, Parikh, and Lee]{LuBPL19}
Jiasen Lu, Dhruv Batra, Devi Parikh, and Stefan Lee.
\newblock Vilbert: Pretraining task-agnostic visiolinguistic representations
  for vision-and-language tasks.
\newblock In \emph{NeurIPS}, 2019.

\bibitem[Luo et~al.(2020)Luo, Ji, Shi, Huang, Duan, Li, Li, Bharti, and
  Zhou]{Luo2020UniVL}
Huaishao Luo, Lei Ji, Botian Shi, Haoyang Huang, Nan Duan, Tianrui Li, Jason
  Li, Taroon Bharti, and Ming Zhou.
\newblock Univl: A unified video and language pre-training model for multimodal
  understanding and generation.
\newblock \emph{arXiv preprint arXiv:2002.06353}, 2020.

\bibitem[Lyu et~al.(2022)Lyu, Zhang, Liu, Qiao, Xu, Wu, Yao, Han, Ding, and
  Wang]{lyu2022maskocr}
Pengyuan Lyu, Chengquan Zhang, Shanshan Liu, Meina Qiao, Yangliu Xu, Liang Wu,
  Kun Yao, Junyu Han, Errui Ding, and Jingdong Wang.
\newblock Maskocr: Text recognition with masked encoder-decoder pretraining.
\newblock \emph{arXiv preprint arXiv:2206.00311}, 2022.

\bibitem[Marino et~al.(2019)Marino, Rastegari, Farhadi, and
  Mottaghi]{marino2019ok}
Kenneth Marino, Mohammad Rastegari, Ali Farhadi, and Roozbeh Mottaghi.
\newblock Ok-vqa: A visual question answering benchmark requiring external
  knowledge.
\newblock In \emph{CVPR}, 2019.

\bibitem[Mishra et~al.(2012)Mishra, Alahari, and Jawahar]{mishra2012scene}
Anand Mishra, Karteek Alahari, and CV~Jawahar.
\newblock Scene text recognition using higher order language priors.
\newblock In \emph{BMVC}, 2012.

\bibitem[Mishra et~al.(2019)Mishra, Shekhar, Singh, and
  Chakraborty]{mishra2019ocr}
Anand Mishra, Shashank Shekhar, Ajeet~Kumar Singh, and Anirban Chakraborty.
\newblock Ocr-vqa: Visual question answering by reading text in images.
\newblock In \emph{ICDAR}, 2019.

\bibitem[Ordonez et~al.(2011)Ordonez, Kulkarni, and Berg]{OrdonezKB11}
Vicente Ordonez, Girish Kulkarni, and Tamara~L. Berg.
\newblock Im2text: Describing images using 1 million captioned photographs.
\newblock In \emph{NeurIPS}, 2011.

\bibitem[Pan et~al.(2020)Pan, Cai, Huang, Lee, Gaidon, Adeli, and
  Niebles]{pan2020spatio}
Boxiao Pan, Haoye Cai, De-An Huang, Kuan-Hui Lee, Adrien Gaidon, Ehsan Adeli,
  and Juan~Carlos Niebles.
\newblock Spatio-temporal graph for video captioning with knowledge
  distillation.
\newblock In \emph{CVPR}, 2020.

\bibitem[Papineni et~al.(2002)Papineni, Roukos, Ward, and Zhu]{PapineniRWZ02}
Kishore Papineni, Salim Roukos, Todd Ward, and Wei{-}Jing Zhu.
\newblock Bleu: a method for automatic evaluation of machine translation.
\newblock In \emph{ACL}, 2002.

\bibitem[Patrick et~al.(2021)Patrick, Huang, Asano, Metze, Hauptmann,
  Henriques, and Vedaldi]{patrick2020support}
Mandela Patrick, Po-Yao Huang, Yuki Asano, Florian Metze, Alexander Hauptmann,
  Joao Henriques, and Andrea Vedaldi.
\newblock Support-set bottlenecks for video-text representation learning.
\newblock In \emph{ICLR}, 2021.

\bibitem[Phan et~al.(2013)Phan, Shivakumara, Tian, and
  Tan]{phan2013recognizing}
Trung~Quy Phan, Palaiahnakote Shivakumara, Shangxuan Tian, and Chew~Lim Tan.
\newblock Recognizing text with perspective distortion in natural scenes.
\newblock In \emph{ICCV}, 2013.

\bibitem[Qiao et~al.(2021)Qiao, Chen, Wang, Chen, Ye, Li, Qi, Gao, and
  Xie]{abs-2106-15332}
Yixuan Qiao, Hao Chen, Jun Wang, Yihao Chen, Xianbin Ye, Ziliang Li, Xianbiao
  Qi, Peng Gao, and Guotong Xie.
\newblock Winner team mia at textvqa challenge 2021: Vision-and-language
  representation learning with pre-trained sequence-to-sequence model.
\newblock \emph{arXiv preprint arXiv:2106.15332}, 2021.

\bibitem[Radford et~al.(2021)Radford, Kim, Hallacy, Ramesh, Goh, Agarwal,
  Sastry, Askell, Mishkin, Clark, Krueger, and Sutskever]{clip}
Alec Radford, Jong~Wook Kim, Chris Hallacy, Aditya Ramesh, Gabriel Goh,
  Sandhini Agarwal, Girish Sastry, Amanda Askell, Pamela Mishkin, Jack Clark,
  Gretchen Krueger, and Ilya Sutskever.
\newblock Learning transferable visual models from natural language
  supervision.
\newblock In \emph{ICML}, 2021.

\bibitem[Raffel et~al.(2020)Raffel, Shazeer, Roberts, Lee, Narang, Matena,
  Zhou, Li, and Liu]{RaffelSRLNMZLL20}
Colin Raffel, Noam Shazeer, Adam Roberts, Katherine Lee, Sharan Narang, Michael
  Matena, Yanqi Zhou, Wei Li, and Peter~J. Liu.
\newblock Exploring the limits of transfer learning with a unified text-to-text
  transformer.
\newblock \emph{J. Mach. Learn. Res.}, 2020.

\bibitem[Rennie et~al.(2017)Rennie, Marcheret, Mroueh, Ross, and Goel]{scst}
Steven~J. Rennie, Etienne Marcheret, Youssef Mroueh, Jerret Ross, and Vaibhava
  Goel.
\newblock Self-critical sequence training for image captioning.
\newblock In \emph{CVPR}, 2017.

\bibitem[Risnumawan et~al.(2014)Risnumawan, Shivakumara, Chan, and
  Tan]{risnumawan2014robust}
Anhar Risnumawan, Palaiahankote Shivakumara, Chee~Seng Chan, and Chew~Lim Tan.
\newblock A robust arbitrary text detection system for natural scene images.
\newblock \emph{Expert Systems with Applications}, 2014.

\bibitem[Seo et~al.(2021)Seo, Nagrani, and Schmid]{SeoNS21}
Paul~Hongsuck Seo, Arsha Nagrani, and Cordelia Schmid.
\newblock Look before you speak: Visually contextualized utterances.
\newblock In \emph{CVPR}, 2021.

\bibitem[Seo et~al.(2022)Seo, Nagrani, Arnab, and Schmid]{abs-2201-08264}
Paul~Hongsuck Seo, Arsha Nagrani, Anurag Arnab, and Cordelia Schmid.
\newblock End-to-end generative pretraining for multimodal video captioning.
\newblock \emph{arXiv preprint arXiv:2201.08264}, 2022.

\bibitem[Sharma et~al.(2018)Sharma, Ding, Goodman, and Soricut]{SoricutDSG18}
Piyush Sharma, Nan Ding, Sebastian Goodman, and Radu Soricut.
\newblock Conceptual captions: {A} cleaned, hypernymed, image alt-text dataset
  for automatic image captioning.
\newblock In \emph{ACL}, 2018.

\bibitem[Shen et~al.(2021)Shen, Li, Tan, Bansal, Rohrbach, Chang, Yao, and
  Keutzer]{abs-2107-06383}
Sheng Shen, Liunian~Harold Li, Hao Tan, Mohit Bansal, Anna Rohrbach, Kai{-}Wei
  Chang, Zhewei Yao, and Kurt Keutzer.
\newblock How much can {CLIP} benefit vision-and-language tasks?
\newblock \emph{arXiv preprint arXiv:2107.06383}, 2021.

\bibitem[Sidorov et~al.(2020)Sidorov, Hu, Rohrbach, and Singh]{abs-2003-12462}
Oleksii Sidorov, Ronghang Hu, Marcus Rohrbach, and Amanpreet Singh.
\newblock Textcaps: a dataset for image captioning with reading comprehension.
\newblock In \emph{ECCV}, 2020.

\bibitem[Singh et~al.(2019)Singh, Natarajan, Shah, Jiang, Chen, Batra, Parikh,
  and Rohrbach]{singh2019towards}
Amanpreet Singh, Vivek Natarajan, Meet Shah, Yu~Jiang, Xinlei Chen, Dhruv
  Batra, Devi Parikh, and Marcus Rohrbach.
\newblock Towards vqa models that can read.
\newblock In \emph{CVPR}, 2019.

\bibitem[Singh et~al.(2021)Singh, Pang, Toh, Huang, Galuba, and
  Hassner]{singh2021textocr}
Amanpreet Singh, Guan Pang, Mandy Toh, Jing Huang, Wojciech Galuba, and Tal
  Hassner.
\newblock {TextOCR}: Towards large-scale end-to-end reasoning for
  arbitrary-shaped scene text.
\newblock 2021.

\bibitem[Sun et~al.(2019)Sun, Myers, Vondrick, Murphy, and
  Schmid]{sun2019videobert}
Chen Sun, Austin Myers, Carl Vondrick, Kevin Murphy, and Cordelia Schmid.
\newblock Videobert: A joint model for video and language representation
  learning.
\newblock In \emph{ICCV}, 2019.

\bibitem[Tan \& Bansal(2019)Tan and Bansal]{TanB19}
Hao Tan and Mohit Bansal.
\newblock {LXMERT:} learning cross-modality encoder representations from
  transformers.
\newblock In \emph{EMNLP}, 2019.

\bibitem[Tang et~al.(2021)Tang, Wang, Zeng, Rao, and Li]{abs-2110-05204}
Mingkang Tang, Zhanyu Wang, Zhaoyang Zeng, Fengyun Rao, and Dian Li.
\newblock Clip4caption ++: Multi-clip for video caption.
\newblock \emph{arXiv preprint arXiv:2110.05204}, 2021.

\bibitem[Vedantam et~al.(2015)Vedantam, Zitnick, and Parikh]{VedantamZP15}
Ramakrishna Vedantam, C.~Lawrence Zitnick, and Devi Parikh.
\newblock Cider: Consensus-based image description evaluation.
\newblock In \emph{CVPR}, 2015.

\bibitem[Venugopalan et~al.(2014)Venugopalan, Xu, Donahue, Rohrbach, Mooney,
  and Saenko]{VenugopalanXDRMS14}
Subhashini Venugopalan, Huijuan Xu, Jeff Donahue, Marcus Rohrbach, Raymond~J.
  Mooney, and Kate Saenko.
\newblock Translating videos to natural language using deep recurrent neural
  networks.
\newblock \emph{arXiv preprint arXiv:1412.4729}, 2014.

\bibitem[Wang et~al.(2022{\natexlab{a}})Wang, Ge, Yan, Ge, Lin, Cai, Wu, Shan,
  Qie, and Shou]{wang2022all}
Alex~Jinpeng Wang, Yixiao Ge, Rui Yan, Yuying Ge, Xudong Lin, Guanyu Cai,
  Jianping Wu, Ying Shan, Xiaohu Qie, and Mike~Zheng Shou.
\newblock All in one: Exploring unified video-language pre-training.
\newblock \emph{arXiv preprint arXiv:2203.07303}, 2022{\natexlab{a}}.

\bibitem[Wang et~al.(2019{\natexlab{a}})Wang, Ma, Zhang, Jiang, Wang, and
  Liu]{wang2019controllable}
Bairui Wang, Lin Ma, Wei Zhang, Wenhao Jiang, Jingwen Wang, and Wei Liu.
\newblock Controllable video captioning with pos sequence guidance based on
  gated fusion network.
\newblock In \emph{ICCV}, 2019{\natexlab{a}}.

\bibitem[Wang et~al.(2020)Wang, Hu, Zhang, Li, Wang, Zhang, Gao, and
  Liu]{abs-2012-06946}
Jianfeng Wang, Xiaowei Hu, Pengchuan Zhang, Xiujun Li, Lijuan Wang, Lei Zhang,
  Jianfeng Gao, and Zicheng Liu.
\newblock Minivlm: {A} smaller and faster vision-language model.
\newblock \emph{arXiv preprint arXiv:2012.06946}, 2020.

\bibitem[Wang et~al.(2021{\natexlab{a}})Wang, Hu, Gan, Yang, Dai, Liu, Lu, and
  Wang]{ufo}
Jianfeng Wang, Xiaowei Hu, Zhe Gan, Zhengyuan Yang, Xiyang Dai, Zicheng Liu,
  Yumao Lu, and Lijuan Wang.
\newblock {UFO:} {A} unified transformer for vision-language representation
  learning.
\newblock \emph{arXiv preprint arXiv:2111.10023}, 2021{\natexlab{a}}.

\bibitem[Wang et~al.(2011)Wang, Babenko, and Belongie]{wang2011end}
Kai Wang, Boris Babenko, and Serge Belongie.
\newblock End-to-end scene text recognition.
\newblock In \emph{ICCV}, 2011.

\bibitem[Wang et~al.(2022{\natexlab{b}})Wang, Yang, Men, Lin, Bai, Li, Ma,
  Zhou, Zhou, and Yang]{abs-2202-03052}
Peng Wang, An~Yang, Rui Men, Junyang Lin, Shuai Bai, Zhikang Li, Jianxin Ma,
  Chang Zhou, Jingren Zhou, and Hongxia Yang.
\newblock Unifying architectures, tasks, and modalities through a simple
  sequence-to-sequence learning framework.
\newblock \emph{arXiv preprint arXiv:2202.03052}, 2022{\natexlab{b}}.

\bibitem[Wang et~al.(2019{\natexlab{b}})Wang, Wu, Chen, Li, Wang, and
  Wang]{wang2019vatex}
Xin Wang, Jiawei Wu, Junkun Chen, Lei Li, Yuan-Fang Wang, and William~Yang
  Wang.
\newblock Vatex: A large-scale, high-quality multilingual dataset for
  video-and-language research.
\newblock In \emph{ICCV}, 2019{\natexlab{b}}.

\bibitem[Wang et~al.(2022{\natexlab{c}})Wang, Li, Xu, Zhou, Lei, Lin, Wang,
  Yang, Zhu, Hoiem, et~al.]{wang2022language}
Zhenhailong Wang, Manling Li, Ruochen Xu, Luowei Zhou, Jie Lei, Xudong Lin,
  Shuohang Wang, Ziyi Yang, Chenguang Zhu, Derek Hoiem, et~al.
\newblock Language models with image descriptors are strong few-shot
  video-language learners.
\newblock \emph{arXiv preprint arXiv:2205.10747}, 2022{\natexlab{c}}.

\bibitem[Wang et~al.(2021{\natexlab{b}})Wang, Yu, Yu, Dai, Tsvetkov, and
  Cao]{wang2021simvlm}
Zirui Wang, Jiahui Yu, Adams~Wei Yu, Zihang Dai, Yulia Tsvetkov, and Yuan Cao.
\newblock Simvlm: Simple visual language model pretraining with weak
  supervision.
\newblock \emph{arXiv preprint arXiv:2108.10904}, 2021{\natexlab{b}}.

\bibitem[Wu et~al.(2016)Wu, Schuster, Chen, Le, Norouzi, Macherey, Krikun, Cao,
  Gao, Macherey, Klingner, Shah, Johnson, Liu, Kaiser, Gouws, Kato, Kudo,
  Kazawa, Stevens, Kurian, Patil, Wang, Young, Smith, Riesa, Rudnick, Vinyals,
  Corrado, Hughes, and Dean]{WuSCLNMKCGMKSJL16}
Yonghui Wu, Mike Schuster, Zhifeng Chen, Quoc~V. Le, Mohammad Norouzi, Wolfgang
  Macherey, Maxim Krikun, Yuan Cao, Qin Gao, Klaus Macherey, Jeff Klingner,
  Apurva Shah, Melvin Johnson, Xiaobing Liu, Lukasz Kaiser, Stephan Gouws,
  Yoshikiyo Kato, Taku Kudo, Hideto Kazawa, Keith Stevens, George Kurian,
  Nishant Patil, Wei Wang, Cliff Young, Jason Smith, Jason Riesa, Alex Rudnick,
  Oriol Vinyals, Greg Corrado, Macduff Hughes, and Jeffrey Dean.
\newblock Google's neural machine translation system: Bridging the gap between
  human and machine translation.
\newblock \emph{arXiv preprint arXiv:1609.08144}, 2016.

\bibitem[Xie et~al.(2022)Xie, Zhou, Dai, Yuan, Bach, Liu, and
  Zeng]{xie2022visual}
Yujia Xie, Luowei Zhou, Xiyang Dai, Lu~Yuan, Nguyen Bach, Ce~Liu, and Michael
  Zeng.
\newblock Visual clues: Bridging vision and language foundations for image
  paragraph captioning.
\newblock \emph{arXiv preprint arXiv:2206.01843}, 2022.

\bibitem[Xu et~al.(2017)Xu, Zhao, Xiao, Wu, Zhang, He, and Zhuang]{xu2017video}
Dejing Xu, Zhou Zhao, Jun Xiao, Fei Wu, Hanwang Zhang, Xiangnan He, and Yueting
  Zhuang.
\newblock Video question answering via gradually refined attention over
  appearance and motion.
\newblock In \emph{ACM Multimedia}, 2017.

\bibitem[Xu et~al.(2021)Xu, Niu, Tan, Luo, Du, and Wu]{abs-2105-03236}
Guanghui Xu, Shuaicheng Niu, Mingkui Tan, Yucheng Luo, Qing Du, and Qi~Wu.
\newblock Towards accurate text-based image captioning with content diversity
  exploration.
\newblock In \emph{CVPR}, 2021.

\bibitem[Xu et~al.(2016)Xu, Mei, Yao, and Rui]{xu2016msr}
Jun Xu, Tao Mei, Ting Yao, and Yong Rui.
\newblock Msr-vtt: A large video description dataset for bridging video and
  language.
\newblock In \emph{CVPR}, 2016.

\bibitem[Xue et~al.(2021{\natexlab{a}})Xue, Huang, Liu, Peng, Fu, Li, and
  Luo]{abs-2106-13488}
Hongwei Xue, Yupan Huang, Bei Liu, Houwen Peng, Jianlong Fu, Houqiang Li, and
  Jiebo Luo.
\newblock Probing inter-modality: Visual parsing with self-attention for
  vision-language pre-training.
\newblock \emph{arXiv preprint arXiv:2106.13488}, 2021{\natexlab{a}}.

\bibitem[Xue et~al.(2021{\natexlab{b}})Xue, Huang, Liu, Peng, Fu, Li, and
  Luo]{xue2021probing}
Hongwei Xue, Yupan Huang, Bei Liu, Houwen Peng, Jianlong Fu, Houqiang Li, and
  Jiebo Luo.
\newblock Probing inter-modality: Visual parsing with self-attention for
  vision-language pre-training.
\newblock In \emph{NeurIPS}, 2021{\natexlab{b}}.

\bibitem[Yang et~al.(2021{\natexlab{a}})Yang, Miech, Sivic, Laptev, and
  Schmid]{YangMSLS21}
Antoine Yang, Antoine Miech, Josef Sivic, Ivan Laptev, and Cordelia Schmid.
\newblock Just ask: Learning to answer questions from millions of narrated
  videos.
\newblock In \emph{ICCV}, 2021{\natexlab{a}}.

\bibitem[Yang et~al.(2022{\natexlab{a}})Yang, Li, Zhang, Xiao, Liu, Yuan, and
  Gao]{Yang_2022_CVPR}
Jianwei Yang, Chunyuan Li, Pengchuan Zhang, Bin Xiao, Ce~Liu, Lu~Yuan, and
  Jianfeng Gao.
\newblock Unified contrastive learning in image-text-label space.
\newblock In \emph{CVPR}, June 2022{\natexlab{a}}.

\bibitem[Yang et~al.(2021{\natexlab{b}})Yang, Gan, Wang, Hu, Ahmed, Liu, Lu,
  and Wang]{abs-2111-12085}
Zhengyuan Yang, Zhe Gan, Jianfeng Wang, Xiaowei Hu, Faisal Ahmed, Zicheng Liu,
  Yumao Lu, and Lijuan Wang.
\newblock Crossing the format boundary of text and boxes: Towards unified
  vision-language modeling.
\newblock \emph{arXiv preprint arXiv:2111.12085}, 2021{\natexlab{b}}.

\bibitem[Yang et~al.(2021{\natexlab{c}})Yang, Lu, Wang, Yin, Flor{\^{e}}ncio,
  Wang, Zhang, Zhang, and Luo]{abs-2012-04638}
Zhengyuan Yang, Yijuan Lu, Jianfeng Wang, Xi~Yin, Dinei A.~F. Flor{\^{e}}ncio,
  Lijuan Wang, Cha Zhang, Lei Zhang, and Jiebo Luo.
\newblock {TAP:} text-aware pre-training for text-vqa and text-caption.
\newblock In \emph{CVPR}, 2021{\natexlab{c}}.

\bibitem[Yang et~al.(2022{\natexlab{b}})Yang, Gan, Wang, Hu, Lu, Liu, and
  Wang]{yang2022empirical}
Zhengyuan Yang, Zhe Gan, Jianfeng Wang, Xiaowei Hu, Yumao Lu, Zicheng Liu, and
  Lijuan Wang.
\newblock An empirical study of gpt-3 for few-shot knowledge-based vqa.
\newblock In \emph{Proceedings of the AAAI Conference on Artificial
  Intelligence}, volume~36, pp.\  3081--3089, 2022{\natexlab{b}}.

\bibitem[Young et~al.(2014)Young, Lai, Hodosh, and Hockenmaier]{young2014image}
Peter Young, Alice Lai, Micah Hodosh, and Julia Hockenmaier.
\newblock From image descriptions to visual denotations: New similarity metrics
  for semantic inference over event descriptions.
\newblock \emph{Transactions of the Association for Computational Linguistics},
  2014.

\bibitem[Yu et~al.(2020)Yu, Li, Zhang, Liu, Han, Liu, and Ding]{yu2020towards}
Deli Yu, Xuan Li, Chengquan Zhang, Tao Liu, Junyu Han, Jingtuo Liu, and Errui
  Ding.
\newblock Towards accurate scene text recognition with semantic reasoning
  networks.
\newblock In \emph{CVPR}, 2020.

\bibitem[Yu et~al.(2022)Yu, Wang, Vasudevan, Yeung, Seyedhosseini, and
  Wu]{yu2022coca}
Jiahui Yu, Zirui Wang, Vijay Vasudevan, Legg Yeung, Mojtaba Seyedhosseini, and
  Yonghui Wu.
\newblock Coca: Contrastive captioners are image-text foundation models.
\newblock \emph{arXiv preprint arXiv:2205.01917}, 2022.

\bibitem[Yuan et~al.(2021)Yuan, Chen, Chen, Codella, Dai, Gao, Hu, Huang, Li,
  Li, Liu, Liu, Liu, Lu, Shi, Wang, Wang, Xiao, Xiao, Yang, Zeng, Zhou, and
  Zhang]{abs-2111-11432}
Lu~Yuan, Dongdong Chen, Yi{-}Ling Chen, Noel Codella, Xiyang Dai, Jianfeng Gao,
  Houdong Hu, Xuedong Huang, Boxin Li, Chunyuan Li, Ce~Liu, Mengchen Liu,
  Zicheng Liu, Yumao Lu, Yu~Shi, Lijuan Wang, Jianfeng Wang, Bin Xiao, Zhen
  Xiao, Jianwei Yang, Michael Zeng, Luowei Zhou, and Pengchuan Zhang.
\newblock Florence: {A} new foundation model for computer vision.
\newblock \emph{arXiv preprint arXiv:2111.11432}, 2021.

\bibitem[Yue et~al.(2020)Yue, Kuang, Lin, Sun, and Zhang]{yue2020robustscanner}
Xiaoyu Yue, Zhanghui Kuang, Chenhao Lin, Hongbin Sun, and Wayne Zhang.
\newblock Robustscanner: Dynamically enhancing positional clues for robust text
  recognition.
\newblock In \emph{ECCV}, 2020.

\bibitem[Zellers et~al.(2021)Zellers, Lu, Hessel, Yu, Park, Cao, Farhadi, and
  Choi]{ZellersLHYPCFC21}
Rowan Zellers, Ximing Lu, Jack Hessel, Youngjae Yu, Jae~Sung Park, Jize Cao,
  Ali Farhadi, and Yejin Choi.
\newblock {MERLOT:} multimodal neural script knowledge models.
\newblock In \emph{NeurIPS}, 2021.

\bibitem[Zeng et~al.(2022)Zeng, Wong, Welker, Choromanski, Tombari, Purohit,
  Ryoo, Sindhwani, Lee, Vanhoucke, et~al.]{zeng2022socratic}
Andy Zeng, Adrian Wong, Stefan Welker, Krzysztof Choromanski, Federico Tombari,
  Aveek Purohit, Michael Ryoo, Vikas Sindhwani, Johnny Lee, Vincent Vanhoucke,
  et~al.
\newblock Socratic models: Composing zero-shot multimodal reasoning with
  language.
\newblock \emph{arXiv preprint arXiv:2204.00598}, 2022.

\bibitem[Zhang \& Peng(2019)Zhang and Peng]{zhang2019object}
Junchao Zhang and Yuxin Peng.
\newblock Object-aware aggregation with bidirectional temporal graph for video
  captioning.
\newblock In \emph{CVPR}, 2019.

\bibitem[Zhang et~al.(2021{\natexlab{a}})Zhang, Li, Hu, Yang, Zhang, Wang,
  Choi, and Gao]{abs-2101-00529}
Pengchuan Zhang, Xiujun Li, Xiaowei Hu, Jianwei Yang, Lei Zhang, Lijuan Wang,
  Yejin Choi, and Jianfeng Gao.
\newblock Vinvl: Making visual representations matter in vision-language
  models.
\newblock In \emph{CVPR}, 2021{\natexlab{a}}.

\bibitem[Zhang et~al.(2020)Zhang, Shi, Yuan, Li, Wang, Hu, and
  Zha]{Zhang_2020_CVPR}
Ziqi Zhang, Yaya Shi, Chunfeng Yuan, Bing Li, Peijin Wang, Weiming Hu, and
  Zheng-Jun Zha.
\newblock Object relational graph with teacher-recommended learning for video
  captioning.
\newblock In \emph{CVPR}, 2020.

\bibitem[Zhang et~al.(2021{\natexlab{b}})Zhang, Qi, Yuan, Shan, Li, Deng, and
  Hu]{zhang2021open}
Ziqi Zhang, Zhongang Qi, Chunfeng Yuan, Ying Shan, Bing Li, Ying Deng, and
  Weiming Hu.
\newblock Open-book video captioning with retrieve-copy-generate network.
\newblock In \emph{CVPR}, 2021{\natexlab{b}}.

\bibitem[Zhao et~al.(2021)Zhao, Wang, and Russakovsky]{zhao2021understanding}
Dora Zhao, Angelina Wang, and Olga Russakovsky.
\newblock Understanding and evaluating racial biases in image captioning.
\newblock In \emph{ICCV}, 2021.

\bibitem[Zheng et~al.(2020)Zheng, Wang, and Tao]{zheng2020syntax}
Qi~Zheng, Chaoyue Wang, and Dacheng Tao.
\newblock Syntax-aware action targeting for video captioning.
\newblock In \emph{CVPR}, 2020.

\bibitem[Zhou et~al.(2018)Zhou, Xu, and Corso]{zhou2018towards}
Luowei Zhou, Chenliang Xu, and Jason~J Corso.
\newblock Towards automatic learning of procedures from web instructional
  videos.
\newblock In \emph{AAAI}, 2018.

\bibitem[Zhou et~al.(2020)Zhou, Palangi, Zhang, Hu, Corso, and
  Gao]{ZhouPZHCG20}
Luowei Zhou, Hamid Palangi, Lei Zhang, Houdong Hu, Jason~J. Corso, and Jianfeng
  Gao.
\newblock Unified vision-language pre-training for image captioning and {VQA}.
\newblock In \emph{AAAI}, 2020.

\bibitem[Zhu \& Yang(2020)Zhu and Yang]{zhu2020actbert}
Linchao Zhu and Yi~Yang.
\newblock Actbert: Learning global-local video-text representations.
\newblock In \emph{CVPR}, 2020.

\bibitem[Zhu et~al.(2019)Zhu, Guo, Yao, Lu, Liu, and Liu]{zhu2019vatex}
Xinxin Zhu, Longteng Guo, Peng Yao, Shichen Lu, Wei Liu, and Jing Liu.
\newblock Vatex video captioning challenge 2020: Multi-view features and hybrid
  reward strategies for video captioning.
\newblock \emph{arXiv preprint arXiv:1910.11102}, 2019.

\end{thebibliography}
\bibliographystyle{tmlr}

\end{document}